\documentclass[a4paper,fleqn]{cas-sc}

\usepackage[numbers]{natbib}

\usepackage{amssymb}
\usepackage{amsmath}

\usepackage{url}
\usepackage{booktabs}
\usepackage{svg}
\usepackage{graphicx}

\usepackage{subcaption}
\usepackage{caption}
\usepackage{float}
\usepackage{threeparttable}
\usepackage{hyperref}
\usepackage{chngcntr}

\def\tsc#1{\csdef{#1}{\textsc{\lowercase{#1}}\xspace}}
\tsc{WGM}
\tsc{QE}

\begin{document}
\let\WriteBookmarks\relax

\shorttitle{Interpretable ML Under the Microscope}    

\shortauthors{M. Billa et~al.}  

\title [mode = title]{Interpretable ML Under the Microscope: Performance, Meta-Features, and the Regression–Classification Predictability Gap}



%
\author[1]{Mattia Billa}[
                        orcid=0009-0005-1979-8918
                        ]

\cormark[1]


\ead{mattia.billa@unimore.it}


\credit{Conceptualization, Methodology, Software, Formal analysis, Investigation, Writing - Original Draft, Visualization}

\affiliation[1]{organization={Department of Physics, Informatics and Mathematics, University of Modena and Reggio Emilia},
    addressline={via Giuseppe Campi, 213/a}, 
    city={Modena},
    postcode={41125}, 
    country={Italy}}

\author[1]{Giovanni Orlandi}[
orcid=0009-0003-6411-4912
]
\ead{318294@studenti.unimore.it}
\credit{Software, Methodology, Visualization}

\author[1]{Veronica Guidetti}[
orcid=0000-0003-2233-0097
]

\ead{veronica.guidetti@unimore.it}
\credit{Conceptualization, Validation, Writing - Original Draft, Writing - Review \& Editing}

\author%
[1]
{Federica Mandreoli}[
orcid=0000-0002-8043-8787
]
\ead{federica.mandreoli@unimore.it}
\credit{Supervision, Conceptualization, Validation, Writing - Review \& Editing}
\cortext[cor1]{Corresponding author}

\begin{abstract}
As machine learning models are increasingly deployed in high-stakes domains, the need for interpretability has grown to meet strict regulatory and accountability constraints. Despite this interest, systematic evaluations of inherently interpretable models for tabular data remain scarce and often focus solely on aggregated performance.

To address this gap, we evaluate sixteen interpretable methods, including Explainable Boosting Machines (EBMs), Symbolic Regression (SR), and Generalized Optimal Sparse Decision Trees, across 216 real-world tabular datasets. We assess predictive accuracy, computational efficiency, and generalization under distributional shifts. Moving beyond aggregate performance rankings, we further analyze how model behavior varies with dataset meta-features and operationalize these descriptors to study algorithm selection.

Our analyses reveal a clear dichotomy: in regression tasks, models exhibit a predictable performance hierarchy dominated by EBMs and SR that can be inferred from dataset characteristics. In contrast, classification performance remains highly dataset-dependent with no stable hierarchy, showing that standard complexity measures fail to provide actionable guidance. Furthermore, we identify an ``interpretability tax'',  showing that models explicitly optimizing for structural sparsity incur significantly longer training times. Overall, these findings provide practical guidance for practitioners seeking a balance between interpretability and predictive performance, and contribute to a deeper empirical understanding of interpretable modeling for tabular data.

\end{abstract}




\maketitle
\section{Introduction}\label{sec:intro}
The field of \emph{Interpretable Machine Learning} (IML) has emerged, emphasizing models whose internal mechanisms are transparent and understandable to human users \citep{molnar2020interpretable}.
This paradigm stands in contrast to the \emph{black box} nature of complex architectures, such as deep neural networks, which prioritize predictive accuracy at the expense of explanatory insight \citep{breiman2001statistical, rudin2019we}. 

It is also necessary to distinguish \emph{intrinsic interpretability} from \emph{post-hoc explainability} (often termed XAI). Post-hoc methods attempt to explain a complex model after it has been trained; however, because these are merely approximations, they may not accurately reflect how the model actually made a decision. In contrast, intrinsically interpretable models are transparent by design, guaranteeing that the explanation is an exact representation of the model's internal logic \citep{rudin_stop_2019}.
For this reason, such techniques are increasingly adopted in domains such as healthcare, finance, and law, where accountability and the understanding of the model decision-making process are essential \citep{doshi2017towards,rudin_stop_2019}. 
This demand is further amplified by evolving regulatory frameworks, such as the European Union's AI Act and the General Data Protection Regulation (GDPR), which mandate stringent transparency requirements and the right to explanation for high-risk automated systems.
In many tabular-data applications, intrinsically interpretable models, such as linear models, sparse decision trees, or symbolic models, are already regarded as strong candidates due to their clarity and ease of validation by domain experts \citep{zeng2017interpretable, la2021contemporary}. 
Despite this growing practical importance, empirical guidance on the comparative performance of interpretable models remains limited.

Most existing benchmarking studies focus on general-purpose machine learning algorithms rather than interpretable ones 
\citep{fernandez2014we,wainer2016comparison, olson2017pmlb,scholz_comparison_2021}. 
As a result, practitioners often lack systematic evidence about 
\emph{(i)} whether some interpretable models tend to perform better than others across diverse real-world datasets, and 
\emph{(ii)} how model performance changes as a function of dataset characteristics such as dimensionality, sample size, linearity, or class imbalance.
A few recent works examine specific subclasses of interpretable models, such as Generalized Additive Models (GAMs) \citep{kruschel2025challenging}, but do not cover the broader landscape of modern IML approaches nor analyze performance variation under different data conditions.
Moreover, comparative evaluations are typically conducted on small collections of hand-selected datasets or on synthetic data, which may limit generalizability.

At the same time, the ecosystem of interpretable methods has expanded substantially over the past decade.
Techniques such as Explainable Boosting Machines (EBMs) \citep{nori2019interpretml}, Interpretable Generalized Additive Neural Networks (IGANNs) \citep{kraus2023interpretable}, 
Generalized Optimal Sparse Decision Trees (GOSDTs) \citep{mctavish2022fast}, and modern Symbolic Regression methods \citep{la2021contemporary}
provide new modeling options whose empirical properties have not yet been systematically compared. Consequently, users are faced with an increasingly fragmented methodological landscape without clear, data-driven recommendations for model selection.

\paragraph{Our contribution}
To address these gaps, we present what is, to the best of our knowledge, the most extensive comparative study of intrinsically interpretable machine learning models to date \footnote{The supplementary material and the code required to reproduce all experiments are available at: \url{https://github.com/mattiabilla/CA-IML}}. 

\begin{itemize}
    \item Using the full Penn Machine Learning Benchmark (PMLB) repository \citep{olson2017pmlb}, we evaluate 16 interpretable models across 137 regression and 79 classification datasets. Our analysis includes both traditional and recent IML methods with stable, well-maintained implementations (9 regression methods and 7 classification methods), assessing both predictive performance and training time. Our experimental design includes 4-fold cross-validation, systematic hyperparameter optimization, and rigorous preprocessing, resulting in over 14,000 model evaluations and more than 700,000 individual training runs.
    
    \item We explicitly evaluate generalization robustness by contrasting standard \emph{in-sample} performance with \emph{out-of-sample} settings. This approach simulates the phenomenon of distributional shift (covariate or label shift), allowing us to assess how well models maintain their performance when the testing distribution diverges from the training data.
    
    \item Beyond aggregated results, we investigate how model performance varies with key dataset characteristics, also called meta-features, including structural properties (e.g., number of features and samples), degree of linearity, and class imbalance \citep{maciel_measuring_2016,lorena_how_2020}. This stratified analysis provides deeper insights into the conditions under which different interpretable models excel or underperform, offering practical guidance for both researchers and practitioners.
    
    \item Furthermore, we operationalize these meta-features to address the Algorithm Selection Problem. By training predictive meta-models, we quantify inherent dataset difficulty and estimate pairwise algorithm superiority, demonstrating the extent to which dataset characteristics can reliably guide a priori model selection across different domains.
\end{itemize}

While \emph{interpretability} is a multifaceted concept, many of its aspects rely on subjective human evaluation \citep{doshi2017towards, lipton2018mythos, arrieta2020explainable}. To enable a consistent large-scale empirical comparison, we focus on structural proxies of interpretability, such as sparsity in linear models and tree size in decision-based methods. Because interpretability is not directly comparable across heterogeneous model classes in a consistent and objective manner, we evaluate intrinsically interpretable methods primarily through their empirical effectiveness, i.e., their predictive performance across datasets with diverse characteristics, and analyze how this effectiveness varies with dataset meta-features.

Our empirical analysis uncovers a fundamental dichotomy between regression and classification tasks regarding both performance hierarchy and predictability. In the regression domain, we observe a clear and consistent stratification of models, where performance is strongly governed by the underlying geometric properties of the data. In contrast, classification exhibits no stable or recoverable hierarchy. Model performances are highly overlapping and strongly dataset-dependent, preventing the emergence of a consistent ranking across problems. 

By operationalizing dataset meta-features, we further demonstrate that this contrast extends to algorithm selection. In regression, meta-features enable a reliable a priori prediction of model performance and pairwise superiority. In classification, however, these same descriptors fail to provide actionable guidance: neither dataset stratification nor meta-modeling yields reliable predictive power, highlighting a fundamental limitation of current data complexity measures in this setting. Furthermore, we quantify a pervasive computational trade-off inherent to interpretable modeling. We show that enforcing structural parsimony or capturing complex non-linear relationships often requires a significant increase in training time compared to standard baselines, a "tax" that practitioners must weigh against the benefits of transparency. Collectively, these findings provide a data-driven framework for balancing predictive accuracy, structural simplicity, and computational cost in high-stakes applications.

The remainder of this paper is organized as follows. Section \ref{sec:related} reviews related work on benchmarking, focusing on interpretable models. Next, Section \ref{sec:exp_design} details the experimental setup, including model selection and dataset preprocessing. We then specify the evaluation protocol in Section \ref{sec:eval_prot}, defining the cross-validation design, generalization scenarios, performance metrics, and meta-features used for stratification. Section \ref{sec:results} presents and analyzes the empirical results for both regression and classification tasks. Finally, Section \ref{sec:discussion} contextualizes these findings within the existing literature to offer practical guidelines, while Section \ref{sec:conclusions} concludes the paper with directions for future research.

\section{Related Work}\label{sec:related}
A number of studies in the literature have conducted comparative analyses of machine learning models \citep{fernandez2014we, wainer2016comparison, olson2017pmlb, zhang2017up,  scholz_comparison_2021}. However, the majority of these works do not focus specifically on IML models. While interpretable models are sometimes included in broader comparisons, recent state-of-the-art approaches such as SR \citep{la2021contemporary}, EBM \citep{nori2019interpretml}, or GOSDT \citep{hu2019optimal,mctavish2022fast} are often omitted.

A recent work focusing on IML, \citet{kruschel2025challenging}, presents a systematic comparison of interpretable models with a focus on GAMs. Their analysis provides strong evidence against the presumed trade-off between accuracy and interpretability. Nevertheless, the scope of their study remains restricted, as other relevant interpretable approaches are not considered.

Moreover, most existing comparative evaluations, such as the work by \citet{fernandez2014we}, rely on aggregated performance results across a collection of datasets, rather than systematically studying how model performance varies under different data conditions. As a result, important nuances in model behavior may be obscured. In contrast, \citet{scholz_comparison_2021} explicitly investigate how classification methods respond to varying structural dataset characteristics, such as dimensionality and target distributions, using synthetic data to evaluate performance under controlled conditions. Their investigation reveals that small heterogeneous ensembles generally yield the best predictive performance. They further identify that specific settings favor certain algorithms: Bagged CART excels in low-dimensional, high-sample scenarios, while nearest shrunken neighbor classifiers are most effective for unbalanced datasets. This type of analysis underscores the importance of accounting for dataset properties in comparative studies, a perspective that remains largely unexplored in the context of IML. While synthetic datasets enable controlled experimentation, they may fail to fully capture the complexity, heterogeneity, and noise patterns found in real-world data, which can limit the external validity of the conclusions regarding which specific models are optimal for a given complexity scenario.

A related approach using real-world datasets was presented by \citet{kraus2023interpretable}, who stratified datasets from the PMLB repository by sample size and dimensionality. However, other critical aspects, such as dataset linearity, and the complexity of decision boundaries, remain insufficiently explored.

To systematically move beyond aggregate metrics and basic structural properties, our work draws inspiration from the meta-learning literature. Meta-learning addresses the Algorithm Selection Problem, building on the premise that no single learning algorithm is universally superior \citep{khan2020literature}. Instead of seeking a universal winner, empirical meta-learning studies map expected algorithm performance to measurable characteristics of datasets, known as meta-features \citep{vanschoren2018meta, rivolli2018characterizing, rivolli2022meta}. Existing results demonstrate that analyzing these meta-features allows researchers to accurately outline the specific ``domains of competence" of different learning algorithms \citep{lorena_how_2020}.

Notably, data complexity measures, a specialized subset of meta-features originally introduced by \citet{ho2002complexity}, have proven highly effective in this regard. By quantifying the geometrical and topological difficulty of problems (such as class overlap, linear separability, and neighborhood density), empirical studies have shown that these measures successfully predict when algorithms like nearest neighbors or specific linear classifiers will succeed or fail \citep{lorena_how_2020}. Recently, these complexity measures have been successfully adapted to the regression domain, demonstrating their effectiveness in predicting regressor performance based on target correlation and function smoothness \citep{lorena2018data}.

Another recurring limitation in many comparative evaluations is the restricted choice of benchmark datasets, often confined to a small number of manually selected examples, typically drawn from well-known repositories such as UCI \citep{uci2019}. This raises concerns regarding the generalizability and robustness of the reported findings. Moreover, rigorous statistical testing is frequently absent, further limiting the reliability of conclusions. As argued by \citet{herrmann2024position}, empirical machine learning research is often presented as confirmatory, even though comparative evaluation should instead be treated as exploratory. Comparative analyses are also commonly conducted in the context of introducing new models, which can unintentionally bias results in favor of the proposed method.

In contrast to these practices, and to prior studies that either relied on narrow subsets of datasets \citep{kruschel2025challenging} or on synthetic data designed to isolate specific conditions \citep{scholz_comparison_2021}, our evaluation will draw on a broad set of real-world datasets from PMLB, excluding only those unsuitable for our setting (see Section \ref{sec:ds_sel_pre}). This approach mitigates the risk of selection bias and ensures a more comprehensive assessment of model behavior across diverse problem types. Furthermore, to move beyond aggregate performance and gain insight into how models behave under different data conditions, we stratify results using several dataset-level metrics, or meta-features (see Section \ref{sec:ds_metrics}), enabling a more fine-grained and interpretable analysis and test generalization properties both in- and out-of-sample.

\section{Experimental Setup}\label{sec:exp_design}
This section describes the experimental setup underlying our large-scale benchmark of IML models. We first present the set of candidate models included in the study and motivate their selection. We then detail the dataset collection process and preprocessing pipeline applied to construct a consistent benchmark. Finally, we describe the computational environment and the hyperparameter search spaces and budgets used to ensure a fair and reproducible comparison across methods.

\subsection{Selection of ML Models}

\begin{table*}[ht!]

\centering
\begin{threeparttable}

\begin{tabular}{p{5.15cm} p{1.5cm} p{1.75cm} p{1.0cm} p{
2.75cm} p{1.5cm}}
\hline
\textbf{Method} & \textbf{Acronym} & \textbf{Task} & \textbf{Type} & \textbf{Reference} & \textbf{Library} \\
\hline
Linear Regression & LR & REGR & DEC & \citet{scikit-learn} & scikit-learn\tnote{1} \\
Logistic Regression & LR & CLF & DEC  & \citet{scikit-learn} & scikit-learn \\
Lasso Regression & LASSO & REGR & DEC  & \citet{scikit-learn} & scikit-learn \\
Generalized Linear Models & GLM & REGR & DEC  & \citet{scikit-learn} & scikit-learn \\
Polynomial Regression w/ LASSO & PR+LASSO & REGR & DEC  & \citet{scikit-learn} & scikit-learn \\
Explainable Boosting Machine & EBM & CLF, REGR & DEC  &\citet{nori2019interpretml} & InterpretML\tnote{2} \\
Interpretable Generalized Additive Neural Network & IGANN & CLF, REGR & DEC  & \citet{kraus2023interpretable} & IGANN\tnote{3} \\
Decision Trees & DT & CLF, REGR & SIM & \citet{scikit-learn} & scikit-learn \\
Generalized Optimal Sparse Decision Trees & GOSDT & CLF & SIM & \citet{mctavish2022fast} & GOSDT\tnote{4} \\
Naive Bayes & NB & CLF & ALG & \citet{scikit-learn} & scikit-learn \\
k-Nearest Neighbors & k-NN & CLF, REGR & ALG & \citet{scikit-learn} & scikit-learn \\
Symbolic Regression & SR & REGR & SIM & \citet{cranmer2023interpretable} & PySR\tnote{5} \\
\hline
\end{tabular}
\begin{tablenotes}
\scriptsize
\item[1]{\url{https://scikit-learn.org}}
\item[2]{\url{https://interpret.ml}}
\item[3]{\url{https://github.com/MathiasKraus/igann}}
\item[4]{\url{https://github.com/ubc-systopia/gosdt-guesses}}
\item[5]{\url{https://ai.damtp.cam.ac.uk/pysr/v1.5.9/}}
\end{tablenotes}
\caption{Overview of the interpretable models included in our comparative evaluation. For each method, we report its acronym, the type of task it addresses (regression and/or classification), its primary interpretability dimension (Simulatability [SIM], Decomposability [DEC], or Algorithmic Transparency [ALG]), key references, and the software library or implementation used in our experiments.}
\label{tab:model_list}
\end{threeparttable}

\end{table*}
For our comparative evaluation, we prioritize IML models with stable and publicly available implementations. This criterion is crucial to ensure that our results are both reproducible and relevant to practitioners deploying these models in real-world scenarios. A complete summary of the methods included in this study, along with their respective implementations, is provided in Table~\ref{tab:model_list}.

In total, we examine nine regression and seven classification methods. We selected these methods to cover a broad spectrum of interpretability and expressiveness, spanning various modeling strategies (e.g., linear, probabilistic, instance-based). Our selection includes both established, widely investigated baselines (e.g., Linear and Logistic Regression, Decision Trees, k-NN) and novel approaches (e.g., Symbolic Regression, Optimal Sparse Decision Trees).

The baseline of our analysis is formed by the family of linear models (Linear/Logistic Regression, Lasso, and Polynomial Regression). While these methods offer the highest degree of transparency, their expressiveness is limited by their fixed functional form. Increasing the level of expressiveness, we also consider Generalized Additive Models. Within this class, we specifically selected Explainable Boosting Machines \citep{nori2019interpretml} and Interpretable Generalized Additive Neural Networks  \citep{kraus2023interpretable}. We focused on these two methods because recent comparative work by \citet{kruschel2025challenging} positions them on the performance–interpretability Pareto frontier among modern GAM variants; furthermore, unlike many experimental alternatives, they offer mature and well-documented implementations.

To explore a further level of flexibility beyond the additive constraints of GAMs, we include Symbolic Regression \citep{kronberger2024symbolic}. Unlike standard parametric or semi-parametric methods, SR does not assume a pre-defined model structure; instead, it searches the space of mathematical expressions to discover the underlying functional form of the data. Given that SR is computationally demanding, we limit our evaluation to a single representative: \texttt{PySR} \citep{cranmer2023interpretable}. We selected PySR over alternative implementations (such as Operon \citep{operon} or QLattice \citep{brolos2021approach}) as recent benchmarks demonstrate that it delivers highly competitive accuracy–interpretability trade-offs \citep{de2024srbench++}, alongside a mature and efficient Python implementation.

Finally, regarding tree-based methods, we include both classical greedy Decision Trees (DTs) and Generalized Optimal Sparse Decision Trees  \citep{mctavish2022fast}. We include GOSDT because it employs a bound-driven search for provably optimal trees on binary variables. This approach results in smaller, more interpretable trees compared to greedy splitting methods used in standard DTs, while maintaining competitive predictive performance.

Altogether, our selection of these methods ensures a comprehensive evaluation that spans three core dimensions of interpretability: simulatability, decomposability, and algorithmic transparency. For a detailed discussion of each individual method, as well as a taxonomy mapping these models to their respective interpretability dimensions, we refer the reader to Appendices \ref{sec:app_models} and \ref{sec:taxonomy}.



\subsection{Dataset Selection and Preprocessing}\label{sec:ds_sel_pre}

\begin{figure*}
\centering
\includegraphics[width=0.85\textwidth]{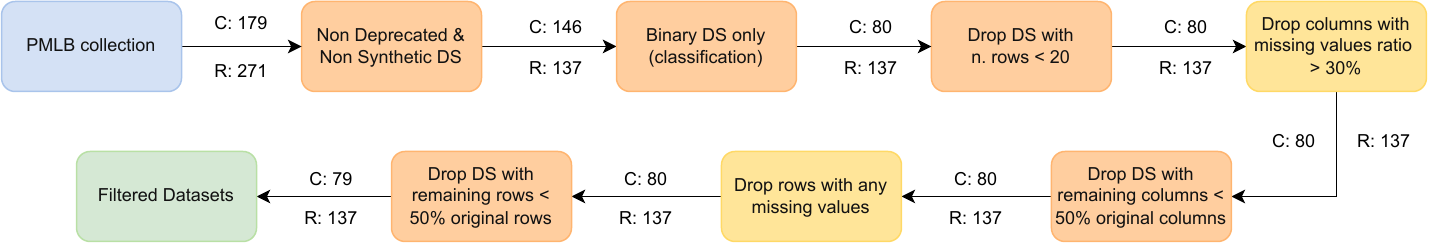}
\caption{Dataset filtering and missing values handling. Orange boxes indicate filtering steps where datasets are removed based on minimum quality criteria, while yellow boxes represent transformation steps where datasets are modified but retained. 
}\label{fig:ds_selection}
\end{figure*}

We base our study on the Penn Machine Learning Benchmark (PMLB) repository \citep{olson2017pmlb}, a curated collection of 450 datasets that has become a widely used benchmark suite, and is frequently used in studies on interpretable models, especially on SR \citep{la2021contemporary,de2024srbench++}. PMLB contains 179 classification datasets and 271 regression datasets. Not all of these are suitable for our purposes, so we apply a sequence of filtering and preprocessing steps summarized in Figure~\ref{fig:ds_selection}, where C and R denote the number of classification and regression datasets retained after each step. Orange boxes indicate filtering (dataset removal), while yellow boxes denote transformations applied to the datasets.

We first remove datasets tagged as ``deprecated'' as well as synthetic datasets. Many synthetic PMLB datasets are derived from exact physics-based equations, or similarly specialized mechanisms; including them may bias performance rankings toward models that can closely adapt to these specific functional forms, and are less representative of typical real-world applications where the generative process is unknown.

For classification, we restrict attention to binary problems. This choice reflects structural constraints of some of the interpretable models considered and simplifies the construction of out-of-sample scenarios described in Section~\ref{sec:oos}. Conceptually, this does not limit generality, as multiclass problems can be decomposed into multiple binary tasks.

Since most interpretable methods do not natively handle missing values, and imputation can distort feature distributions in a method-dependent way, we rely on simple deletion. First, we drop columns with more than 30\% missing values, retaining only those datasets where at least half of the original columns remain. Next, we remove rows containing any missing values, keeping only datasets that preserve at least 50\% of the original rows. Datasets requiring more extensive deletion are discarded, as they would be too heavily altered and could bias the results. Notably, under these criteria, only a single classification dataset was excluded. To further ensure that this deletion strategy does not introduce bias, Appendix \ref{sec:missing_values} provides a detailed quantification of the missing data, demonstrating that the proportion of removed cells in the retained datasets is negligible, with most datasets containing no missing values at all.

After these selection steps, 216 datasets remain: 137 regression datasets and 79 binary classification datasets.

\begin{figure}
\centering
\includegraphics[width=0.6\columnwidth]{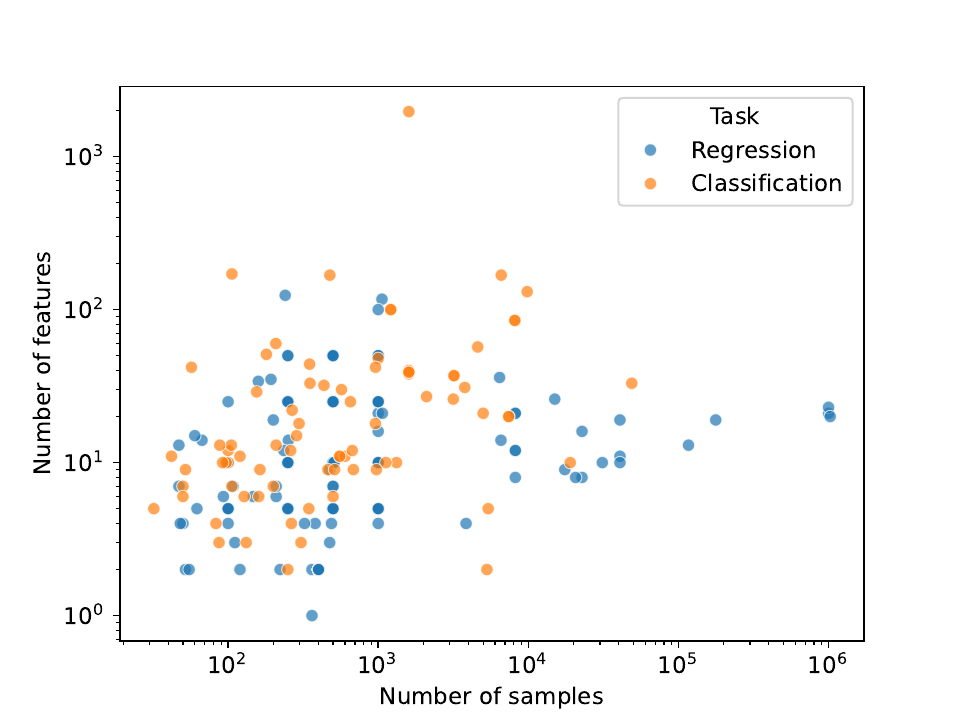}
\caption{Number of samples and features of the selected datasets.}\label{fig:pmlb}
\end{figure}

For each dataset, categorical variables are transformed via one-hot encoding and numerical features are scaled using median/IQR scaling, which reduces sensitivity to outliers (\texttt{RobustScaler} from \textit{scikit-learn}). The scaler is fitted only on the training split of each dataset, and the learned parameters are then applied to the corresponding validation and test sets to avoid information leakage. 
Figure~\ref{fig:pmlb} reports the distribution of the number of samples and features for the final set of selected and preprocessed datasets.

\subsection{Hyperparameter Optimization}

%
Hyperparameter optimization is necessary to ensure that performance comparisons are not confounded by arbitrary configuration choices. We therefore tune hyperparameters independently for each model–dataset combination.

When available, the search space was initialized based on the hyperparameter ranges suggested in the papers introducing the corresponding methods. Hyperparameter ranges were further constrained to promote smaller and more comprehensible models, in line with interpretability objectives. For example, for Decision Trees, the maximum tree depth was limited to 6, as deeper trees would typically result in hundreds or thousands of nodes, making them difficult to interpret meaningfully. The full list of hyperparameters and their corresponding search ranges used in our experiments is reported in Table D1 in Appendix \ref{sec:app_hyper}. 
The hyperparameter optimization objective was the maximization of the performance measures described in Section \ref{sec:perf_metrics}.

Hyperparameter search was conducted using the Tree-structured Parzen Estimator (TPE) algorithm \citep{bergstra2011algorithms}, as implemented in the Optuna Python library \citep{akiba2019optuna}. This Bayesian optimization method allows for a more efficient exploration of the hyperparameter space compared to random search, which is particularly advantageous for high-dimensional configurations in some of the tested models. We executed at most 50 optimization trials per setting, limiting the total optimization time to 6000 seconds, corresponding to an average of approximately 2 minutes per trial. For models with higher computational demands, such as SR, this time constraint was explicitly enforced to ensure a feasible evaluation. While unconstrained search times might yield marginal improvements for computationally heavy methods, our time budget was strictly enforced to reflect realistic operational constraints and to make such a large-scale evaluation possible.

In this study, which involves more than two hundred datasets, we balance the need for reliable hyperparameter selection with the computational constraints imposed by large-scale experimentation. To this end, we adopt a K-fold cross-validation strategy for each model–dataset combination. Hyperparameters are optimized on the validation subset of each fold using a dedicated search procedure, rather than via fully nested cross-validation, in order to keep the computational cost tractable. While less expensive than a nested scheme, this approach still provides a robust and unbiased estimate of model generalization performance.

After selecting the optimal configuration for a given fold, models are retrained on the combined training and validation data under a fixed time budget of 300 seconds, and evaluated on the held-out test set. Final performance scores are obtained by aggregating test-set results across the four folds using the median.

\subsection{Compute environment}
All experiments were executed on a shared compute cluster managed by SLURM. Each node was equipped with two Intel Xeon Gold 5320 processors (2.2 GHz), providing a total of 52 cores and 512 GB of shared memory. To guarantee a fair comparison, every individual job was allocated a single CPU core and 16 GiB of RAM, without GPU acceleration. This strict standardization enhances reproducibility and prevents discrepancies arising from differing multiprocessing or threading implementations that could otherwise bias the results.

\section{Evaluation protocol}\label{sec:eval_prot}
This section specifies the evaluation protocol used to assess model performance across datasets. We define a cross-validation design and an experimental framework to probe model robustness in both standard settings and in the presence of controlled distributional shifts. To provide a rigorous foundation for these comparisons across diverse tasks, we first establish a common notation for our data structures and predictive objectives.

Formally, we denote each dataset as $\mathcal{D} = \{(\mathbf{x}_i, y_i)\}_{i=1}^N$, where $\mathbf{x}_i \in \mathbb{R}^d$ represents the $d$-dimensional vector of processed features for the $i$-th sample, and $y_i$ denotes the corresponding ground truth target. This framework encompasses both classification and regression; the target domain is defined as $y_i \in \{0, 1\}$ for the classification task and $y_i \in \mathbb{R}$ for regression task. We let $\hat{y}_i$ denote the prediction produced by the evaluated methods for the input $\mathbf{x}_i$.
Using this notation, the following subsections detail how these data are partitioned and manipulated to simulate various testing environments.

\subsection{Cross-validation design}
We adopt a 4-fold cross-validation scheme to balance robustness of performance estimation with computational efficiency. This
provides sufficient variability across data partitions to explore different training and validation configurations, while remaining sustainable for small datasets and large-scale benchmarking.

Within each fold, data are split into training (50\%), validation (25\%), and test (25\%) subsets. This protocol is used in the in-sample experiments both for hyperparameter optimization on the validation splits and for performance estimation on the corresponding test splits. For out-of-sample experiments, hyperparameters are similarly optimized on the validation split and are not re-tuned, ensuring that performance under distributional shift is assessed without additional adaptation.

\subsection{In-sample and out-of-sample generalization}\label{sec:oos}
In our evaluation, we distinguish between two critical experimental scenarios: in-sample and out-of-sample settings. The in-sample setting represents the standard evaluation scenario where the training and test sets are drawn from the same underlying data distribution (e.g., a simple random split), allowing us to assess a model's performance on data statistically identical to what it was trained on. In contrast, the out-of-sample setting simulates the more challenging real-world phenomenon of distributional shift (also known as covariate shift or label shift), where the training and test sets exhibit different statistical properties.

To systematically assess the in- and out-of-sample generalization capabilities of the evaluated models across both regression and binary classification tasks, we designed a two-stage evaluation protocol ensuring statistical consistency within samples and controlled distributional shifts across samples. 


\paragraph{In-sample Generalization}

In-sample generalization was evaluated using stratified sampling to preserve the distributional properties of the target variable $y$ across the training and test subsets. 

\begin{itemize}
    \item \emph{Classification.} The stratification was performed with respect to the binary target variable, ensuring that class proportions remained consistent across splits. 
    \item \emph{Regression.} Since the target is continuous, we partitioned $y$ into deciles and performed stratified sampling over these bins to ensure that both subsets reflected the overall distribution of target values.
\end{itemize}

\paragraph{Out-Of-Sample Generalization}

To evaluate model robustness under distributional shifts, we constructed test sets with deliberately altered target distributions. 
\begin{itemize}
    \item \emph{Classification.} To introduce class-imbalance shifts, we modified the fraction of positive samples in the test set. Let $f$ denote the test fraction and $\alpha_{\mathcal{D}}$ the overall positive rate. We constrain the test proportion $\alpha$ to lie in the interval 
    \begin{equation}
        \frac{l}{f}\alpha_{\mathcal{D}} \le \alpha \le \frac{u}{f}\alpha_{\mathcal{D}},
    \end{equation}
    where $l$ and $u$ denote lower and upper bounds on the share of positives assignable to the test set. In our experiments, we set $f=1/4$, $l=1/8$, and $u=1/2$, yielding shifts of up to twofold differences in class proportions between training and testing subsets. To ensure strong distributional change between training and test sets, we require that 
    \begin{equation}\left\{
    \begin{array}{lll}
    \frac{l}{f}\;\alpha_\mathcal{D} \leq \alpha  \leq  \alpha_\mathcal{D} & \qquad \text{if}\quad \alpha_\mathcal{D}> 0.5\\[10pt]
    \alpha_\mathcal{D} \leq \alpha \leq \frac{u}{f} \;  \alpha_\mathcal{D} & \qquad \text{if}\quad \alpha_\mathcal{D}\leq 0.5\\[10pt]
    \end{array}\right.\end{equation}
    so that the balancing between training and test set gets reverted, when possible. 
    \item \emph{Regression.} We simulated out-of-distribution conditions by systematically excluding one quartile of the target distribution to form the test set. Each quartile was omitted once in a fourfold procedure, ensuring that each portion of the distribution was tested independently.
\end{itemize}

This procedure produces controlled target shifts that enable a fair assessment of each model’s generalization under distributional changes, while preserving sufficient statistical diversity within each dataset. 

\subsection{Performance Metrics}\label{sec:perf_metrics}
To evaluate model performance, we primarily focus on the coefficient of determination ($R^2$) for regression and the F1 score for classification. These metrics were selected as the most representative measures of predictive quality for our analysis. For regression, $R^2$ provides a normalized measure of goodness of fit that accounts for the variance (and implicitly, the noise) in the target variable. It is defined as:
\begin{equation}
R^2 = 1 - \frac{\sum_{i=1}^{n} (y_i - \hat{y}_i)^2}{\sum_{i=1}^{n} (y_i - \bar{y})^2},
\end{equation}
where $\bar{y}$ denotes the mean of the observed targets.

For classification, the F1 score offers a balanced assessment of precision and recall, making it particularly suitable for datasets where class distribution may be imbalanced. The F1 score is computed as:
\begin{equation}
\text{F1} = 2 \cdot \frac{\text{precision} \cdot \text{recall}}{\text{precision} + \text{recall}}.
\end{equation}
We also computed several other standard metrics, such as Mean Squared Error (MSE) and Mean Absolute Error (MAE) for regression, as well as precision, recall, accuracy, and Brier score for classification. Because these yielded results highly correlated with our primary metrics, we focus our discussion on $R^2$ and F1 to maintain clarity and avoid redundancy. Detailed results for all other metrics are provided in the supplementary material.






\subsection{Meta-Features for Dataset Stratification and Algorithm Selection} \label{sec:ds_metrics}
To better understand how IML models behave across heterogeneous problems, we frame our dataset characterization within the broader context of meta-learning. We characterize each dataset $\mathcal{D}$ using a small set of complexity descriptors, or meta-features, that were proposed for classification \citep{lorena_how_2020} and regression \citep{maciel_measuring_2016}. These descriptors summarize size, sparsity, class/target structure, and local neighborhood overlap, and are subsequently used to stratify performance across the PMLB benchmark.

For classification problems, let $\mathcal{C}$ denote the set of classes and $p_c$ the empirical proportion of class $c \in \mathcal{C}$. We compute metrics belonging to the following classes.

\paragraph{\textbf{Size and sparsity}}
We characterize dataset size and sparsity using the following indicators:
\begin{itemize}
    \item \emph{Number of samples ($N$).} 
    \item \emph{Number of features ($d$).} 
    \item \emph{Average number of features per point (T2):} $\text{T2} = d / N$.
    \item \emph{Average number of PCA dimensions per point (T3):} $\text{T3} = d' / N\,,\;$ where $d'$ is the number of principal components explaining $95\%$ of the variance in $\mathbf{X}$.
\end{itemize}
These meta-features jointly describe the scale and effective sparsity of a dataset. Larger values of $N$ typically stabilize empirical performance estimates, whereas small-sample regimes increase sensitivity to noise and model specification. High dimensionality ($d \gg 1$) can favor models with built-in regularization or feature selection while penalizing methods relying on local neighborhoods. The ratios T2 and T3 explicitly capture sparsity by relating dimensionality to sample size: large values indicate that data occupy a high-dimensional space relative to the number of observations, a regime in which decision boundaries are poorly supported and both parameter estimation and generalization become more challenging. In particular, T3 approximates intrinsic dimensionality after removing linear correlations, highlighting cases where many effective degrees of freedom must be learned from limited data.

\paragraph{\textbf{Classification-specific structure}}
We characterize class separability, local overlap, and class balance using the following complexity measures:
\begin{itemize}
    \item \emph{Maximum Fisher’s discriminant ratio (F1):}
    \begin{equation}
        \text{F1} = \frac{1}{1 + \max_j r_j} \in [0,1]\,;
    \end{equation}
    where
    \begin{equation} r_j =
        \frac{ \sum_{c \in \mathcal{C}} p_c (\mu_{cj} - \mu_j)^2}
             { \sum_{c \in \mathcal{C}} p_c \sigma_{cj}^2}\,\end{equation}
    and $\mu_{cj}$ and $\sigma_{cj}^2$ are the class $c\in\mathcal{C}$ conditional mean and variance of feature  $j = 1,\dots,d$, and $\mu_j$ is the global mean.

    \item \emph{Error rate of the 1-nearest neighbor classifier (N3):}
    \begin{equation}
        \text{N3} = \frac{1}{N} \sum_{i=1}^N 
        \mathbb{I}\bigl(y_i^{\text{NN}} \neq y_i\bigr) \in [0,1].
    \end{equation}
    where $y_i^{\text{NN}}$ is the label of the nearest neighbor of sample $\mathbf{x}_i$.

    \item \emph{Entropy of the 5 nearest neighbors (E5):}
    \begin{equation}
        \text{E5} = - \frac{1}{N} \sum_{i=1}^N 
        \left[p_i \log p_i + (1-p_i)\log(1-p_i)\right]\in [0,1]\, 
    \end{equation}
    where
    \begin{equation}p_i = \frac{1}{5} \sum_{j \in \mathcal{N}_5(i)} \mathbb{I}(y_j \neq y_i)\,,\end{equation}
    and $\mathcal{N}_5(i)$ is the set of the five nearest neighbors of $\mathbf{x}_i$.

    \item \emph{Entropy of class proportions $C1_{\text{c}}$:}
    \begin{equation}
        \text{C1}_{\text{c}} 
        = - \frac{1}{\log C} \sum_{c \in \mathcal{C}} p_c \log p_c \in [0,1].
    \end{equation}
    where $p_c$ is the proportion of class $c$ in the dataset and $C = |\mathcal{C}|$ the number of classes.
\end{itemize}

These indicators jointly capture different aspects of classification difficulty. The Fisher-based measure $F1$ quantifies the extent to which at least one feature can separate classes through a simple threshold. The numerator of $r_j$ captures between-class variability, while the denominator captures within-class variability. Thus, low F1 values indicate near-axis-aligned separability and higher values reflect increasingly entangled class structure. The neighborhood-based metrics N3 and E5 probe local class overlap at different spatial scales: high values indicate irregular or ambiguous decision boundaries where nearby samples frequently belong to different classes. Finally, $\text{C1}_{\text{c}}$ measures global class balance, highlighting regimes in which skewed label distributions may disproportionately affect certain learning algorithms. Together, these descriptors allow us to stratify performance by separability, locality, and imbalance.

\paragraph{\textbf{Regression-specific structure}}
We characterize functional simplicity, linearity, and local smoothness using the following complexity measures:
\begin{itemize}
    \item \emph{Maximum feature–target correlation $C1_{\text{r}} $} 
    \begin{equation}
        \text{C1}_{\text{r}} = \max_{j=1,\dots,d} |\rho(x_j, y)| \in [0,1].
    \end{equation}
    where $\rho(x_j, y)$ denotes the Spearman correlation between feature $j$ and the target.

    \item \emph{Distance to a linear model (L1):}
    \begin{equation}
        \text{L1} = \frac{1}{N} \sum_{i=1}^N |y_i - f(\mathbf{x}_i)|,
    \end{equation}
    where $f$ is a fitted multiple linear regressor on the dataset $\mathcal{D}$.

    \item \emph{Error of a 1-nearest neighbor regressor (S3):}
    \begin{equation}
        \text{S3} = \frac{1}{N} \sum_{i=1}^N \bigl(\widehat{y}_i^{\text{NN}} - y_i\bigr)^2,
    \end{equation}
    where $\widehat{y}_i^{\text{NN}}$ denotes the leave-one-out 1-NN prediction.
\end{itemize}

\begin{figure*}
    \centering
    \begin{subfigure}[b]{0.48\textwidth}
        \centering
        \includegraphics[width=\columnwidth]{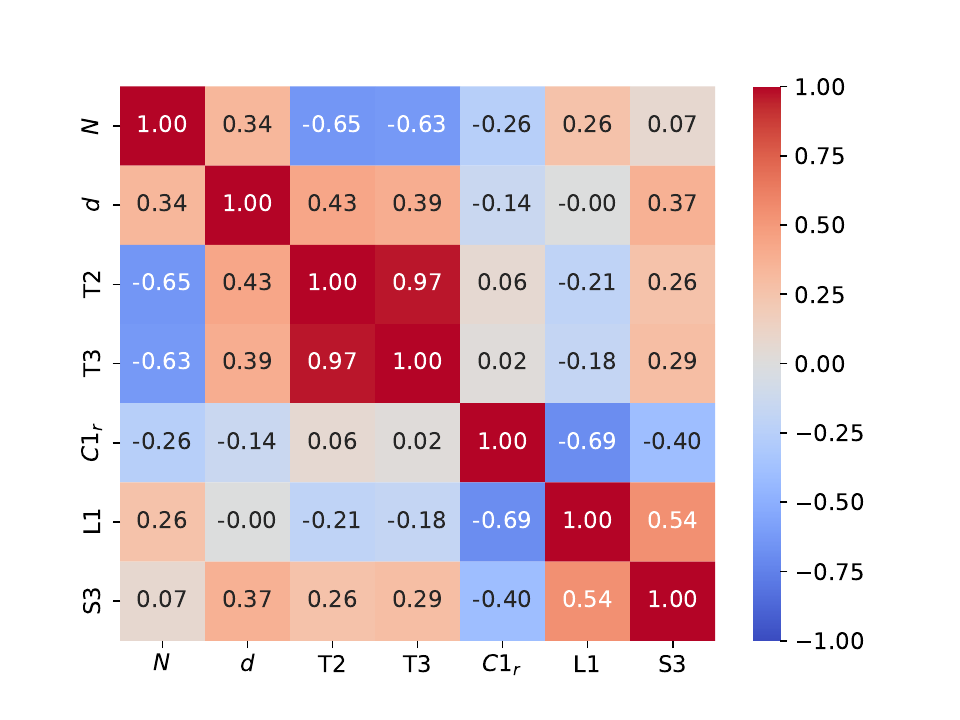}
        \caption{Regression}\label{fig:ds_metrics_corr_regr}
    \end{subfigure}
    \hfill
    \begin{subfigure}[b]{0.48\textwidth}
        \centering
        \includegraphics[width=\columnwidth]{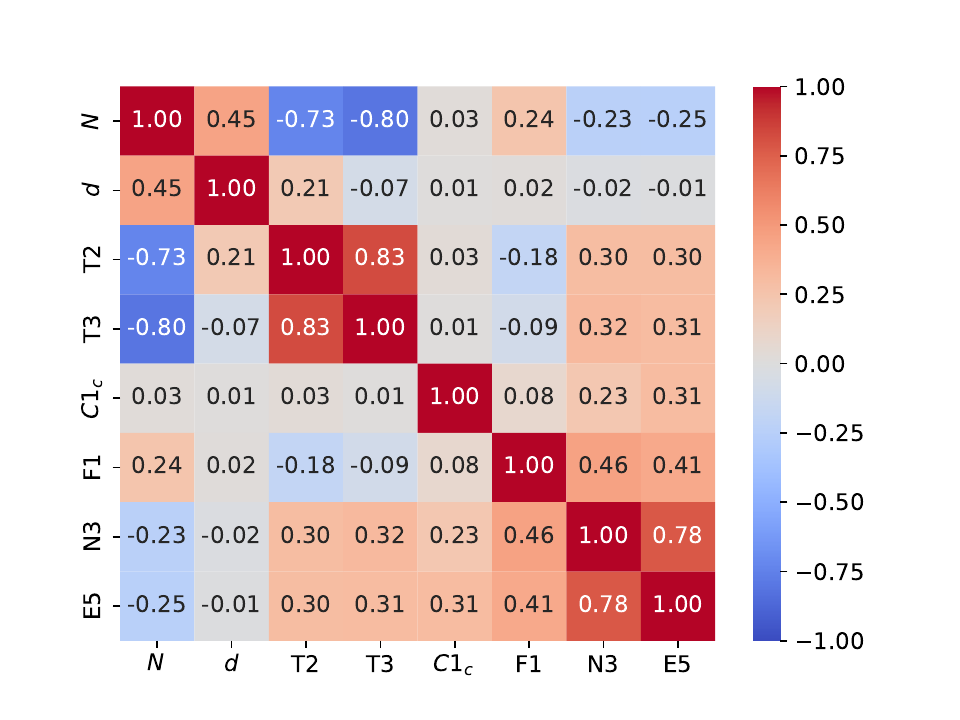}
        \caption{Classification}\label{fig:ds_metrics_corr_clf}
    \end{subfigure}
    \caption{Correlation plot of the meta-features for (a) regression and (b) classification datasets computed using the Spearman correlation.}
    \label{fig:metrics_corr}
\end{figure*}

\begin{table*}[t]
\centering

\footnotesize
\begin{tabular}{|c|l|c|c|}
\hline
\textbf{Metric} & \textbf{Measured Property} & \textbf{Complexity ($\nearrow$)} & \textbf{Tasks} \\
\hline $N$ & Dataset size   & $\searrow$ & Class., Reg. \\
\hline $d$ & Dimensionality & $\nearrow$ & Class., Reg. \\
\hline
T2 & Data sparsity & $\nearrow$ & Class., Reg. \\
\hline
T3 & Intrinsic data sparsity & $\nearrow$ & Class., Reg. \\
\hline
F1 & Single feature discrimination & $\nearrow$ & Class. \\
\hline
N3 & Local class overlap & $\nearrow$ & Class. \\
\hline
E5 & Local class overlap & $\nearrow$ & Class. \\
\hline
$\text{C1}_\text{c}$ & Class Balance & $\searrow$ & Class. \\
\hline
$\text{C1}_\text{r}$ &  Single feature discrimination & $\searrow$  & Reg. \\
\hline
L1 & Linearity level & $\nearrow$ & Reg. \\
\hline
S3 & Smoothness / Sampling quality & $\nearrow$ & Reg. \\
\hline
\end{tabular}
\caption{Meta-Features for Classification and Regression. }
\label{tab:complexity_metrics}

\end{table*}

These metrics capture complementary aspects of regression difficulty. The correlation-based measure $\text{C1}_{\text{r}}$ quantifies whether the target is strongly driven by a single feature, indicating low intrinsic complexity and near one-dimensional structure. The linearity measure L1 assesses how well the data conform to a global linear model, with low values signaling simple relationships or low noise levels, and higher values reflecting the need for increased model expressiveness. Finally, the neighborhood-based error S3 probes local smoothness of the regression surface: high values indicate that nearby points may have substantially different outputs, suggesting irregular functions or sparse sampling of the input space.

The set of meta-features considered, summarized in Table \ref{tab:complexity_metrics}, spans multiple dimensions of dataset complexity. Their complementarity is assessed by examining pairwise Spearman correlations, reported in Figures \ref{fig:ds_metrics_corr_regr} and \ref{fig:ds_metrics_corr_clf}. The predominantly low correlations across most pairs suggest these descriptors provide complementary information, justifying their joint use in our stratification. Notable exceptions include the expected structural dependencies between T2 and T3, as well as their natural correlation with $N$ and $d$. In classification, we observe a moderate correlation between N3 and E5, as both metrics probe local neighborhood purity. Similarly, for regression, $\text{C1}_{\text{r}}$ and L1 show a distinct correlation, reflecting their shared sensitivity to the linearity of the feature-target relationship.

In the following, we use these meta-features to partition the PMLB benchmark into strata and analyze whether the relative performance of interpretable methods remains stable across these regions or changes substantially under more complex data conditions. Building on this stratification, we then operationalize these metrics by training meta-models to predict both pairwise algorithm superiority and inherent dataset difficulty.

\subsection{Evaluation of Interpretability}
As discussed, measuring true cognitive notion of model comprehensibility often relies on subjective human evaluations, which lack universal consensus and are nearly impossible to standardize across diverse model architectures. Since we cannot directly compare interpretability across heterogeneous model classes, this study primarily quantifies predictive performance, generalization, and computational efficiency.

To enable a more controlled comparison, we instead assess interpretability within homogeneous model families. In this context, we operationalize interpretability through structural parsimony. As detailed in Section \ref{sec:model_compl_res}, we introduce objective proxies for model complexity: for linear models, interpretability is quantified by the number of non-zero coefficients, while for tree-based methods, we consider the total number of nodes. These measures provide consistent, quantifiable bounds on the cognitive effort required to understand the models.

\section{Results}\label{sec:results}

\begin{figure*}
    \centering
    \begin{subfigure}[b]{0.48\textwidth}
        \centering
        \includegraphics[width=\columnwidth]{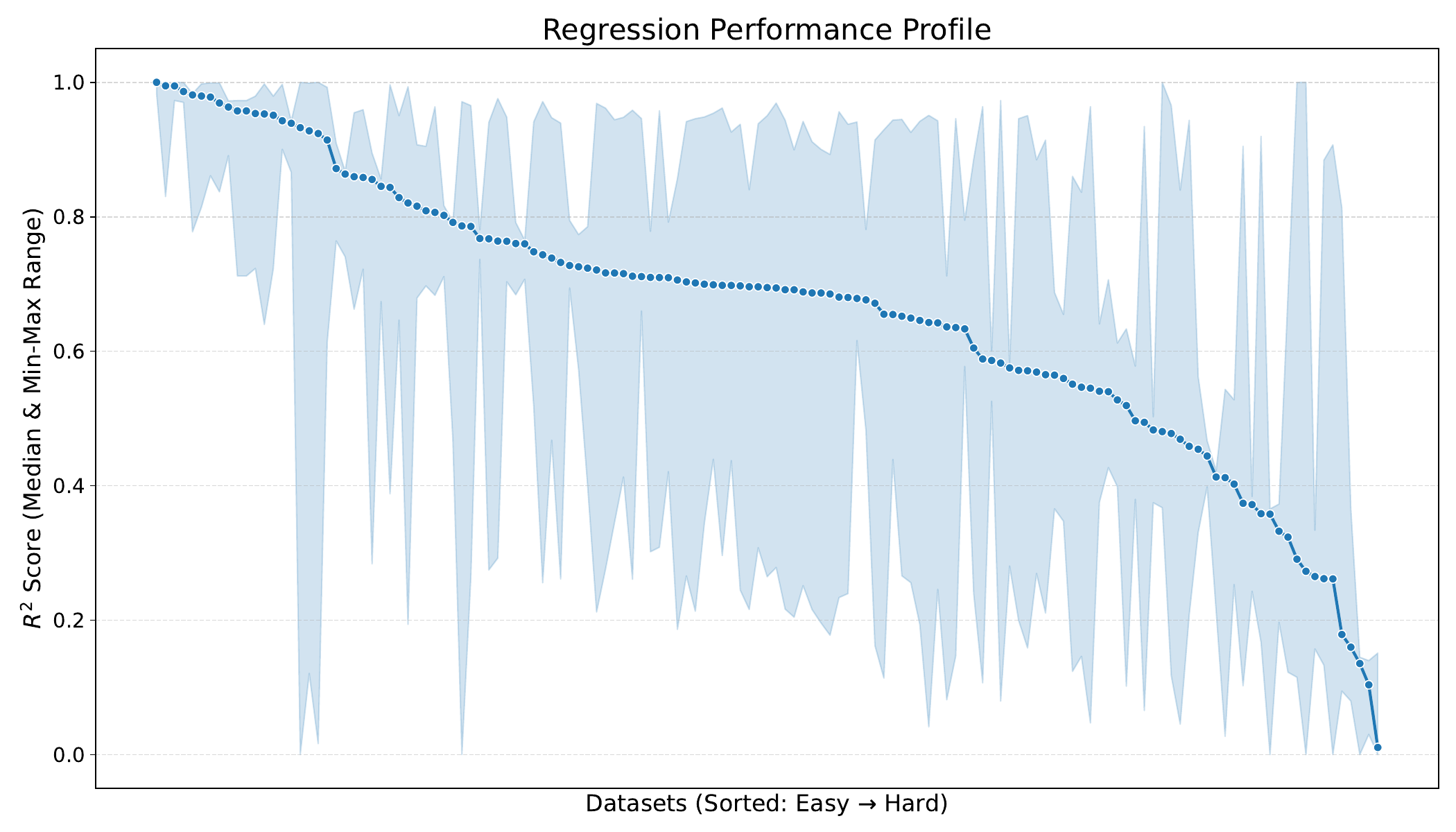}
        \caption{Regression datasets}\label{fig:regr_is_prof}
    \end{subfigure}
    \hfill
    \begin{subfigure}[b]{0.48\textwidth}
        \centering
        \includegraphics[width=\columnwidth]{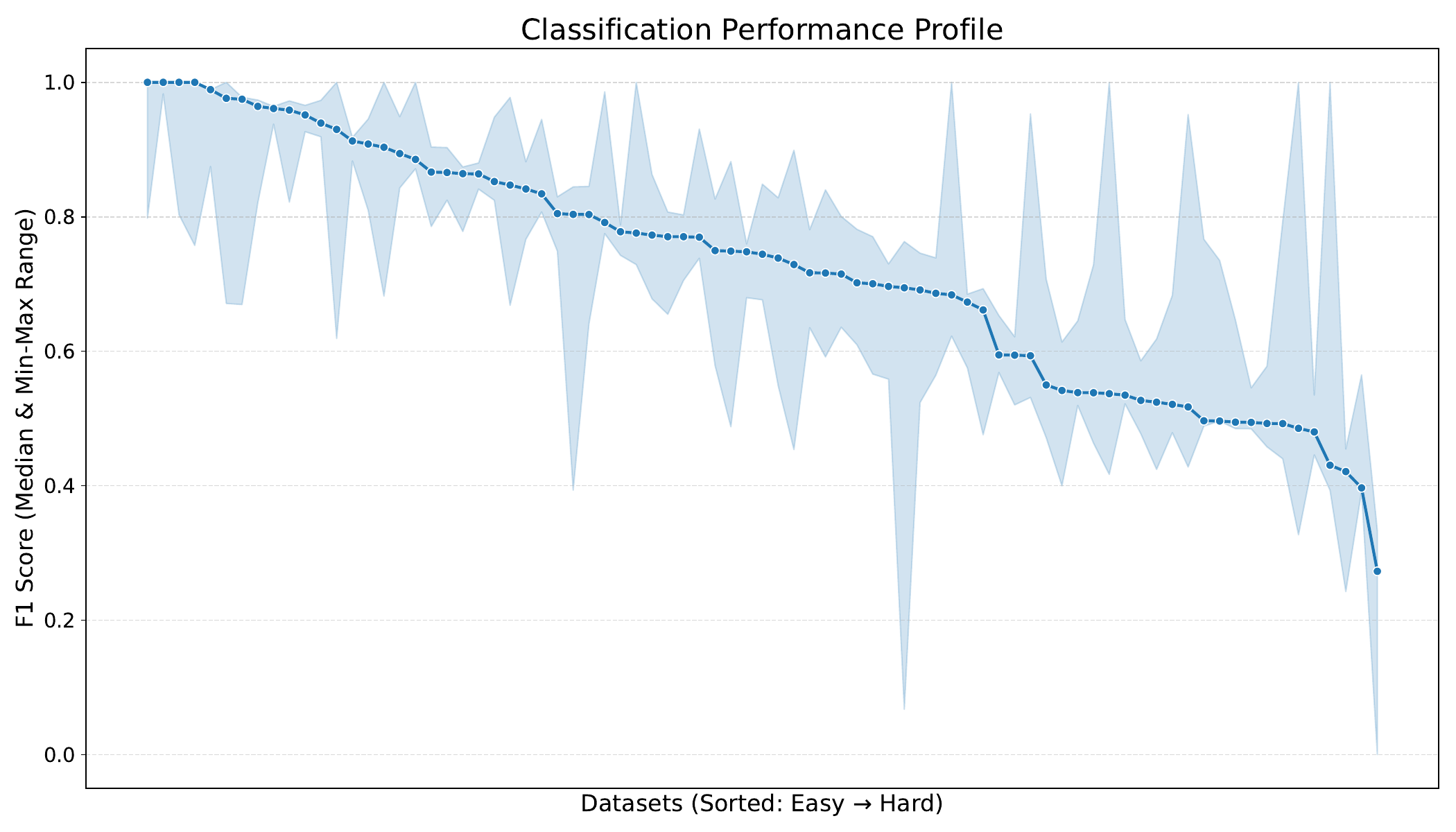}
        \caption{Classification datasets}\label{fig:clf_is_prof}
    \end{subfigure}
    \caption{(a) Regression and (b) classification performance profile across datasets (sorted along the x-axis by the median performance).}
    \label{fig:prof}
\end{figure*}

\begin{figure*}
    \centering
    \begin{subfigure}[b]{0.48\textwidth}
        \centering
        \includegraphics[width=\columnwidth]{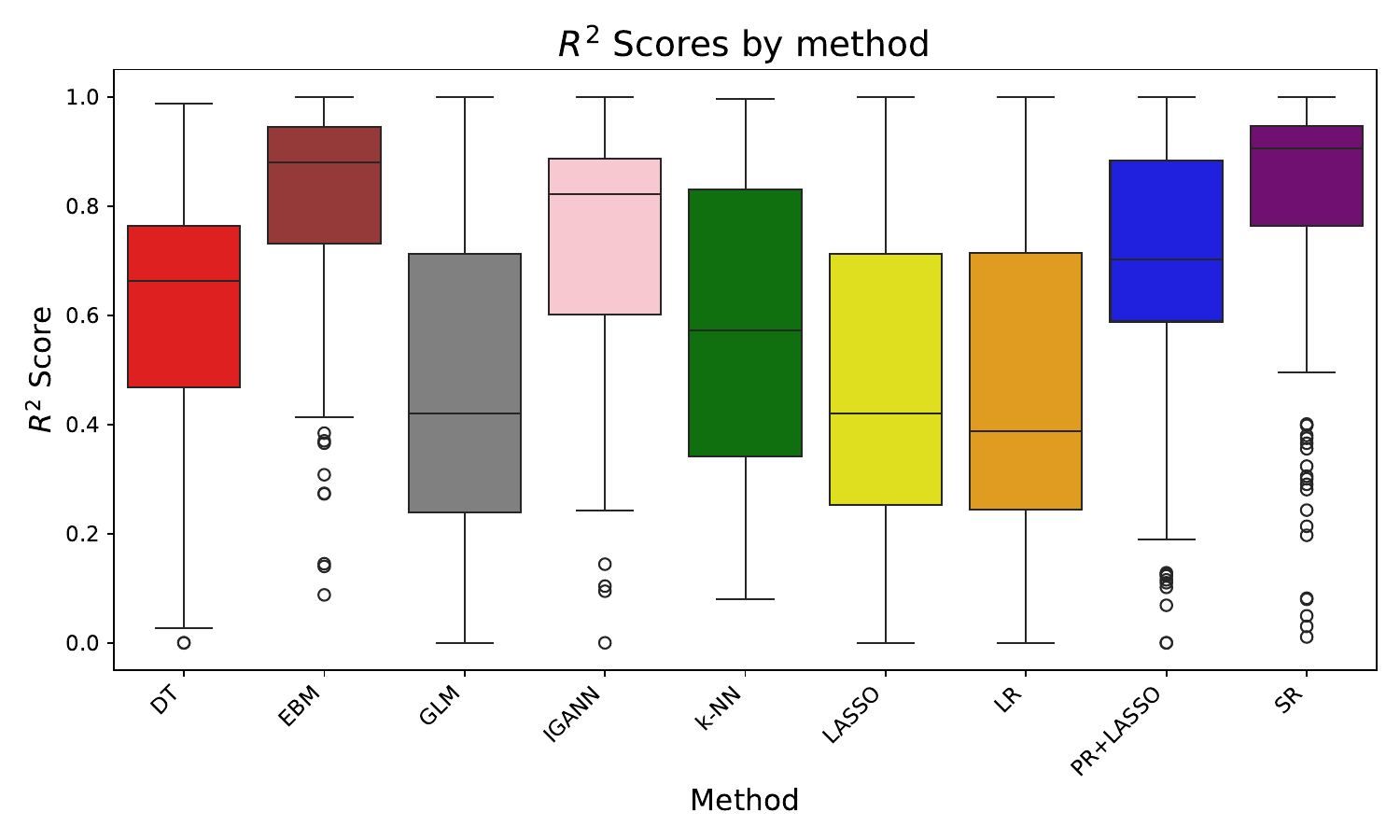}
        \caption{Regression
        }\label{fig:regr_is_box}
    \end{subfigure}
    \hfill
    \begin{subfigure}[b]{0.48\textwidth}
        \centering
        \includegraphics[width=\columnwidth]{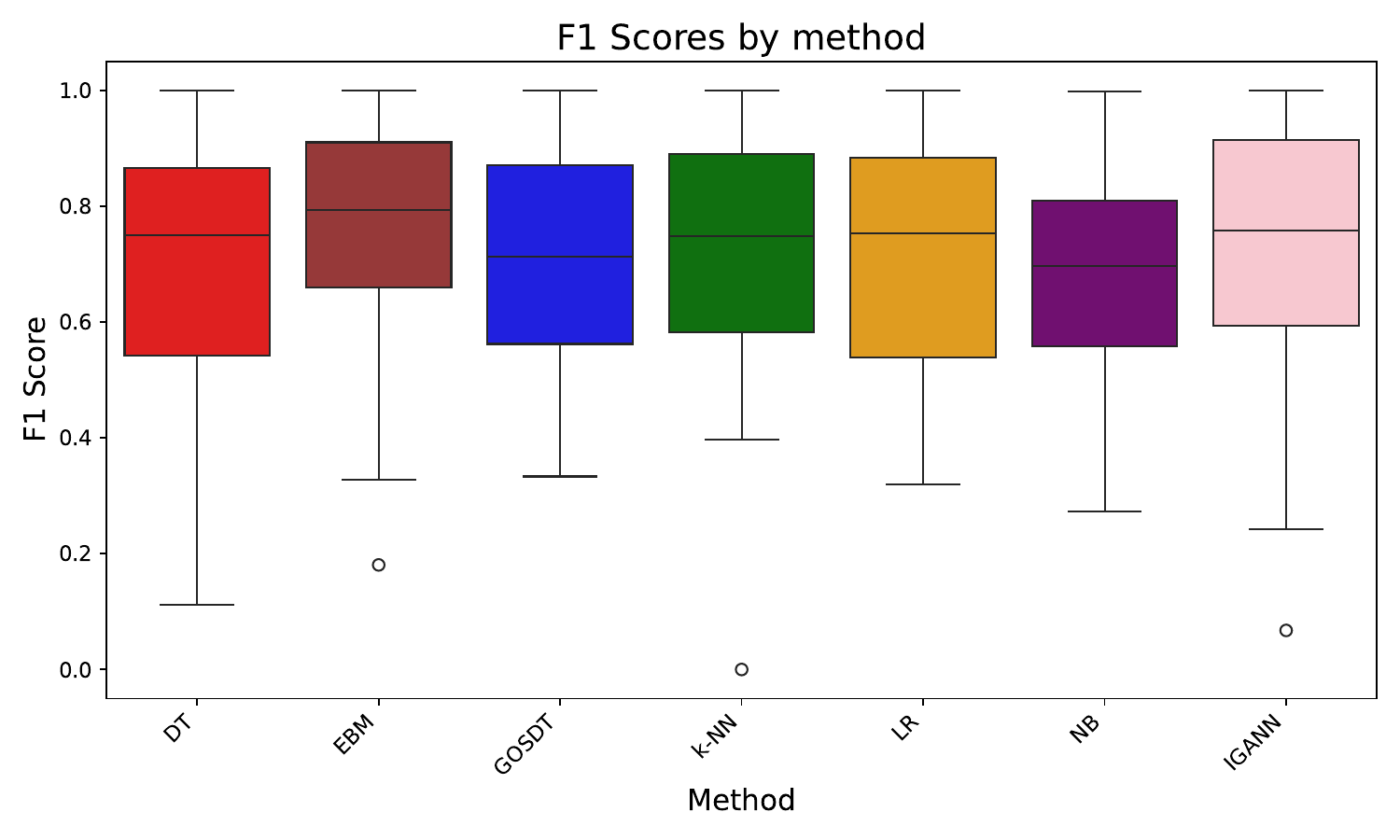}
        \caption{Classification
        }\label{fig:clf_is_box}
    \end{subfigure}
    \caption{Predictive accuracy distribution of the methods on all the datasets for (a) regression ($R^2$ scores) and (b) classification (F1 scores).}
    \label{fig:box}
\end{figure*}

Before comparing the specific methods, we assess the performance variability across the dataset collection to ensure that the benchmark provides a sufficiently diverse and challenging evaluation testbed.

By analyzing the median predictive performance ($R^2$ for regression, Figure \ref{fig:regr_is_prof}, and F1 score for classification, Figure \ref{fig:clf_is_prof}) on each dataset alongside the range between the best and worst-performing models, we observe a wide spectrum of dataset difficulties. For both tasks, the benchmark includes datasets where models achieve near-perfect scores (approaching 1) and others where models struggle significantly. Notably, the performance range in regression is quite wide. In classification, while the range is less pronounced, there remains a non-negligible gap between the best and worst models. This indicates that the benchmark is not saturated; the models have not converged on an intrinsic performance ceiling, ensuring that the datasets are suitably complex for the comparative analysis.

Moving to the performance of the individual methods (Figure \ref{fig:regr_is_box} for regression, and Figure \ref{fig:clf_is_box} for classification), the score distributions reveal distinct behaviors between the two tasks. In regression, we observe a considerable difference among the methods. EBM and Symbolic Regression achieve the highest overall performance, followed by IGANN and PR+LASSO.
k-NN exhibits high variability, followed by Decision Trees, while the linear models (Linear Regression, LASSO, and GLM) perform similarly at the lower end of the spectrum.

In classification, the distinction between models is much less clear. EBM maintains a slight performance advantage, but the remaining models exhibit highly overlapping performance distributions. The only notable exception is Naive Bayes, which performs poorly across the majority of the datasets.

\subsection{Method Rankings and Pairwise Comparisons}

\begin{figure*}
    \centering
    \begin{subfigure}[b]{0.48\textwidth}
        \centering
        \includegraphics[width=\textwidth]{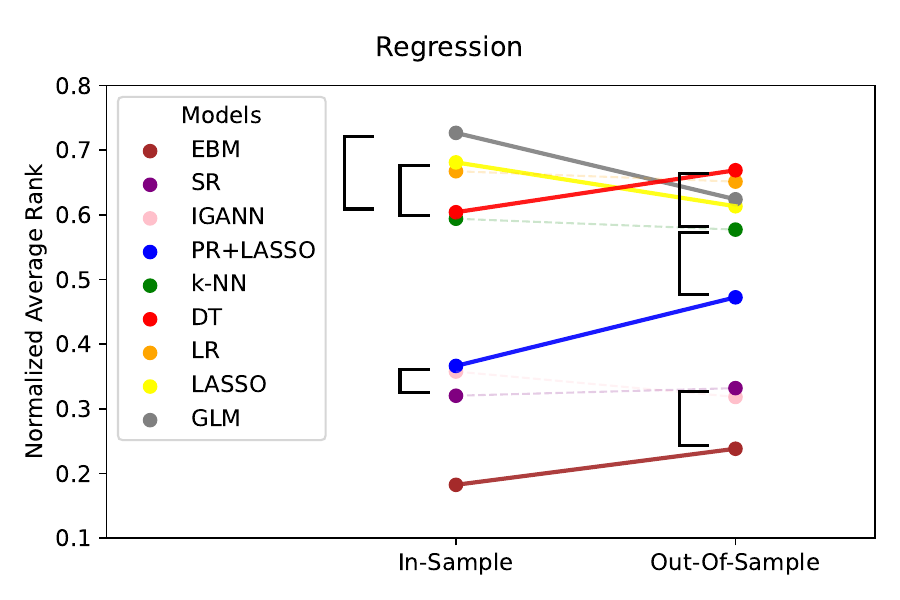}
        \caption{In-sample vs. Out-Of-Sample rankings (regression)}
        \label{fig:regr_is_os}
    \end{subfigure}
    \hfill
    \begin{subfigure}[b]{0.48\textwidth}
        \centering
        \includegraphics[width=\textwidth]{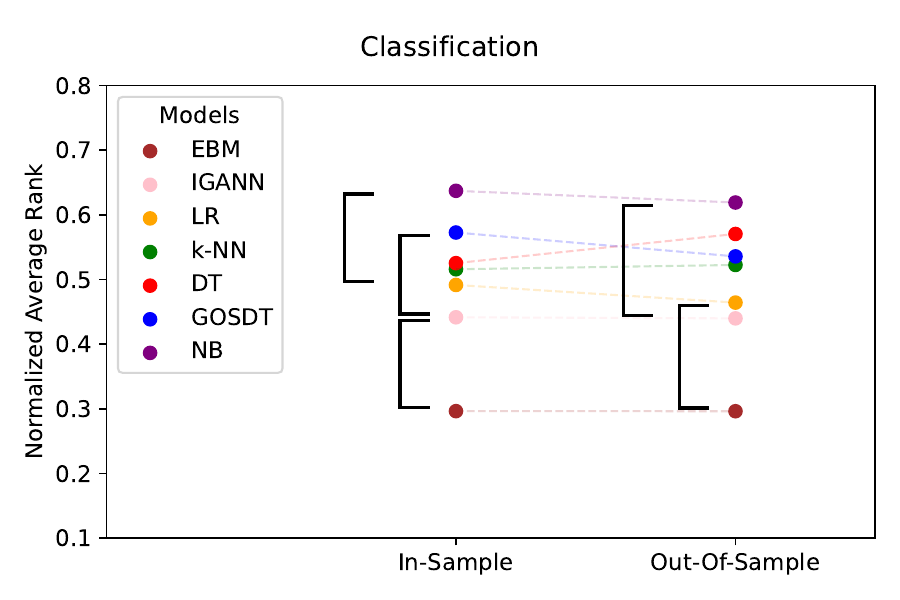}
        \caption{In-sample vs. Out-Of-Sample rankings (classification)}
        \label{fig:clf_is_os}
    \end{subfigure}
    \caption{Ranking comparison for (a) the regression and (b) the classification tasks between in-sample and out-of-sample settings.}
    \label{fig:is_vs_os}
\end{figure*}

\begin{figure*}
    \centering
    \begin{subfigure}[b]{0.48\textwidth}
        \centering
        \includegraphics[width=\textwidth]{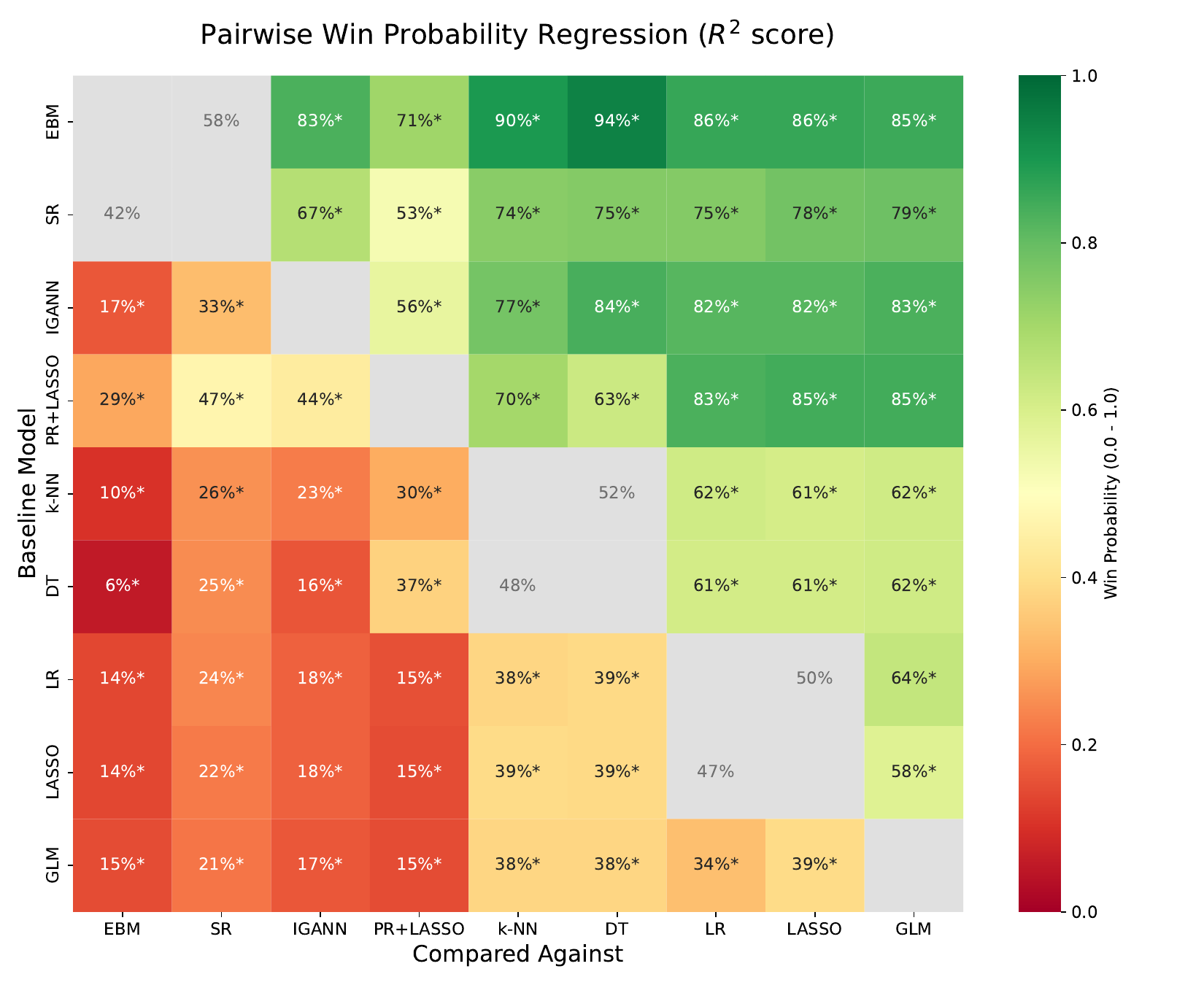}
        \caption{Regression}
        \label{fig:regr_pair}
    \end{subfigure} 
    \hfill
    \begin{subfigure}[b]{0.48\textwidth}
        \centering
        \includegraphics[width=\textwidth]{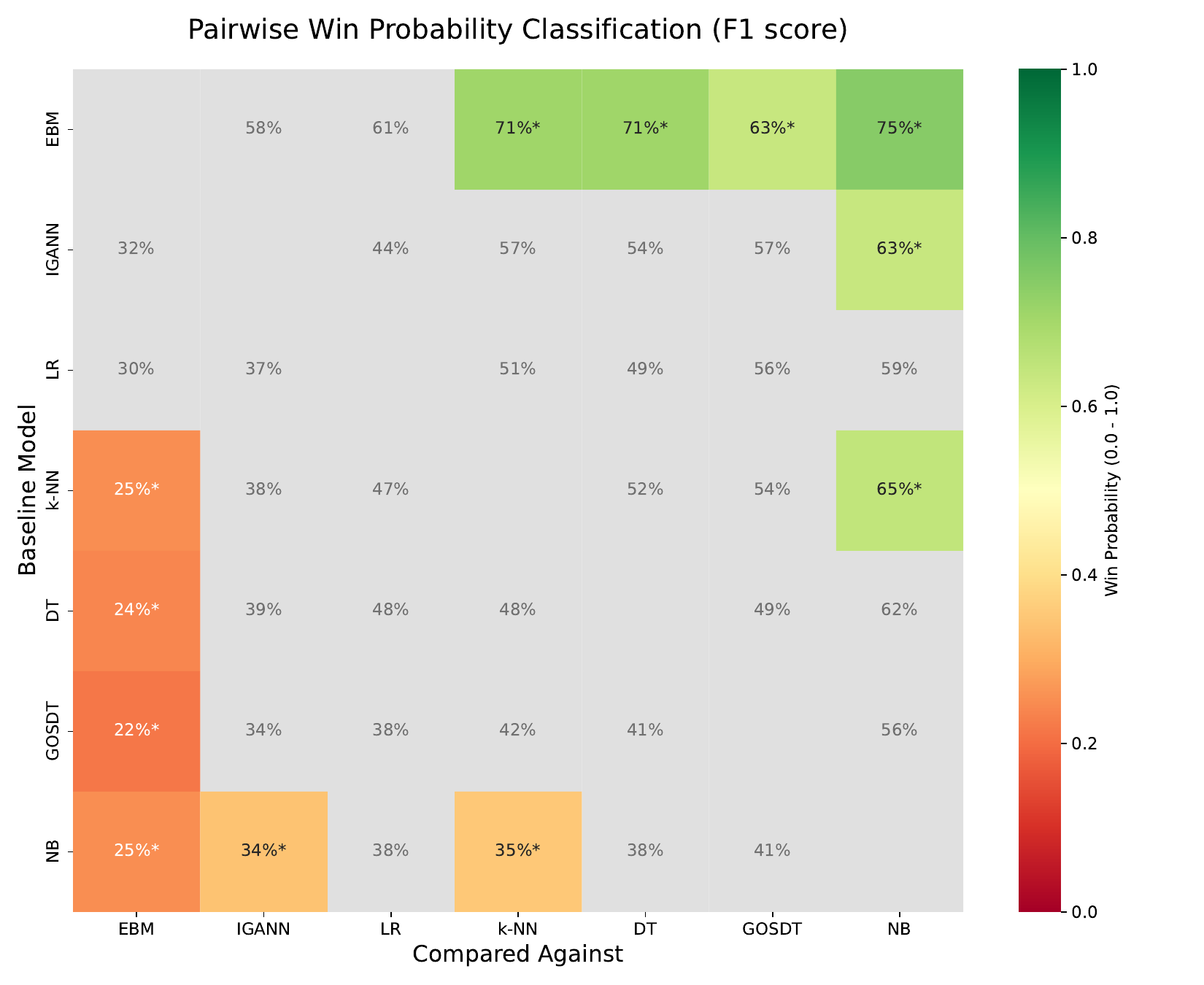}
        \caption{Classification}
        \label{fig:clf_pair}
    \end{subfigure}
    \caption{Pairwise comparison between methods in the in-sample setting for (a) regression and (b) classification.}
    \label{fig:aggr_pair}
\end{figure*}

To rigorously establish a performance hierarchy, we evaluated the statistical significance of the model rankings. Specifically, we utilized the Friedman test, a non-parametric alternative to repeated-measures ANOVA used to detect overall differences across multiple datasets, paired with the Nemenyi post-hoc test, which identifies specific pairwise differences by comparing mean ranks. This approach follows established best practices for comparing machine learning algorithms \citep{demvsar2006statistical}. In our critical difference visualizations, brackets connect methods whose performances are not statistically distinguishable.
Furthermore, to compare the in-sample and out-of-sample settings, we applied the Wilcoxon Signed-Rank Test, a non-parametric method used to assess whether the median difference between paired observations is statistically significant \citep{demvsar2006statistical}. To control the family-wise error rate across repeated measures, we incorporated a Holm correction. Significant trends in these ranks are denoted by solid lines.

For regression in the in-sample setting (Figure \ref{fig:regr_is_os}), three distinct groups emerge. EBM isolates itself as the top-performing method. The second tier comprises SR, IGANN, and PR+LASSO, while the final tier includes the remaining algorithms. In the out-of-sample setting, the overall hierarchy remains largely intact, though PR+LASSO suffers a relative decrease in rank, together with the decision trees. EBM, SR, and IGANN consistently remain the top performers, and regularized linear models like LASSO tend to be more robust in this setting. Conversely, the classification rankings, in Figure \ref{fig:clf_is_os}, are highly overlapping, making it difficult to distinguish clear, statistically significant tiers among the majority of the algorithms.

This trend is further corroborated by our analysis of the pairwise win probabilities. We calculated the probability of one method outperforming another, employing the Wilcoxon signed-rank test with a Holm correction to identify statistically significant pairs. Crucially, unlike previous rank-based evaluations, this test assesses raw performance values; thus, marginal differences may lack statistical significance even if one method consistently outranks another. In the regression task (Figure \ref{fig:regr_pair}), the performance hierarchy is distinct, with differences being almost universally significant. However, there are two notable exceptions where no statistically significant difference is observed: between EBM and SR, as well as between LR and LASSO. These exceptions are particularly relevant as they demonstrate that (i) SR, despite being constrained to provide analytical and highly interpretable models, yields highly competitive results, and (ii) the application of LASSO regularization does not compromise predictive performance relative to standard linear regression.

In classification (Figure \ref{fig:clf_pair}), the situation is drastically different. Aside from EBM's dominance and Naive Bayes' underperformance, a definitive hierarchy among the remaining methods cannot be established, corroborating the ranking analysis.

These findings highlight an intrinsic difference between the regression and classification tasks. While the regression analysis reveals distinct behavioral differences among the methods, the classification results exhibit a less pronounced performance hierarchy. In the subsequent sections, we deepen our analysis by characterizing the datasets using the meta-features introduced in Section \ref{sec:ds_metrics}. Guided by our initial findings, the regression analysis will focus on the relationship between these meta-features and model performance. Conversely, the classification analysis will investigate whether meta-features can establish a clearer hierarchy, or if underlying patterns and method clusters exist that were missed in the aggregated analysis.

\subsection{Meta-Features Effects on Regressors}\label{sec:regr_strat}

\begin{figure*}
    \centering
    \begin{subfigure}[b]{0.48\textwidth}
        \centering
        \includegraphics[width=\columnwidth]{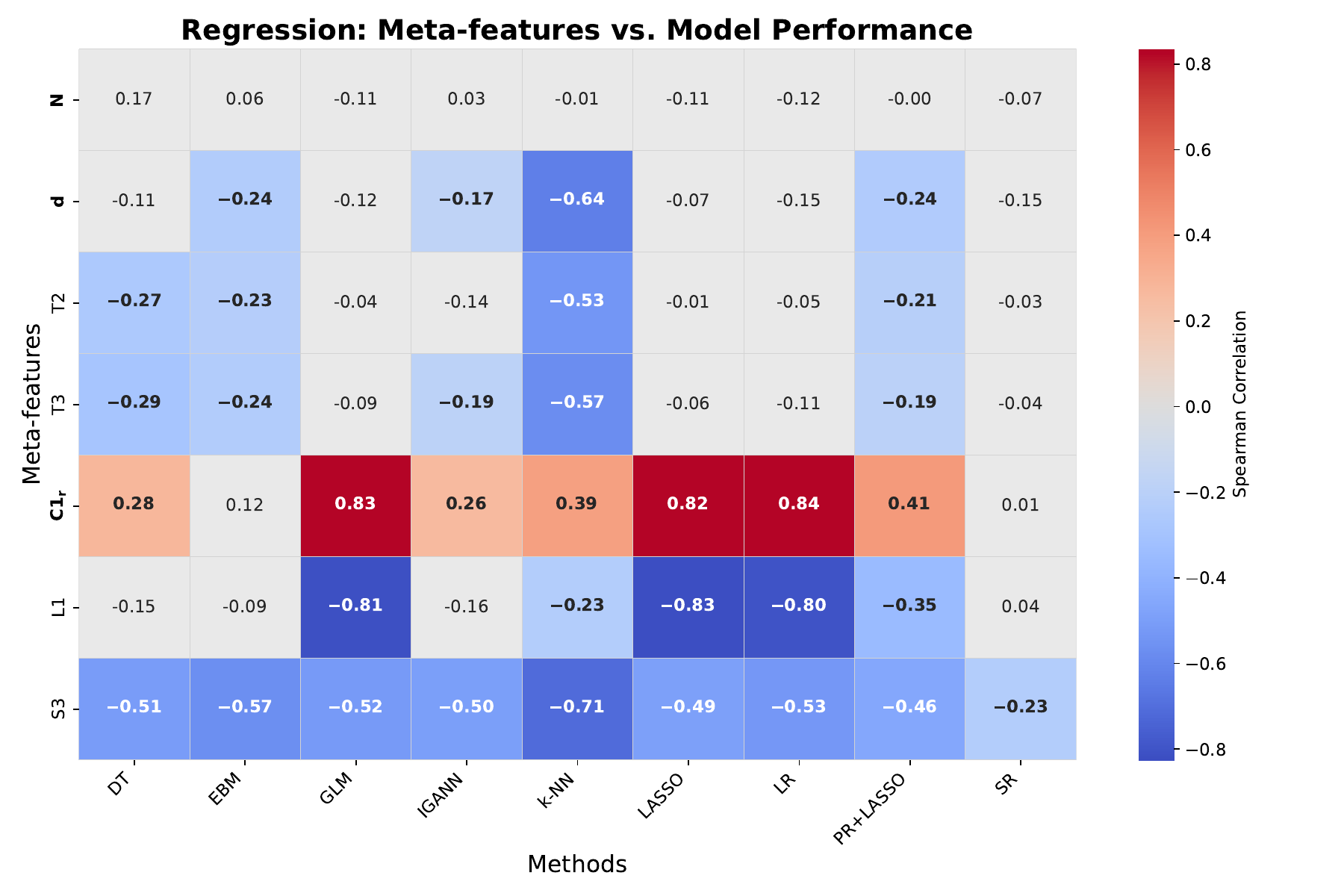}
        \caption{Regression}\label{fig:regr_is_corr}
    \end{subfigure}
    \hfill
    \begin{subfigure}[b]{0.48\textwidth}
        \centering
        \includegraphics[width=\columnwidth]{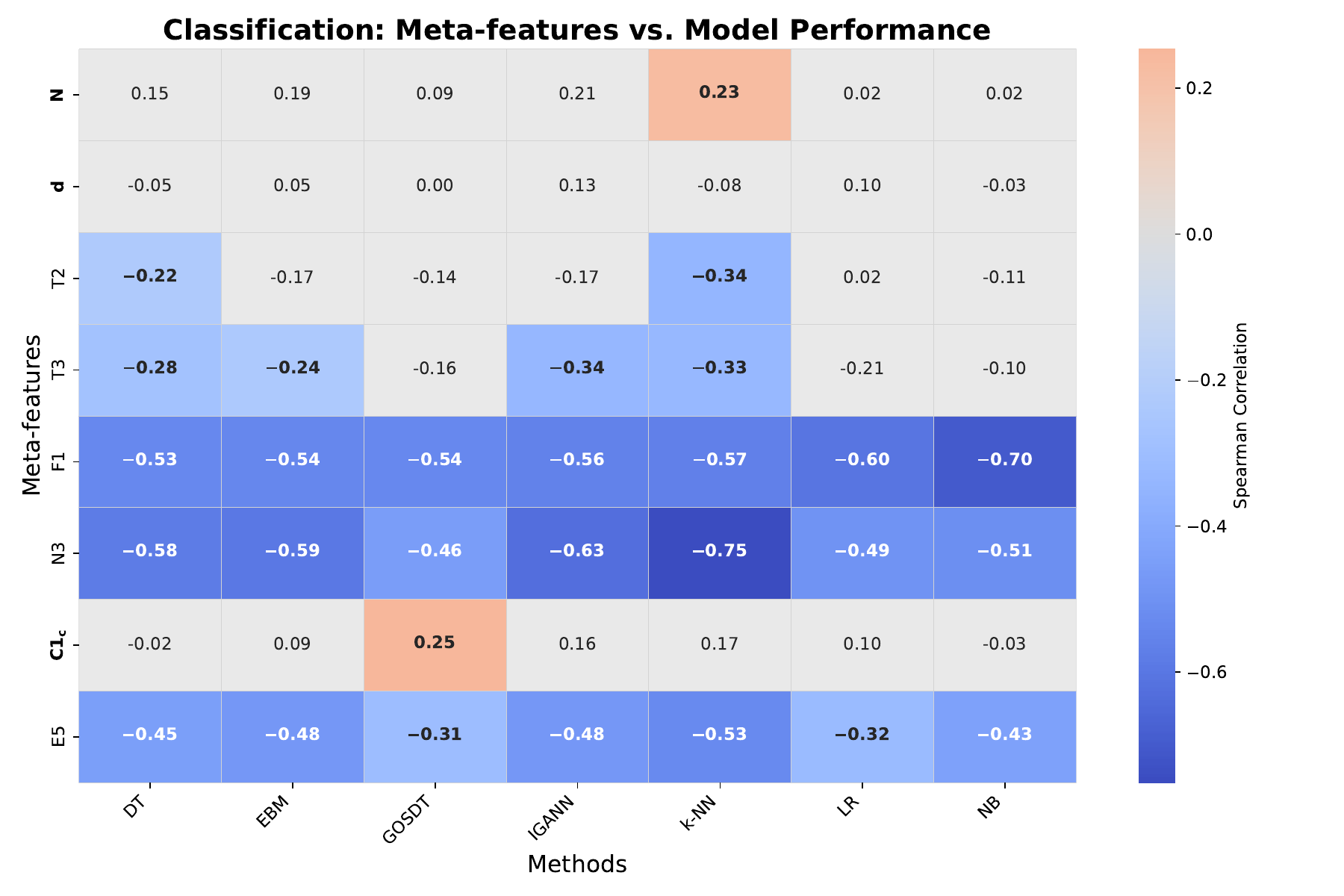}
        \caption{Classification}\label{fig:clf_is_corr}
    \end{subfigure}
    \caption{Spearman correlation between dataset meta-features and (a) regressor or (b) classifier performance.}
    \label{fig:meta_corr}
\end{figure*}

\begin{figure*}
\centering
\includegraphics[width=\textwidth]{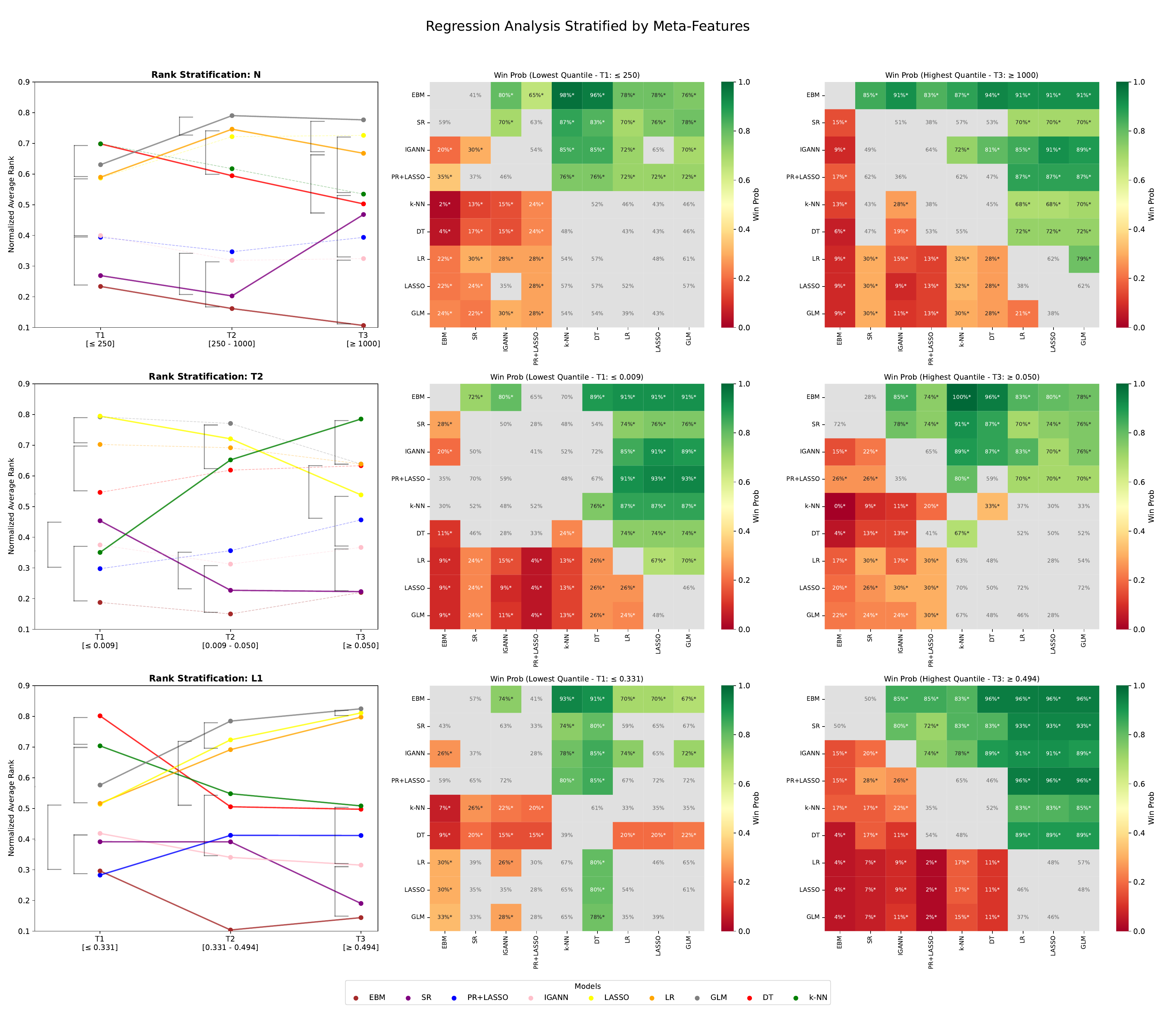}
\caption{Stratified results for the regression task (in-sample). 
}\label{fig:regr_is_strat}
\end{figure*}

\begin{figure*}
\centering
\includegraphics[width=\textwidth]{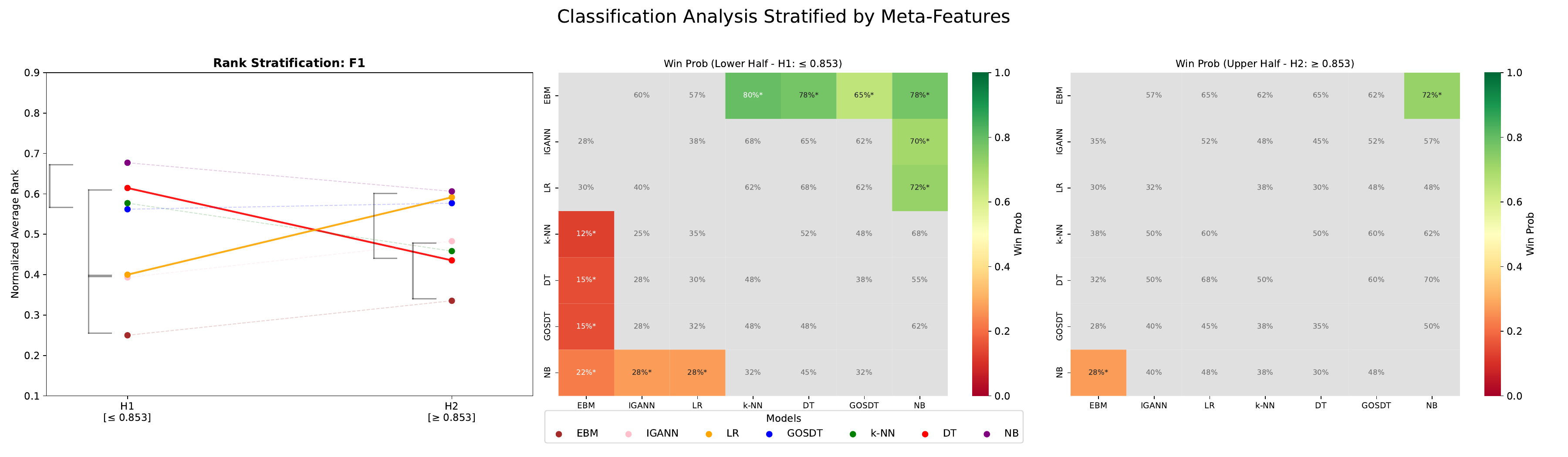}
\caption{Stratified results for the classification task (in-sample). 
}\label{fig:clf_is_strat}
\end{figure*}

To understand the drivers of model performance, we analyze the Spearman correlation between the dataset meta-features and the models' predictive scores (Figure \ref{fig:regr_is_corr}). Only the significant values are colored.

Several strong correlations emerge:
\begin{itemize}
    \item Target Correlation ($\text{C1}_\text{r}$): Strong positive correlation with performance, especially for linear-based models.
    \item Non-linearity (L1): Negatively impacts linear models, while SR, GAMs, and decision trees demonstrate robustness and model non-linearity effectively.
    \item Smoothness/Sampling Quality (S3): Negatively affects all methods.
    \item Dimensionality and Sparsity: Negatively impacts EBM, DT, IGANN, PR+LASSO, and severely impacts k-NN. Linear models and SR remain relatively robust.
\end{itemize}



To capture more nuanced behaviors, we stratified the regression datasets into terciles based on their meta-features. Figure \ref{fig:regr_is_strat} presents a selection of these results, with the remainder detailed in Appendix \ref{sec:compl_strat}. The rankings within these strata reveal behaviors largely consistent with theoretical expectations. For instance, k-NN is heavily penalized by data sparsity, whereas LASSO benefits from it. Furthermore, as expected, linear models underperform in non-linear scenarios (L1), where SR conversely excels. At the same time, non-parametric models, such as decision trees and k-NN, underperform in highly linear settings. 

Interestingly, while SR performs exceptionally well in low-sample settings, we observe a surprising relative degradation in performance as the sample size increases. We rule out training time constraints as the primary cause for this trend, as confirmed by an ablation study detailed in Appendix \ref{sec:ablation_sr}. Instead, we note that the resulting equations remain remarkably small even with larger sample sizes. This suggests that the relative underperformance is likely driven by the strict parsimony of the model selection criterion, which may overly penalize complexity and prevent SR from capturing finer patterns in large datasets. Conversely, SR demonstrates unexpected robustness in high-dimensional settings. This resilience to a large number of features is non-trivial; it demonstrates that the initial feature selection step utilized by PySR effectively mitigates the combinatorial explosion that has traditionally been considered one of the most severe limitations of this approach.

Furthermore, when a feature is highly correlated with the target variable, all methods tend to perform similarly. Finally, target noise (S3) does not appear to differentiate model performance, aside from an expected detrimental effect on k-NN, suggesting that noise degrades the performance of all methods relatively uniformly.

\subsection{Investigating the Classification Performance Landscape}

\begin{figure}
\centering
\includegraphics[width= 0.6\columnwidth]{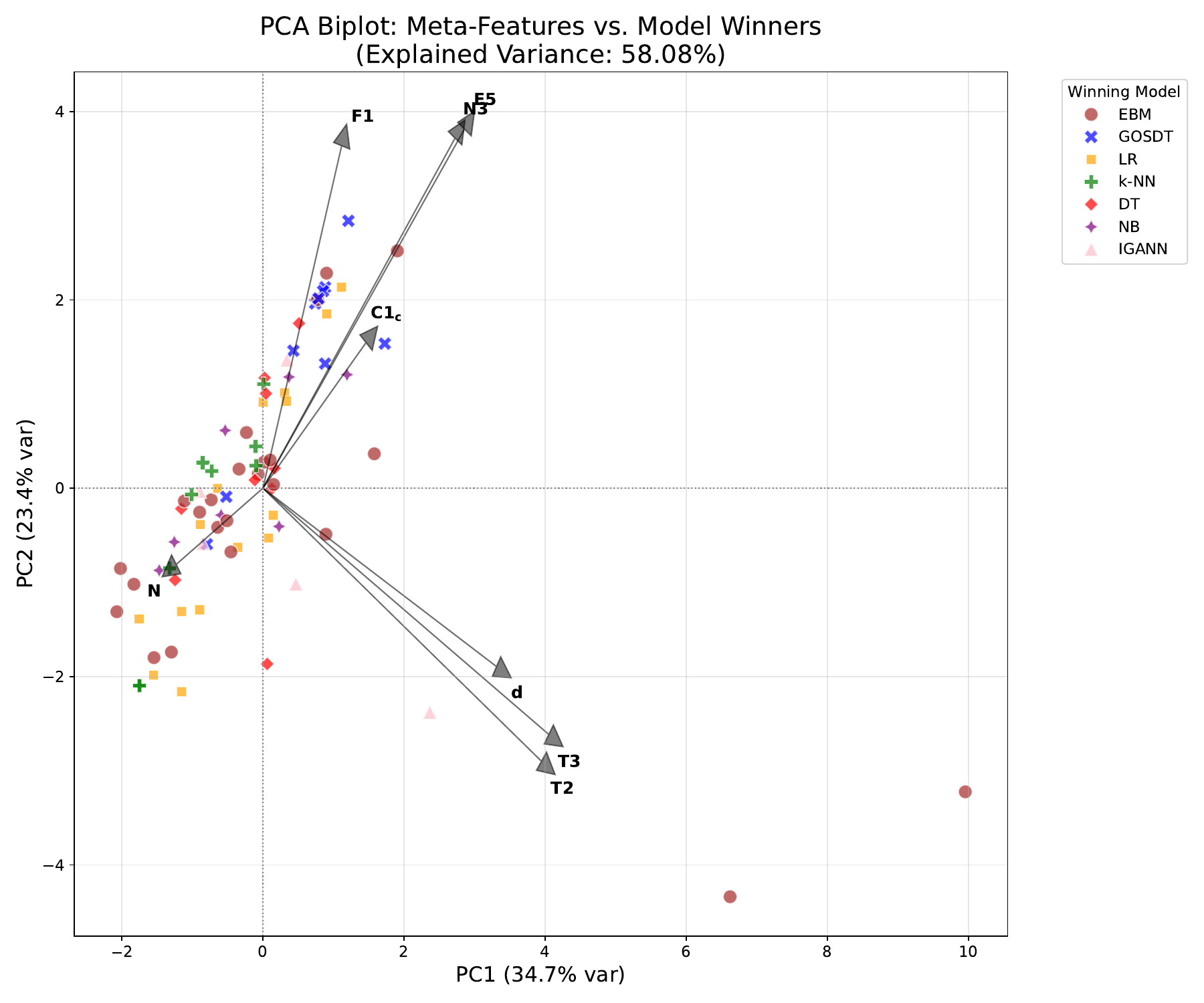}
\caption{PCA biplot of the dataset meta-features projected onto the first two principal components. Individual datasets are color-coded by the winning algorithm.}\label{fig:clf_is_pca}
\end{figure}

\begin{figure}
\centering
\includegraphics[width=0.6\columnwidth]{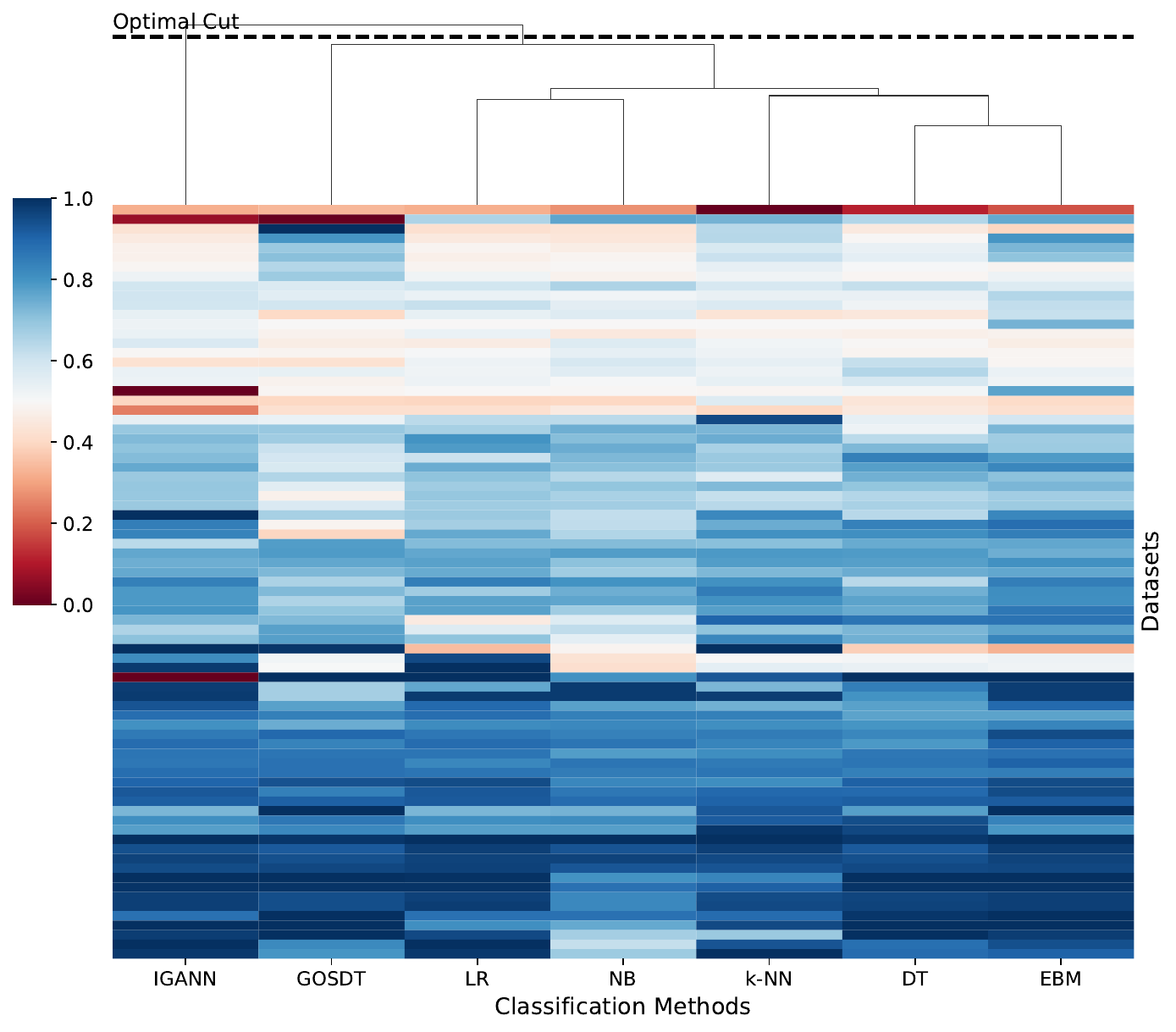}
\caption{Hierarchical clustering of method performance profiles across classification datasets.}\label{fig:clf_is_cluster}
\end{figure}

In Figure \ref{fig:clf_is_corr}, we observe that classifier performance moderately correlates with certain meta-features, particularly those related to dimensionality and noise, broadly consistent with patterns seen in regression. An additional effect is observed for GOSDT, which exhibits a distinct sensitivity to class imbalance.

Unlike regression, however, stratifying the classification datasets fails to produce any consistent or interpretable performance hierarchy. Figure \ref{fig:clf_is_strat} illustrates an example of this meta-feature-induced stratification. To nearly match the sample sizes used in the regression analysis, we divide the datasets into halves rather than terciles. The only distinct pattern emerges when a highly discriminative feature (the F1 meta-feature) is present: under these conditions, Logistic Regression improves its relative standing, whereas traditional Decision Trees experience a relative decline in performance. Stratifications based on the remaining meta-features (detailed in Appendix \ref{sec:compl_strat}) do not reveal any significant patterns.

Given the absence of a clear hierarchy, even after stratification, we investigate whether underlying structures or clusters exist that our previous aggregated analyses might have missed.

First, we project the meta-features into a reduced space using Principal Component Analysis (PCA). The resulting biplot (Figure \ref{fig:clf_is_pca}), which captures 58\% of the variance, plots the datasets on the first two principal components, colored by the winning algorithm. This visualization reveals no clear clustering of winning models in the meta-feature space, indicating that even linear combinations of these meta-features fail to meaningfully separate the winning models, indicating that the meta-feature space does not encode a usable structure for model selection.

Next, to determine if the performance profiles themselves naturally cluster, we applied hierarchical clustering (using Euclidean distance and Ward's method) directly to the methods' performances across all classification datasets. This yielded a poor overall silhouette score of 0.22 (Figure \ref{fig:clf_is_cluster}). When we forced the algorithm to find a number of clusters equal to the number of methods, the silhouette score decreased further to 0.05. For reference, the silhouette score computed using this same methodology for the regression tasks was 0.54 and 0.30, respectively. These metrics confirm that distinct performance clusters do not exist in this classification benchmark; rather, model successes and failures are highly interleaved across datasets and do not align with the selected meta-features. 

To ensure this behavior was not merely an artifact of the discrete nature of the F1 score, we confirm that this lack of hierarchy persists even when evaluating the models using continuous proper scoring rules, such as the Brier score. We include these results in Appendix \ref{sec:clf_compl_res}.

Overall, we do not find any evidence of a stable or recoverable performance hierarchy in classification. Taken together, these results indicate that classifier performance cannot be organized into a low-dimensional or meta-feature-driven structure, and that current meta-features are insufficient for reliable characterization.

\subsection{Operationalizing Algorithm Selection and Dataset Difficulty}
\label{sec:operationalize}
\begin{figure*}
    \centering
    \begin{subfigure}[b]{0.48\textwidth}
        \centering
        \includegraphics[width=\columnwidth]{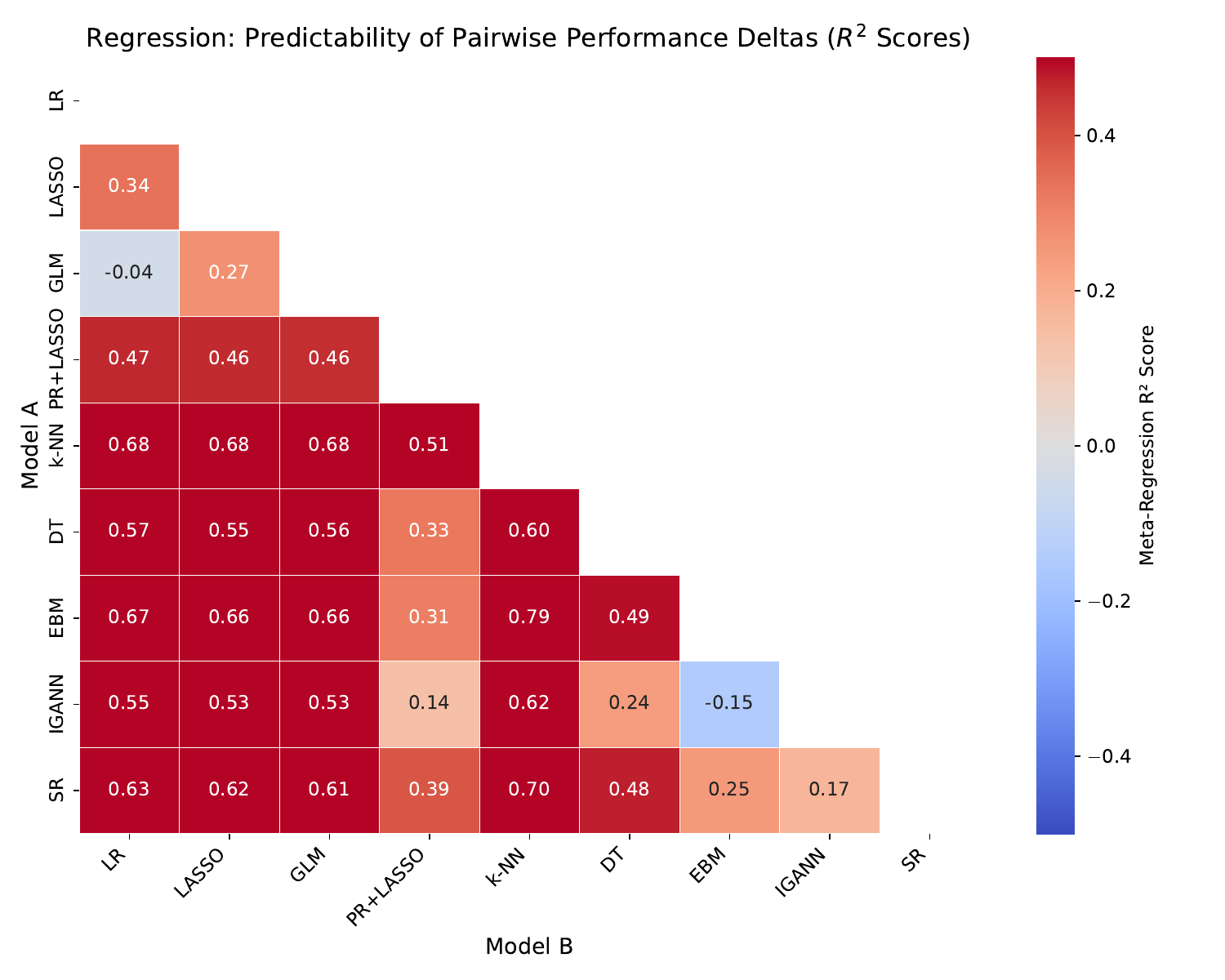}
        \caption{Regression (in-sample)}\label{fig:regr_is_metap}
    \end{subfigure}
    \hfill
    \begin{subfigure}[b]{0.48\textwidth}
        \centering
        \includegraphics[width=\columnwidth]{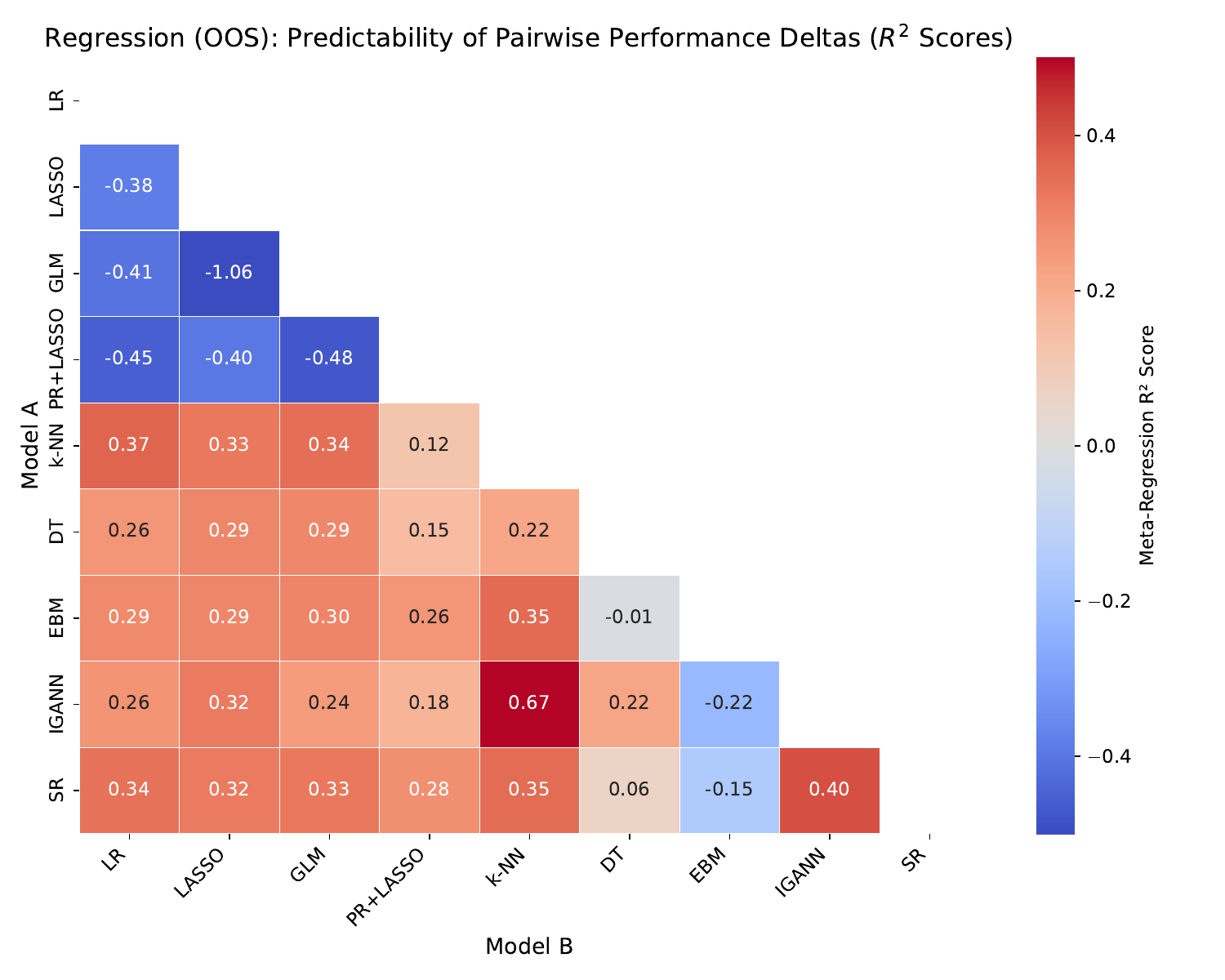}
        \caption{Regression (out-of-sample)}\label{fig:regr_oos_metap}
    \end{subfigure}
    \caption{$R^2$ scores of the meta-models predicting the $\Delta$ performance between pairs of models in regression (a) in-sample and (b) out-of-sample.}
    \label{fig:meta_p}
\end{figure*}

\begin{figure*}
    \centering
    \begin{subfigure}[b]{0.48\textwidth}
        \centering
        \includegraphics[width=\columnwidth]{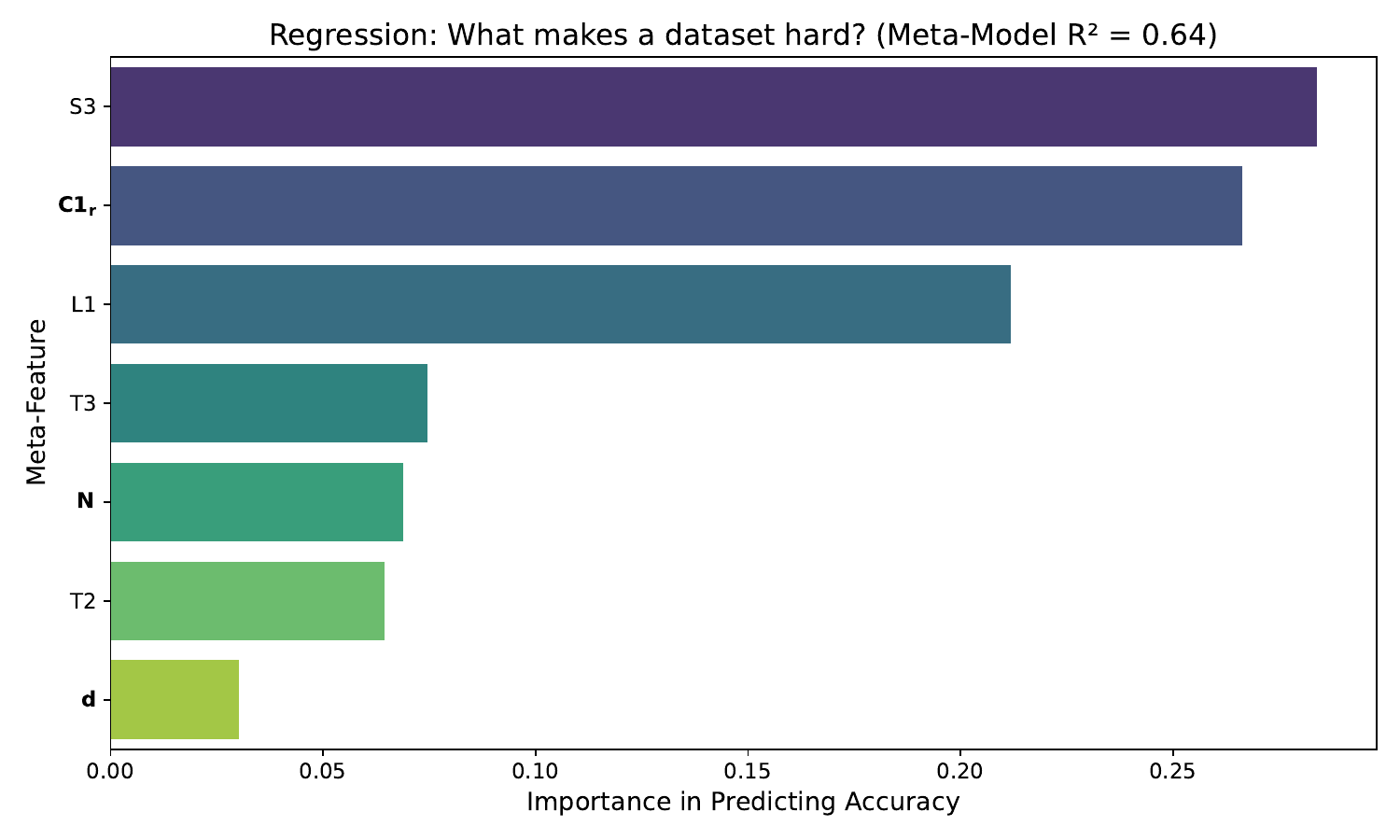}
        \caption{Regression (in-sample)}\label{fig:regr_is_metaf}
    \end{subfigure}
    \hfill
    \begin{subfigure}[b]{0.48\textwidth}
        \centering
        \includegraphics[width=\columnwidth]{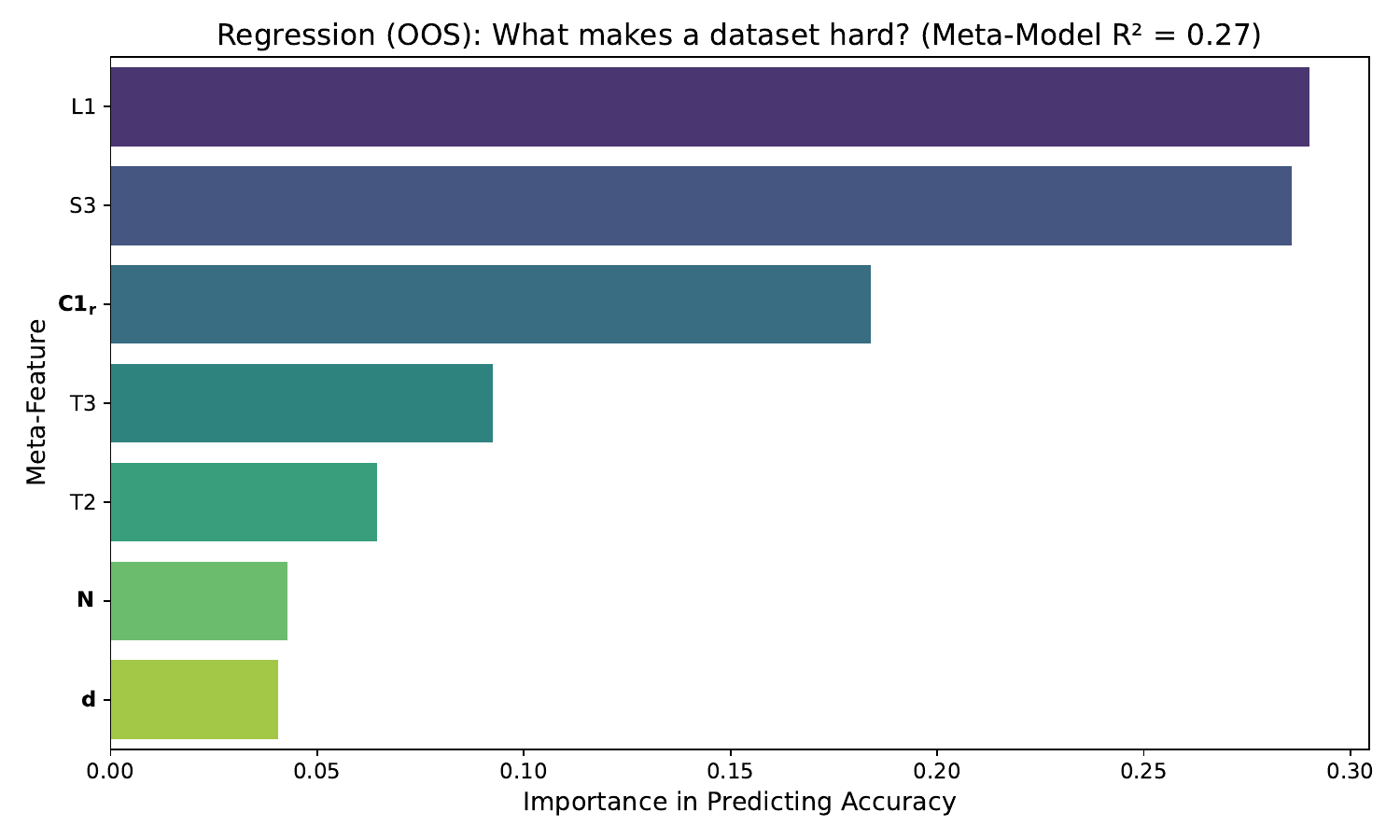}
        \caption{Regression (out-of-sample)}\label{fig:regr_oos_metaf}
    \end{subfigure}
    \caption{Random Forest feature importances for predicting dataset difficulty in regression (a) in-sample and (b) out-of-sample.}
    \label{fig:meta_f}
    
\end{figure*}

Given the clear contrast between the structured behavior of regression models and the highly interleaved nature of classification performance, a natural next step is to test whether we can leverage these insights predictively. To this aim, we evaluate whether meta-features can be operationalized for algorithm selection and dataset difficulty estimation.

To answer whether we can operationalize algorithm selection using meta-features, we trained a Random Forest meta-model (evaluated via 5-fold cross-validation) to predict the pairwise difference in performance between methods $i$ and $j$, defined as:
\begin{equation}
\Delta_{i,j} = S_i - S_j
\end{equation}
where $S$ represents the $R^2$ or F1 score on a dataset.

In regression (Figure \ref{fig:regr_is_metap}), the meta-model achieves consistently positive $R^2$ values, effectively capturing differences in inductive biases. The only exceptions are pairs of highly similar methods, such as EBM versus IGANN or closely related linear models. When evaluating out-of-sample behavior (Figure \ref{fig:regr_oos_metap}), predictive performance decreases, indicating that meta-features are less reliable for anticipating generalization. While the overall structure remains similar to the in-sample setting, the discrepancies are amplified. In particular, the relative performance between certain model families, such as linear and additive methods, becomes difficult to predict.

We further explored what makes a dataset inherently ``hard'' by training a meta-model to predict the average performance across methods on a given dataset. For regression in-sample (Figure \ref{fig:regr_is_metaf}), the meta-model achieves strong performance ($R^2 = 0.64$), with S3, $\text{C1}_\text{r}$, and L1 as the dominant features, while dimensionality-related features play a minor role. If we look at the feature importance for out-of-sample regression (Figure \ref{fig:regr_oos_metaf}), we still notice that meta-features related to dimensionality are less important. The only difference is that the linearity of the problem seems to have a larger impact on the difficulty of the dataset.

In contrast, and as expected, the same analyses on classification tasks show that the meta-model rarely achieves positive $R^2$ values, indicating that meta-features do not provide reliable guidance for algorithm selection. Similarly, dataset difficulty is only weakly captured ($R^2 = 0.28$), suggesting that classification performance cannot be effectively characterized using the available meta-features. Detailed results and figures are provided in Appendix \ref{sec:clf_compl_res}.

\subsection{Structural Interpretability and Model Complexity}\label{sec:model_compl_res}
\begin{figure*}
    \centering
    \begin{subfigure}[b]{0.48\textwidth}
        \centering
        \includegraphics[width=\columnwidth]{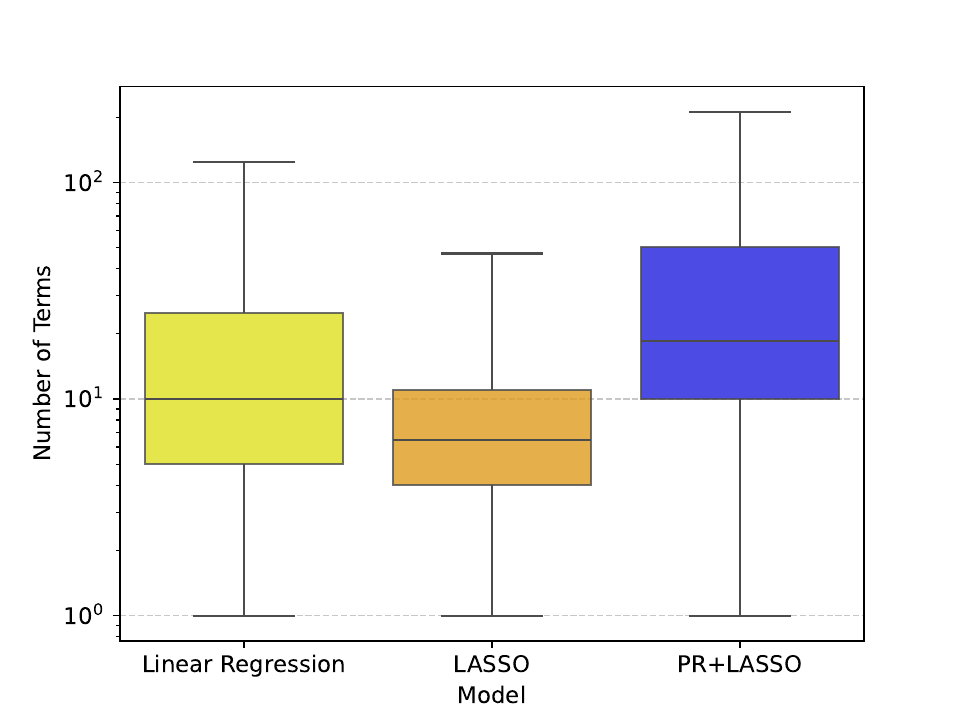}
        \caption{Linear Models}\label{fig:lin_lasso}
    \end{subfigure}
    \hfill
    \begin{subfigure}[b]{0.48\textwidth}
        \centering
        \includegraphics[width=\columnwidth]{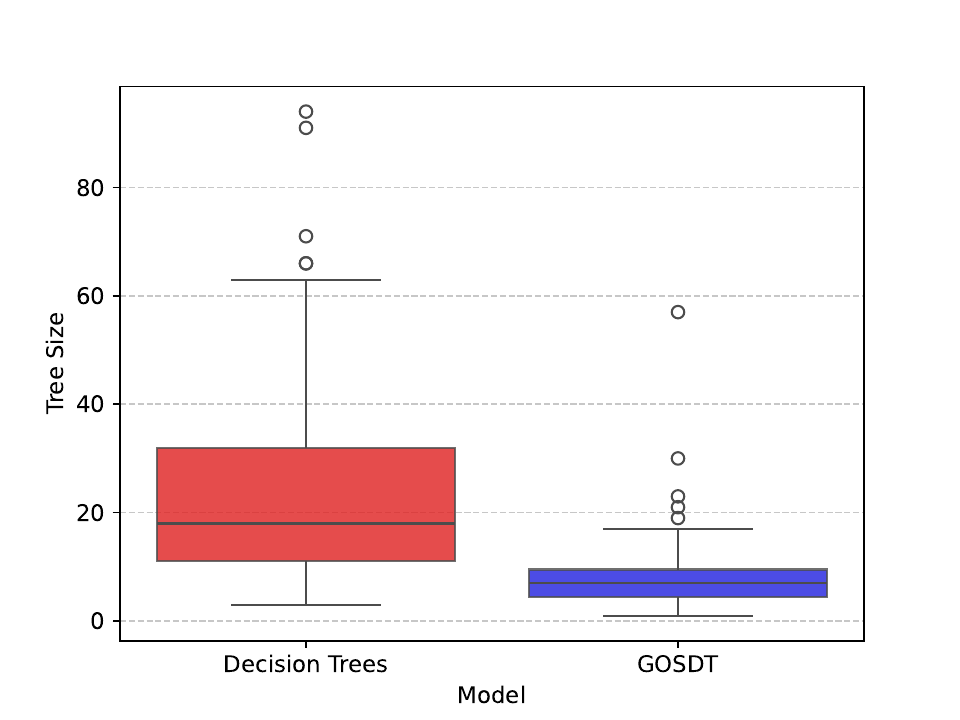}
        \caption{Decision Tree Models}\label{fig:gosdt_dt}
    \end{subfigure}
    \caption{Comparison of model complexity measured by (a) the number of non-zero linear terms, between Linear Regression, LASSO, and Polynomial Regression with LASSO; and by (b) tree size, between traditional Decision Trees and GOSDT.}
    \label{fig:compl}
\end{figure*}

Having established the predictive performance landscape and the role of dataset meta-features, we now shift our focus to the operational properties of the models. In IML, predictive accuracy is only one facet of utility; the structural complexity of the resulting model is equally critical.

Quantifying interpretability is inherently architecture-specific, precluding a direct, one-to-one comparison across heterogeneous models (e.g., k-NN versus Symbolic Regression). However, by examining models within the same structural families, we can objectively assess their relative complexities.

Within the family of linear models, interpretability can be mathematically approximated by analyzing the sparsity of the resulting equations. To quantify this, we apply a numerical threshold, treating any coefficient with an absolute value below $10^{-6}$  as zero. Figure \ref{fig:lin_lasso} illustrates the distribution of the remaining non-zero coefficients. The results demonstrate that LASSO effectively minimizes model complexity compared to standard Linear Regression. Conversely, PR+LASSO exhibits a marked increase in the number of terms due to the polynomial expansion of the feature space. Beyond the sheer number of terms, the inclusion of higher-order transformations and interaction effects makes Polynomial Regression inherently less transparent than purely linear models. Nevertheless, this added structural complexity allows the model to capture non-linear relationships that strict linear formulations cannot fundamentally represent. PR+LASSO thus exemplifies a clear performance-interpretability trade-off: increased expressive power can yield improved predictive accuracy but generates elaborate symbolic representations that require significantly greater effort to interpret.

A similar structural comparison can be made within tree-based models. Traditional Decision Trees and GOSDT exhibit comparable predictive performance, with a slight advantage favoring the former. Because both methods yield identical model structures (decision trees), their interpretability can be directly compared via the number of resulting nodes. Figure \ref{fig:gosdt_dt} plots the distribution of node counts across all datasets. As anticipated, GOSDT produces significantly sparser trees, delivering more compact and therefore more easily interpretable models.

\subsection{Computational Efficiency and Training Time}

\begin{figure*}
    \centering
    \begin{subfigure}[b]{0.48\textwidth}
        \centering
        \includegraphics[width=\columnwidth]{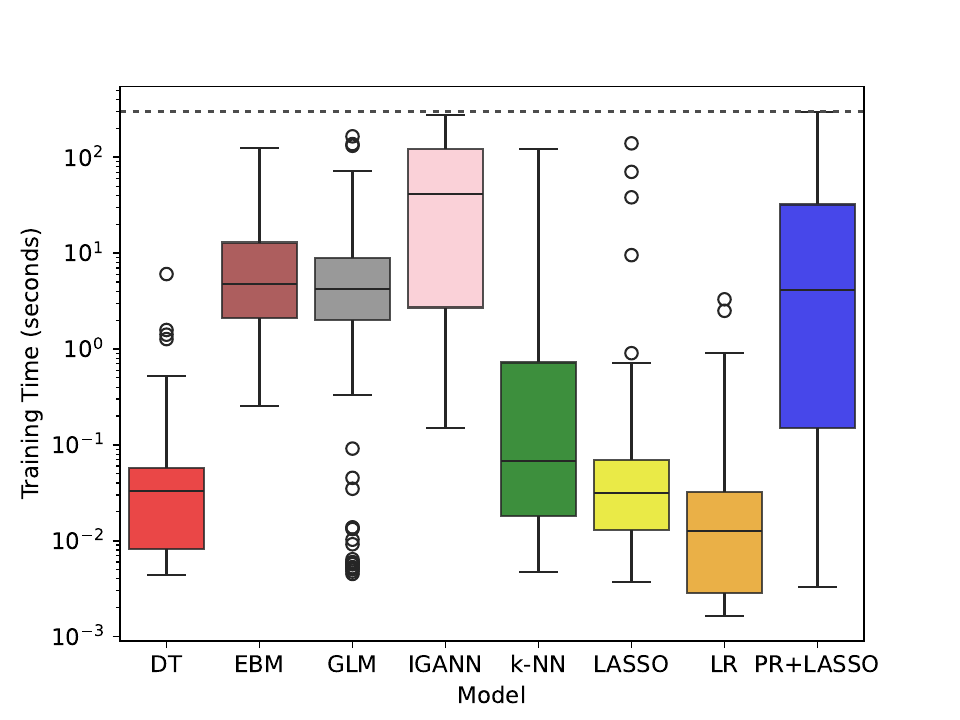}
        \caption{Regression Training Time}\label{fig:training_time_regr}
    \end{subfigure}
    \hfill
    \begin{subfigure}[b]{0.48\textwidth}
        \centering
        \includegraphics[width=\columnwidth]{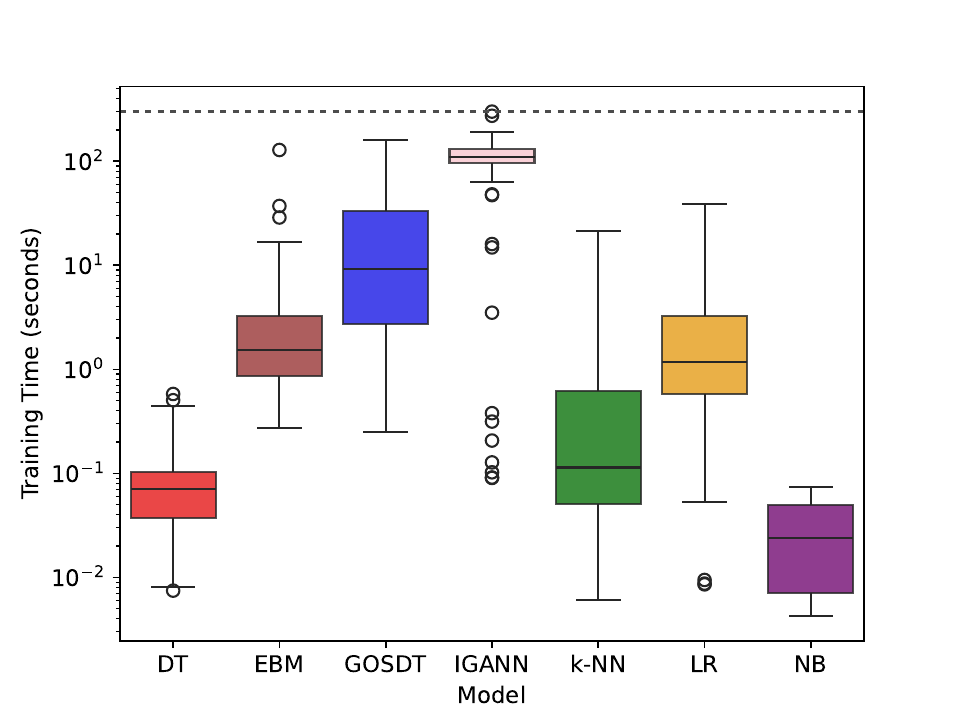}
        \caption{Classification Training Time}\label{fig:training_time_clf}
    \end{subfigure}
    \caption{Training times, measured in seconds, for (a) the regression models and (b) the classification models. The horizontal dashed line represents the maximum allowed time for each method (300 seconds).}
    \label{fig:time}
\end{figure*}

Beyond structural complexity, training time remains a critical operational constraint, particularly in environments requiring rapid retraining or responsive model updates.

Figure \ref{fig:training_time_regr} details the training times for the regression models. A high variance in execution time is evident across the methods, underscoring substantial differences in computational efficiency among interpretable models. IGANN requires the longest execution times, likely due to its reliance on neural networks to compute shape functions, a process that would heavily benefit from GPU acceleration. PR+LASSO is the next slowest, primarily burdened by the expanded feature space created during polynomial transformation. Interestingly, GLM demonstrates relatively long training times despite its comparatively weak predictive performance. At the other end of the spectrum, k-NN, Decision Trees, LASSO, and Linear Regression train rapidly.

Symbolic Regression was omitted from Figure \ref{fig:training_time_regr} because its execution time was intentionally set to reach the maximum allowable threshold. This extended duration was necessary to accommodate the extensive exploration of the functional space that SR requires. Because every candidate expression must eventually be evaluated across all samples, this exploration becomes exceptionally computationally demanding on larger datasets. We confirmed, however, that this time constraint did not artificially degrade performance on larger datasets; supplementary tests on a random subset showed that extending the training duration did not improve predictive accuracy. Additional details are provided in Appendix \ref{sec:ablation_sr}.


The classification training times, presented in Figure \ref{fig:training_time_clf}, largely mirror the expectations set by the regression task. IGANN again exhibits the longest training durations. GOSDT ranks second, taking approximately two orders of magnitude longer than standard Decision Trees, which remain among the fastest methods overall. EBM and Logistic Regression follow in duration, while k-NN, Decision Trees, and Naive Bayes are highly efficient. It is worth noting that Logistic Regression trains slower than its regression counterpart (Linear Regression) because it relies on iterative gradient descent optimization, whereas Linear Regression utilizes a closed-form Ordinary Least Squares solution.

Ultimately, these timing metrics reveal a pervasive trade-off spanning both regression and classification tasks.  More expressive methods capable of representing richer function classes (e.g., EBM, IGANN, PR+LASSO) require one to two orders of magnitude more computation. Furthermore, algorithms explicitly designed to search for sparser, more interpretable solutions (such as GOSDT) or smoother shape functions (like IGANN) incur a heavy computational penalty compared to their standard, less constrained counterparts.

\section{Discussion}\label{sec:discussion}
To contextualize our results, we first examine them alongside existing comparative benchmarks and the broader literature on data complexity. Following this, we explore the implications of our meta-learning analysis for algorithm selection. We conclude by providing practical guidelines for practitioners seeking to implement interpretable methods.

\subsection{Contextualization within Comparative Benchmarks}
When examining specific model performances, our results both align with and challenge existing literature. The work of \citet{kraus2023interpretable}, who introduced IGANN, used a stratified analysis on PMLB datasets and showed that IGANN could outperform other interpretable models, including EBM, especially on small or low-dimensional data. This contrasts with our findings that EBM often exhibits stronger predictive performance, a difference we attribute to the limited hyperparameter search space used in their study for EBM (where the number of possible interactions was set to zero). However, our classification results and observations regarding IGANN's significantly higher execution times are entirely consistent with their findings.

Consistent with the findings of \citet{la2021contemporary} and \citet{wilstrup2021symbolic}, our evaluation confirms that Symbolic Regression consistently outperforms traditional Linear Regression models. \citet{wilstrup2021symbolic} concluded that SR models generalize better to out-of-sample data than Linear Regression, LASSO, and Decision Trees. This aligns with our observation that SR captures non-linear relationships that simple linear models miss.
Expanding upon these prior studies, our results highlight two critical behaviors of SR regarding data dimensionality and sample size. First, SR demonstrates remarkable robustness in high-dimensional settings. This is a non-trivial outcome, as it emphasizes that modern implementations utilizing initial feature selection steps, such as PySR, successfully mitigate the combinatorial explosion that has historically been the primary limitation of symbolic search. Second, contrary to its strength with many features, we observe a surprising degradation in SR's relative performance as the number of samples increases. As discussed in our results, this underperformance is not driven by training time bottlenecks over large datasets, but rather stems from the strict parsimony of the model selection criterion, which overly penalizes complexity and limits the equation's ability to capture subtle patterns in massive datasets.

Comparisons involving traditional classification families also align with \citet{wainer2016comparison}, who found that k-NN generally outperforms Naive Bayes, with penalized linear models positioning between the two. Finally, \citet{scholz_comparison_2021} stratified synthetic data and found that no classifier consistently dominates in all settings. While we agree that LASSO-type regularization does not always guarantee top-tier performance and that individual Decision Trees are rarely the strongest performers, we diverge regarding k-NN. Although \citet{scholz_comparison_2021} found that k-NN achieved high ranks in small, high-dimensional synthetic datasets, our analysis on real-world data highlights this specific regime as a weakness for k-NN, particularly in out-of-sample settings where the feature space is not sufficiently tiled by training examples.

\subsection{Algorithm Selection and the Predictability Gap}
A primary contribution of this study is the meta-analysis of algorithm performance, which directly addresses the classic Algorithm Selection Problem (ASP) \citep{khan2020literature}. The ASP posits that no single algorithm is universally superior, a concept formalized by the "No Free Lunch" theorem \citep{wolpert2002no, vanschoren2018meta}, and seeks to map measurable characteristics of datasets to expected algorithm performance.
Interestingly, while the vast majority of meta-learning literature has historically focused on classification tasks \citep{khan2020literature}, our meta-models reveal a profound  ``predictability gap'' between regression and classification. 

For regression, algorithm performance and pairwise superiority are highly predictable from dataset meta-features. In contrast, classification performance proves highly idiosyncratic, with meta-features failing to reliably predict task hardness or pairwise differences between methods. Across all analyses, we find no evidence of a stable or recoverable performance hierarchy in classification. Neither stratification, dimensionality reduction, nor clustering reveals any consistent structure, indicating that current meta-features are insufficient for reliable characterization or prediction.

This suggests that, unlike regression, where fitting the underlying data manifold reveals the distinct inductive biases of each model, classification operates on overlapping class distributions, where performance depends on both boundary placement and probability calibration. As a result, the mapping from data geometry to model performance becomes inherently less stable and highly dataset-specific, preventing the emergence of a consistent correspondence between dataset characteristics and algorithm behavior. Consequently, although models induce different decision functions and probability estimates, their inductive biases are not reliably reflected in observable performance patterns, even when evaluated using proper scoring rules such as the Brier score.

As a result, reliable \emph{a priori} algorithm selection in classification remains fundamentally challenging.


\subsection{Driving Factors of Dataset Difficulty}
By extracting feature importances from our meta-models, we identified the specific data complexity measures that drive problem difficulty.
For regression, our analysis highlights target correlation ($C1_r$), smoothness (S3), and non-linearity (L1) as the most critical determinants of model success. This perfectly corroborates the foundational regression complexity framework recently proposed by \citet{maciel_measuring_2016}, who introduced these metrics to the regression domain. Specifically, \citet{lorena2018data} adapted maximum feature correlation, nearest-neighbor error (smoothness), and linear error from classification to regression, and our findings confirm that these specific geometric properties, rather than basic dimensionality or sparsity metrics (N,d, T2), dictate the success of interpretable regression models.
For classification, our meta-model relied heavily on the Error Rate of the Nearest Neighbor Classifier (N3) and the Maximum Fisher's Discriminant Ratio (F1). These metrics are foundational classification complexity measures originally proposed by \citet{ho2002complexity} and are widely recognized in modern meta-learning surveys as powerful descriptors of class overlap and decision boundary complexity \citep{rivolli2018characterizing}. The prominence of F1 and N3 in our results confirms that the main source of difficulty in interpretable classification lies in the geometric interleaving of classes, rather than structural data properties \citep{rivolli2018characterizing}.
We must note a discrepancy in predictive strength, however: while these meta-features act as robust predictors of model success in regression, their influence is significantly less pronounced in classification scenarios.

\subsection{ Interpretability and Computational Tax}
In interpretable machine learning, predictive accuracy must be weighed against structural complexity and computational efficiency. Measuring interpretability objectively is notoriously ambiguous, as noted by \citet{kruschel2025challenging}, whose criteria yielded counter-intuitive results (e.g., ranking Decision Trees low on interpretability).

By comparing models within the same structural families, we observed clear trade-offs. Within linear models, extending the feature space (PR+LASSO) captures non-linear relationships and boosts accuracy but inherently sacrifices transparency by exploding the number terms compared to standard linear models. Similarly, for tree-based models, GOSDT produces significantly sparser and more interpretable trees than traditional Decision Trees, but this structural parsimony demands up to two orders of magnitude more computational time. More expressive methods (EBM, IGANN, PR+LASSO, SR) universally incur heavy computational penalties, highlighting that the search for optimal, inherently interpretable representations is fundamentally bounded by execution time constraints.

\subsection{Practical Guidelines for Practitioners
}
Based on our analysis, we offer the following guidelines for selecting interpretable machine learning models. The decision process is heavily dependent on the specific type of task, the available computational budget, and the desired form of interpretability.

\paragraph{For Regression Tasks}
The regression landscape is highly structured, allowing practitioners to make informed, a priori algorithm selections based on dataset characteristics.
\begin{itemize}
    \item Default to EBM or SR for Accuracy: EBM and Symbolic Regression consistently yield the highest predictive performance.
    \item Account for Out-of-Sample Robustness: Models like Decision Trees and PR+LASSO are more sensitive and prone to performance degradation on unseen data, whereas strictly regularized models like LASSO provide superior out-of-sample stability.
    \item Beware of Scalability Limits: While SR is highly competitive, its computationally expensive search process makes it impractical for very large datasets or strictly time-constrained environments. 
    \item Leverage Dataset Meta-Features: Practitioners should compute dataset meta-features to guide model selection (e.g., use LASSO over k-NN when dealing with highly sparse data, or rely on SR, GAMs, or Decision Trees rather than linear models when the data exhibits high non-linearity).
    \item Weigh Expressivity Against Time: Models that capture complex, non-linear relationships (like PR+LASSO or IGANN) incur a massive computational penalty compared to standard linear models or decision trees.
\end{itemize}

\paragraph{For Classification Tasks}
Unlike regression, classification performance is highly idiosyncratic and heavily dataset-dependent, making a priori algorithm selection difficult.

\begin{itemize}
    \item Adopt an Empirical Approach: Apart from a slight overall advantage for EBM, performance distributions overlap significantly. Because meta-features cannot reliably predict the best-performing model, practitioners must allocate time to train and evaluate multiple models.
    \item Factor in the Interpretability Tax: For practitioners who strictly require a specific model structure (e.g., decision trees), there is a stark trade-off between structural simplicity and training time. For instance, while GOSDT produces significantly sparser and more interpretable trees than traditional Decision Trees, it requires up to two orders of magnitude more computational time.
\end{itemize}

\section{Conclusions}\label{sec:conclusions}

This study presented an extensive empirical evaluation of intrinsically interpretable machine learning models, addressing the lack of systematic guidance for practitioners in the field. By benchmarking sixteen diverse methods across more than two hundred real-world datasets, we moved beyond aggregate performance metrics to uncover how model behavior is influenced by meta-features such as sample size, dimensionality, linearity, and class imbalance. Our analysis confirms that while modern interpretable approaches can achieve competitive predictive accuracy, model selection should be carefully tailored to the specific properties of the data at hand rather than relying on a one-size-fits-all solution.

Our central finding is a profound dichotomy between regression and classification tasks. In regression, a clear and predictable performance hierarchy emerges, led by EBM and Symbolic Regression, and can be reliably anticipated from dataset meta-features such as target correlation, non-linearity, and local smoothness. In contrast, classification exhibits no stable or recoverable performance hierarchy. Model performances are heavily overlapping, and meta-features fail to reliably predict pairwise algorithm superiority. This suggests that classifier performance cannot be organized into a consistent, meta-feature-driven structure, and instead depends on complex, dataset-specific class geometry rather than broad structural properties.

Cutting across both tasks, we identified a pervasive interpretability tax: methods that optimize for structural parsimony or expressive non-linearity can demand up to two orders of magnitude more training time than simpler baselines, a cost practitioners must weigh explicitly against the transparency benefits.

In this work, we deliberately focused on predictive performance and computational efficiency, and did not attempt to quantify interpretability itself, reflecting the absence of a universally accepted and objective interpretability metric. Developing rigorous, model-agnostic frameworks for measuring interpretability, therefore, remains an important direction for future research. Additionally, to accommodate the structural constraints of certain models and standardize our out-of-sample shift simulations, we restricted classification tasks to binary problems. While multiclass tasks can conceptually be decomposed into binary problems, native multiclass extensions introduce distinct complexities that may alter the observed performance hierarchy and computational tax. Future benchmarking should evaluate natively multiclass environments to verify if the classification predictability gap persists.

Ultimately, these findings contribute to a deeper empirical understanding of interpretable modeling for tabular data. They provide researchers and practitioners with a data-driven foundation for algorithm selection, emphasizing that the pursuit of inherently interpretable representations must be carefully balanced against specific task dynamics, out-of-sample robustness requirements, and available computational budgets.

\printcredits

\section*{Declaration of Competing Interest}
The authors declare that they have no known competing financial interests or personal relationships that could have appeared to
influence the work reported in this paper

\section*{Acknowledgement}
We acknowledge the CINECA award under the ISCRA initiative, for the availability of high performance computing resources and support.
\bibliographystyle{cas-model2-names}

\bibliography{cas-refs}



\appendix
\counterwithin{equation}{section}
\counterwithin{table}{section}
\counterwithin{figure}{section}

\section{Classification and Regression Methods}\label{sec:app_models}
In this study, we evaluate a diverse set of interpretable machine learning models for both classification and regression tasks. These methods are selected to represent a range of modeling paradigms, from linear and tree-based approaches to probabilistic, instance-based, and symbolic methods. We focus on models that are intrinsically interpretable and whose implementations are readily available and stable. The following subsections provide an overview of each model category, highlighting their interpretability characteristics and core assumptions.

\subsection{Linear models: Logistic, Linear, Lasso, GLM, GAM}
Linear models are among the most widely used interpretable methods due to their simplicity and transparency. Linear Regression models continuous outcomes as a linear combination of input features. In classification tasks, Logistic Regression estimates the probability of categorical outcomes by applying a logistic function to a linear combination of features, mapping logit values to probabilities.

Extensions of linear models, such as LASSO regression, introduce regularization to promote sparsity, improving interpretability by selecting a subset of relevant features through an L1 penalty \citep{tibshirani1996regression}. Generalized Linear Models extend linear models to accommodate non-normal target distributions (e.g., Poisson, Gamma) via appropriate link functions \citep{mccullagh2019generalized}. Generalized Additive Models (GAMs) further enhance flexibility by allowing non-linear transformations of individual features, such as splines, or interactions between feature pairs, while maintaining an additive structure that preserves interpretability \citep{hastie1986generalized}.
Different formulations and implementations of GAMs have been proposed. 
Explainable Boosting Machines (EBMs) extend the concept of GAMs using gradient boosting to capture complex feature interactions while preserving interpretability through per-feature visualizations and monotonic constraints \citep{nori2019interpretml}.

Recently, \citet{kraus2023interpretable} introduced a novel variant, the Interpretable Generalized Additive Neural Network (IGANN), which promotes linearity by initializing shape functions linearly and allowing deviations only when required by the underlying features.

\subsection{Tree-based models: Decision Tree, GOSDT}
Tree-based models provide a rule-based representation of decision processes. Decision Trees recursively partition the feature space into homogeneous regions, producing hierarchical rules that are easy to visualize \citep{breiman1984classification}. However, their interpretability and predictive performance can be compromised by suboptimal or overly large trees \citep{lipton2018mythos}, as classical algorithms often rely on greedy splitting heuristics. 
Optimal Sparse Decision Trees (OSDT) introduced a practical, bound-driven search for provably optimal trees on binary variables, improving sparsity compared to greedy methods \citep{hu2019optimal}. Subsequent work generalized and scaled this approach to handle continuous features and imbalanced data \citep{lin2020generalized}. More recent methods apply reference ensembles from black-box models to guide the search, substantially reducing computation while producing sparse trees that remain interpretable \citep{mctavish2022fast}.

\subsection{Probabilistic models: Naive Bayes}
Naive Bayes is a probabilistic approach grounded in Bayes' theorem, assuming conditional independence among features \citep{hastie2009elements}. Despite its simplicity, it is effective in many real-world scenarios and provides interpretable probabilistic outputs. Feature contributions to predictions can be directly assessed through the model's likelihood functions, offering insights into how individual variables influence the predicted probabilities.

\subsection{Instance-based models: KNN}
Instance-based methods, such as k-Nearest Neighbors (k-NN), make predictions by comparing new instances to stored examples in the training data \citep{cover1967knn}. k-NN is often considered interpretable because predictions can be traced back to the most similar observed instances, providing an intuitive understanding of how the model makes decisions.

While some may argue that instance-based methods offer only local explanations, similar to post-hoc methods like LIME \citep{ribeiro2016should}, k-NN's predictions are fully determined by the nearest training samples, without any additional approximation or model transformation. Consequently, the behavior of the model across the entire dataset can, in principle, be understood by examining the stored instances, and the influence of each feature is directly observable through the similarity metric.

\subsection{Symbolic models: Symbolic Regression}
Symbolic Regression seeks to discover explicit mathematical expressions that describe the relationship between input features and the target variable \citep{la2021contemporary}. By generating human-readable formulas, Symbolic Regression offers a high degree of interpretability, allowing domain experts to understand and validate the underlying patterns captured by the model. Methods in this category balance predictive accuracy with simplicity, often guided by constraints on expression complexity.

\section{Taxonomy of Interpretability}\label{sec:taxonomy}
While the term ``interpretability'' is ubiquitous in current literature, it remains an ill-defined concept, often treated as a monolithic property rather than a multifaceted and often subjective one. 
While we distinguished intrinsic interpretability from post-hoc explainability in Section \ref{sec:intro}, the category of inherently interpretable models itself is far from homogeneous.
Following the frameworks established by \citet{lipton2018mythos} and \citet{molnar2020interpretable}, we categorize our candidate models based on three core dimensions of transparency: simulatability, decomposability, and algorithmic transparency. 
\paragraph{Simulatability}
A model is considered simulatable if a human can reasonably inspect the entire model at once and calculate its output for a given input. This property is strictly constrained by the cognitive load required to trace a decision path. In our experimental design, this is operationalized through structural constraints. For tree-based methods, we ensure simulatability for standard Decision Trees by limiting maximum depth, and explicitly test GOSDT, which optimize for sparsity to produce provably minimal representations. Similarly, Symbolic Regression targets simulatability by searching for parsimonious mathematical expressions that can be inspected directly.

\paragraph{Decomposability}
Decomposability requires that the individual components of a model, such as input features, weights, or shape functions, can be analyzed independently. The linear models in our benchmark satisfy this by assigning scalar weights to features, where the magnitude and sign directly indicate contribution. We extend this category to include GAMs (specifically EBM and IGANN). While computationally complex, these models remain decomposable because they isolate feature effects into additive shape functions or interaction terms that can be visualized and understood independently of the rest of the model.

\paragraph{Algorithmic Transparency}
Algorithmic transparency refers to models where the mathematical process from input to output is fully open and can be scrutinized. This category includes $k$-NN (and instance-based algorithms in general) and Naive Bayes. While these models do not always produce a compact formula, they are transparent because their decision-making process is grounded in clear mathematical axioms: $k$-NN predictions are fully determined by the most similar stored instances, and Naive Bayes relies on explicit conditional independence and likelihood functions.

\section{Missing Values}\label{sec:missing_values}
\begin{figure*}
    \centering
    \begin{subfigure}[b]{0.48\textwidth}
        \centering
        \includegraphics[width=\columnwidth]{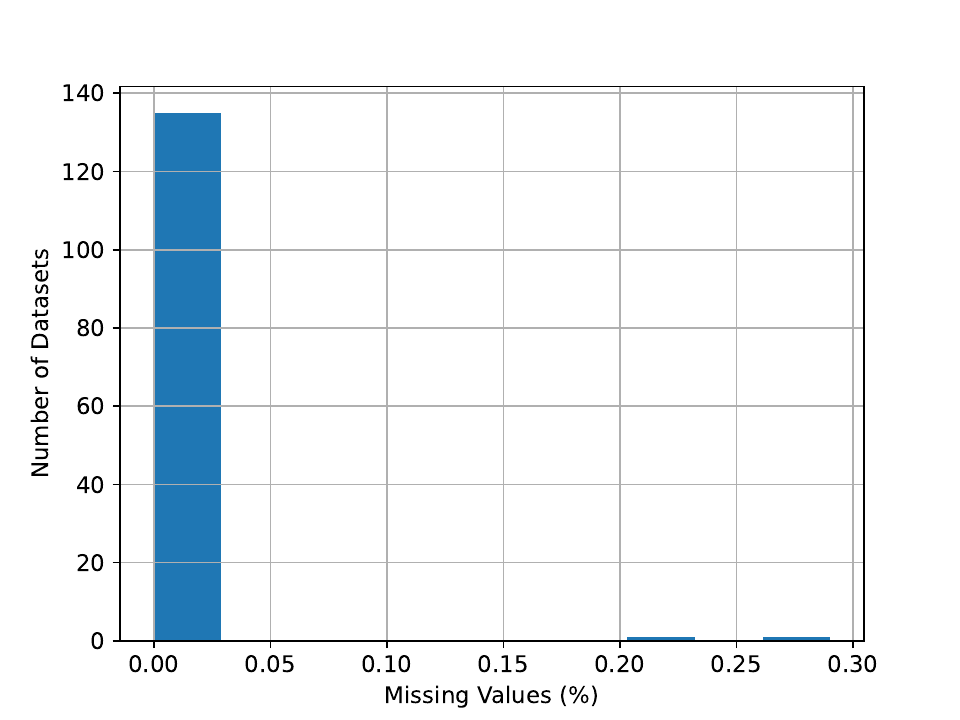}
        \caption{Regression}\label{fig:missing_regr}
    \end{subfigure}
    \hfill
    \begin{subfigure}[b]{0.48\textwidth}
        \centering
        \includegraphics[width=\columnwidth]{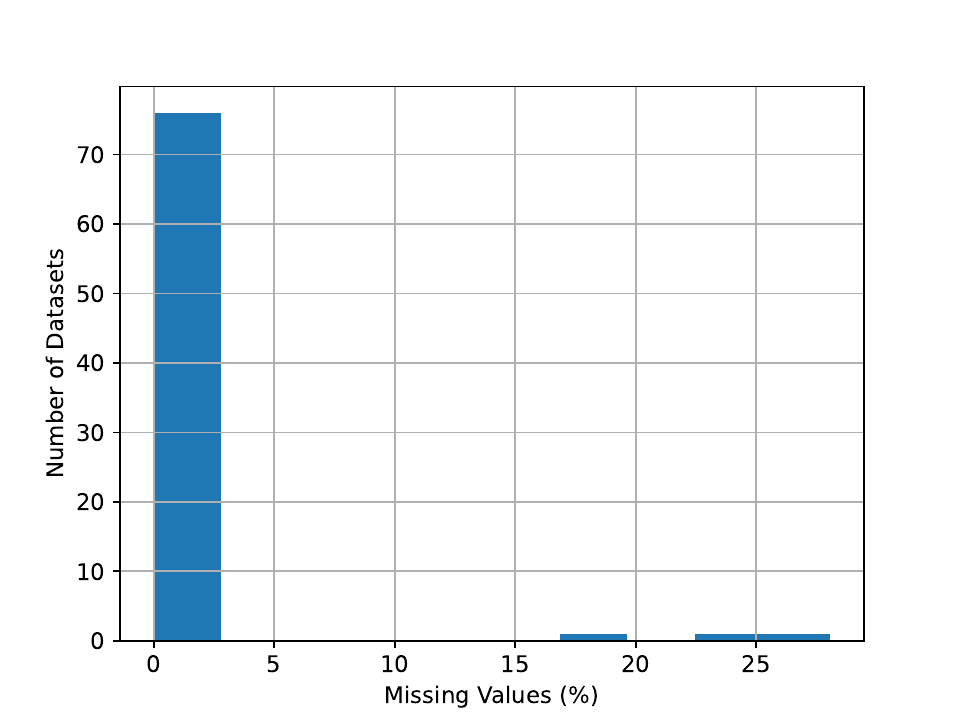}
        \caption{Classification}\label{fig:missing_clf}
    \end{subfigure}
    \caption{Percentage of missing values in the benchmark datasets.}
    \label{fig:missing}
\end{figure*}

In the main text, we described the preprocessing pipeline used in our experiments (Section \ref{sec:ds_sel_pre}). A key step in this process was the exclusion of rows and columns containing missing values. While discarding data, much like arbitrary imputation, can potentially introduce bias, we provide a quantification of these missing values here to demonstrate their minimal impact. Figures \ref{fig:missing_regr} and \ref{fig:missing_clf} illustrate the percentage of missing cells across all datasets.

As observed, the vast majority of the datasets contain no missing values. Specifically, only two regression datasets and seven classification datasets exhibit any missing data. Furthermore, the proportion of missing cells is negligible (under 1\%) for all regression datasets and for all but three classification datasets. Consequently, we are confident that the deletion of these missing values did not meaningfully alter our findings.

\section{Hyperparameter Ranges}\label{sec:app_hyper}
Table~\ref{tab:hyperparams} reports the hyperparameter ranges used for model selection across all interpretable methods. Search spaces were chosen to reflect standard practice while enforcing constraints that favor parsimonious and interpretable model configurations.

\begin{table*}[ht!]
\centering

\begin{tabular}{lll}
\toprule
\textbf{Model} & \textbf{Tuning Parameters} & \textbf{Range} \\
\midrule
k-NN (clf/regr) & $n\_neighbors$ & $[3, \min(100, 0.2 \cdot n\_{samples})]$ \\
\midrule
Decision Tree (clf) & $min\_samples\_leaf$ & $[0.01, 0.3]$ \\
                   & $criterion$ & \{gini, entropy\} \\
                   & $max\_depth$ & $[2, \min(6, n\_{features}/2 + 1)]$ \\
Decision Tree (regr) & $max\_depth$ & $[2, \min(6, n\_{features}/2 + 1)]$ \\
\midrule
Logistic Regression & $C$ & $[10^{-2}, 10^1]$ (log-scale) \\
                    & $penalty$ & \{l2, None\} \\
\midrule
GOSDT & $GDBT\_n\_estimators$ & $[20, 40]$ \\
      & $GDBT\_max\_depth$ & $[2, 5]$ \\
      & $GOSDT\_regularization$ & $[0.01, 0.2]$ (log-scale) \\
      & $GOSDT\_similar\_support$ & \{False, True\} \\
      & $GOSDT\_depth\_budget$ & $[2, 6]$ \\
\midrule
Naive Bayes & $var\_smoothing$ & $[10^{-12}, 10^{-1}]$ (log-scale) \\
\midrule
EBM (clf/regr) & $interactions$ & $[0, 2]$ \\
          & $learning\_rate$ & $[0.01, 0.05]$ (log-scale) \\
          & $max\_rounds$ & $[100, 200]$ \\
          & $min\_samples\_leaf$ & $[2, 5]$ \\

\midrule
IGANN (clf/regr) & $init\_reg$ & $[10^{-7}, 1]$ (log-scale) \\
                 & $n\_hid$ & $[5, 20]$ \\
                 & $elm\_alpha$ & $[10^{-7}, 1]$ (log-scale) \\
                 & $elm\_scale$ & $[1, 10]$ \\
                 & $boost\_rate$ & $[0.1, 1]$ \\
                 & $n\_estimators$ & $[100, 5000]$ \\
\midrule
Symbolic Regression & $select\_k\_features$ & $[2, \min(n\_{features}, 12)]$ \\
                    & $operators$ & $\{+, -, \times, /, exp, \log, \sin, \cos\}$\\
                    & $populations$ & $[6, 20]$ \\
                    & $population\_size$ & $[32, 96]$ \\
                    & $weight\_optimize$ & $[10^{-3}, 10^{-1}]$ (log-scale) \\
                    & $parsimony$ & $[10^{-3}, 1]$ (log-scale) \\
                    & $maxsize$ & $[15, 70]$ \\
                    & $batching$ & $Auto$ \\
                    
\midrule
Linear Regr. & None & None \\
\midrule
Lasso & $alpha$ & $[10^{-4}, 10^2]$ (log-scale) \\
\midrule
GLM & $alpha$ & $[10^{-4}, 10^1]$ (log-scale) \\
\midrule
Polynomial Regr. + Lasso & $poly\_\_degree$ & \{2, 3\} \\
& $alpha$ & $[10^{-4}, 10^2]$ (log-scale) \\
\bottomrule
\end{tabular}
\caption{Hyperparameter search spaces for classification and regression models.}
\label{tab:hyperparams}
\end{table*}


\section{Complete Stratification Results}\label{sec:compl_strat}
\begin{figure*}
\centering
\includegraphics[width=\textwidth]{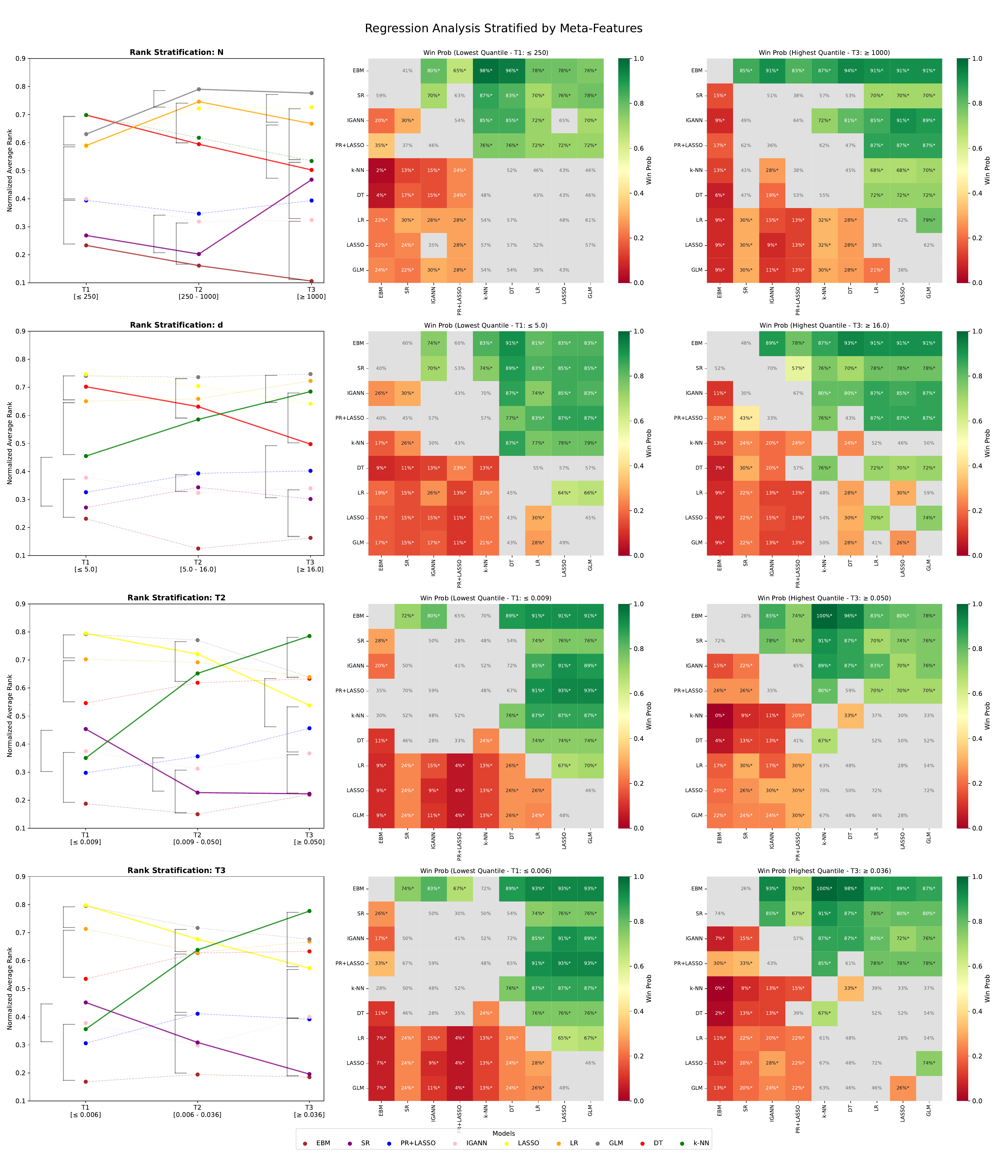}
\caption{Stratified results for the regression task (in-sample), part 1. 
}\label{fig:regr_is_strat_c1}
\end{figure*}

\begin{figure*}
\centering
\includegraphics[width=\textwidth]{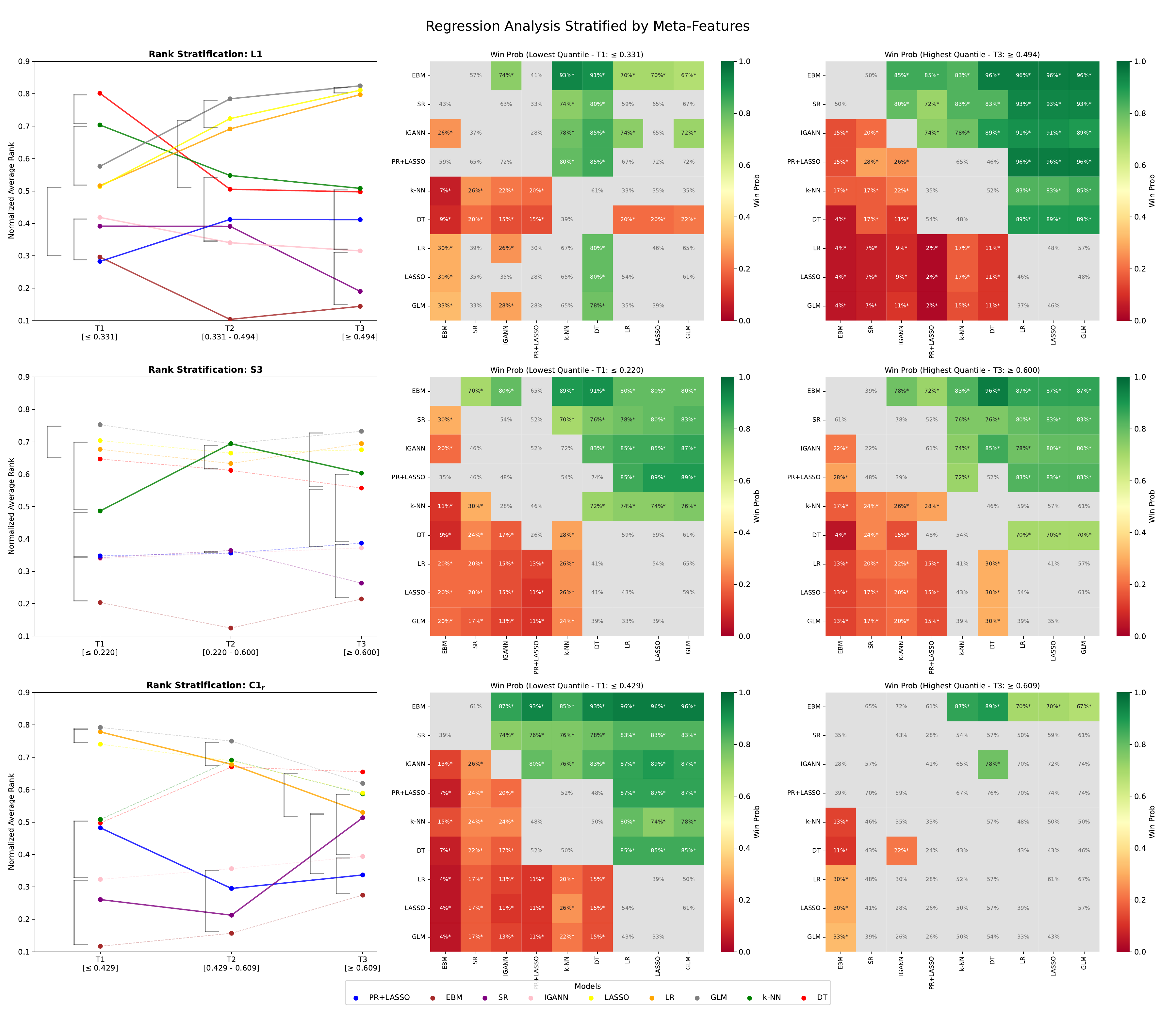}
\caption{Stratified results for the regression task (in-sample), part 2. 
}\label{fig:regr_is_strat_c2}
\end{figure*}

\begin{figure*}
\centering
\includegraphics[width=\textwidth]{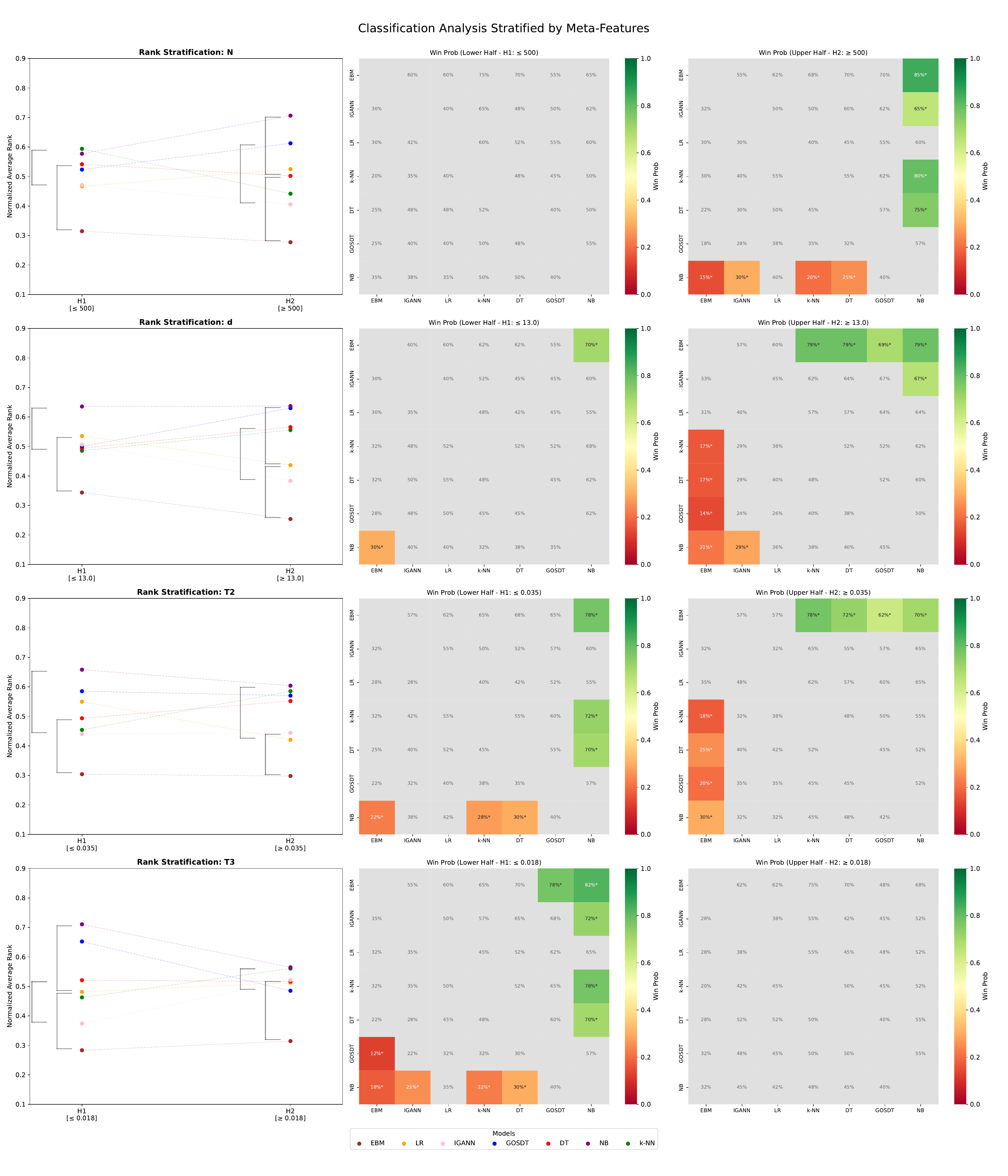}
\caption{Stratified results for the classification task (in-sample), part 1. 
}\label{fig:clf_is_strat_c1}
\end{figure*}

\begin{figure*}
\centering
\includegraphics[width=\textwidth]{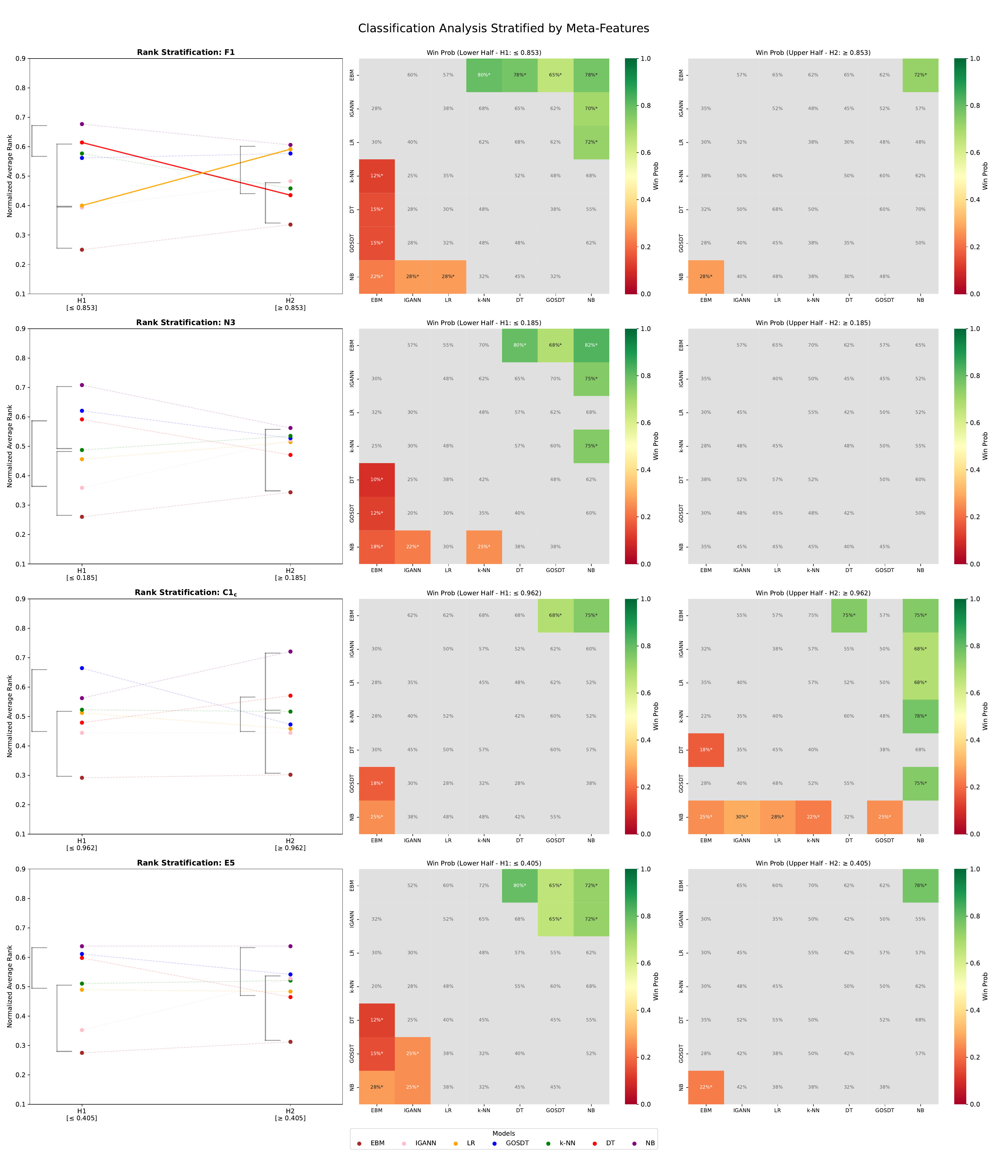}
\caption{Stratified results for the classification task (in-sample), part 2. 
}\label{fig:clf_is_strat_c2}
\end{figure*}

Figures \ref{fig:regr_is_strat_c1}, \ref{fig:regr_is_strat_c2}, \ref{fig:clf_is_strat_c1}, and \ref{fig:clf_is_strat_c2} present the stratified results across all evaluated meta-features. As discussed in the main text, we observe that while regression provides good stratification between the methods, this is not the case for classification.




\section{Ablation Study and Analysis for Symbolic Regression}\label{sec:ablation_sr}
\subsection{Training Time}
Throughout our comparative evaluation, model training was restricted to a maximum time budget of 300 seconds. While the execution times for the majority of the evaluated methods fell safely below this threshold, Symbolic Regression was a notable exception; its execution time was intentionally set to reach this maximum allowable limit to accommodate its extensive search of the functional space.
This naturally raises the question of whether this strict time limit artificially bottlenecked the method's capabilities, and if allocating a larger computational budget would have yielded improved predictive performance. This concern is particularly relevant for datasets with large sample sizes, where SR's ranks were observed to degrade compared to other methods.

To investigate this, we randomly selected 10 datasets from the tercile characterized by the largest sample sizes ($N$). We then re-evaluated SR on these datasets with a substantially increased time budget of 900 seconds (three times the original limit).

\begin{figure*}
\centering
\includegraphics[width=0.75\textwidth]{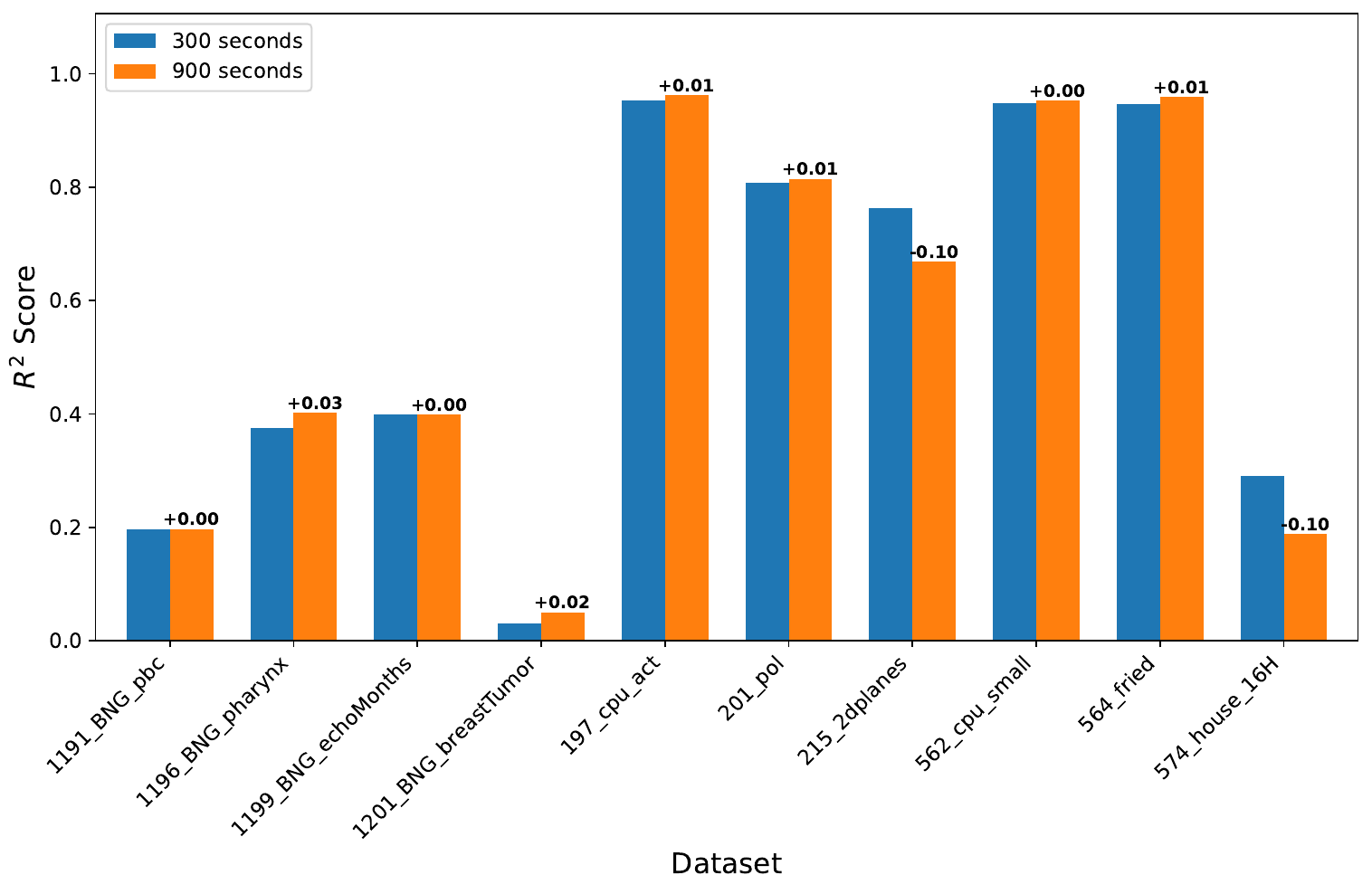}
\caption{Predictive accuracy comparison of Symbolic Regression given an extended time budget.}
\label{fig:ablation_symbolic}
\end{figure*}

Figure \ref{fig:ablation_symbolic} illustrates the difference in $R^2$ scores achieved under the two time constraints. We observe that the performance difference is negligible, leading us to conclude that the 300-second time budget was not a primary limiting factor for SR's predictive accuracy in our main experiments.

\subsection{Model Size}
Having ruled out computational constraints as the root cause of Symbolic Regression's relative performance degradation on large datasets, we investigate the role of the model selection process. SR algorithms, including the PySR implementation used in our study, typically generate a Pareto front of candidate equations that balance predictive accuracy against mathematical complexity. The final equation is chosen based on a scoring criterion that penalizes complexity to enforce parsimony.

\begin{figure*}
\centering
\includegraphics[width=0.75\textwidth]{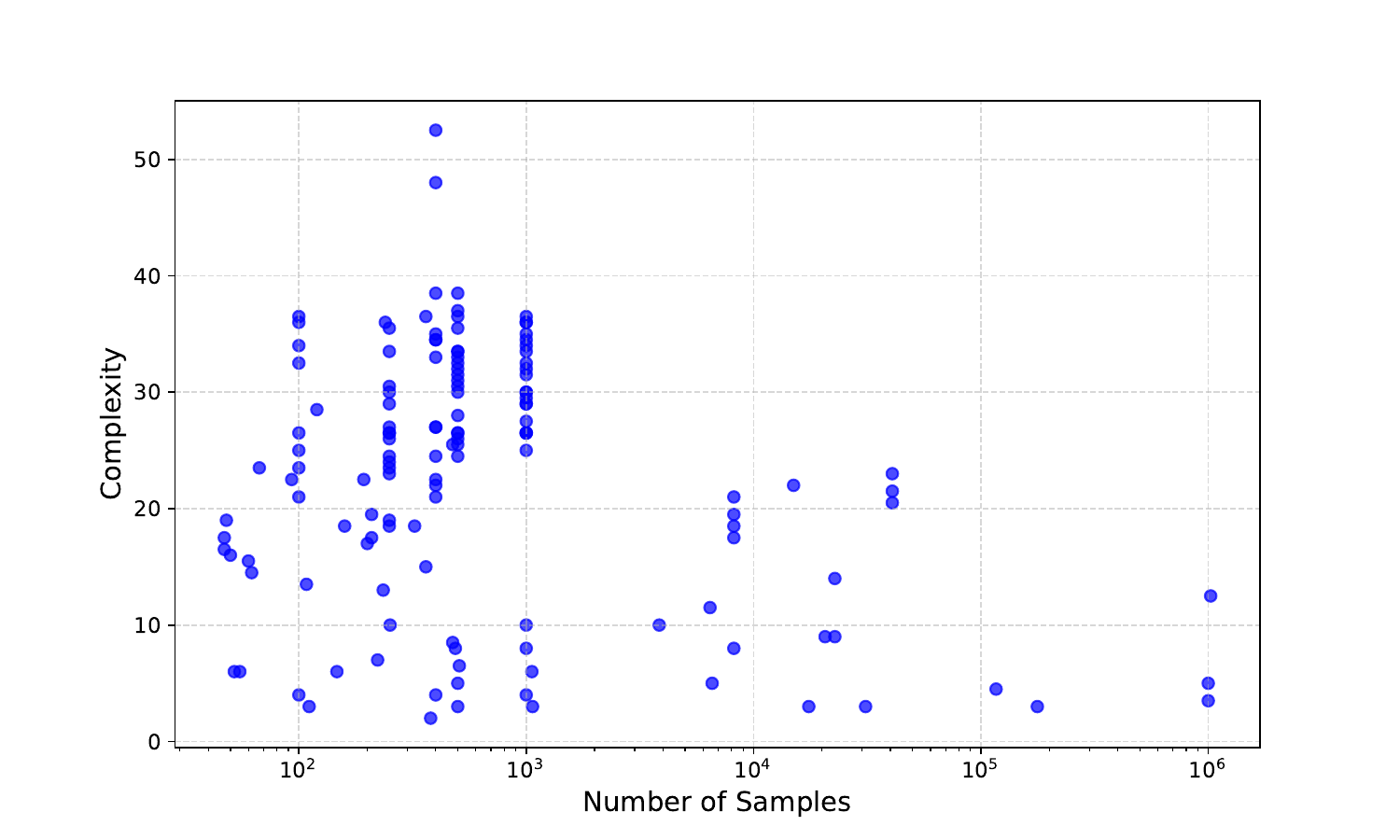}
\caption{Distribution of selected Symbolic Regression model complexities across varying dataset sample sizes.}
\label{fig:model_size_symbolic}
\end{figure*}

As illustrated in Figure \ref{fig:model_size_symbolic}, the structural complexity of the selected equations remains remarkably low, even as the dataset sample size ($N$) increases significantly. In large-sample regimes, there is generally sufficient statistical evidence to support more complex models capable of capturing finer, non-linear patterns without a high risk of overfitting. However, the strict parsimony penalty prevents the SR algorithm from selecting these more expressive formulations.


\section{Algorithm Selection and Classification Performance Landscape}\label{sec:clf_compl_res}
\begin{figure*}[t]
    \centering
    \begin{subfigure}[b]{0.48\textwidth}
        \centering
        \includegraphics[width=\columnwidth]{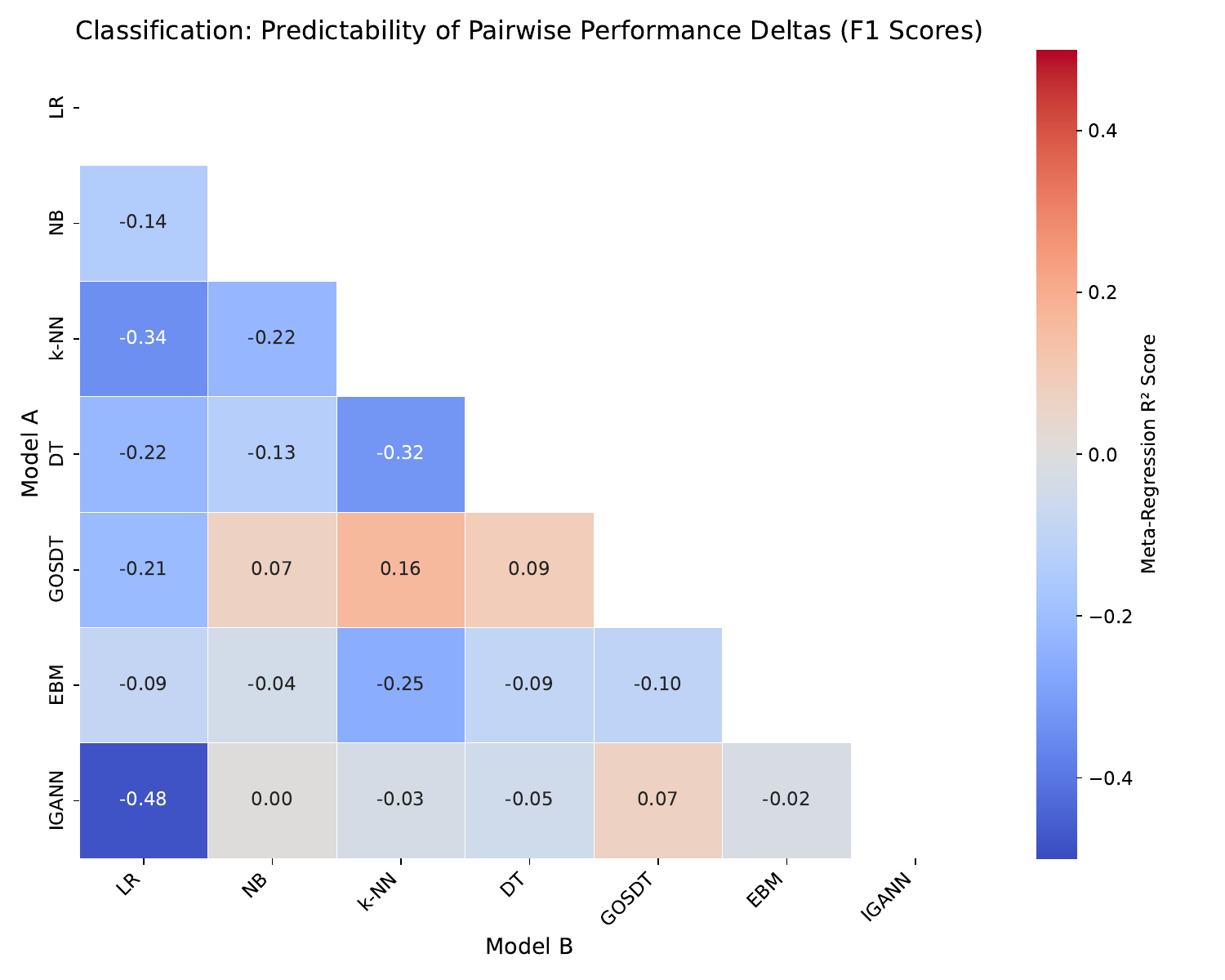}
        \caption{Pairwise}\label{fig:clf_is_metap}
    \end{subfigure}
    \begin{subfigure}[b]{0.48\textwidth}
        \centering
        \includegraphics[width=\columnwidth]{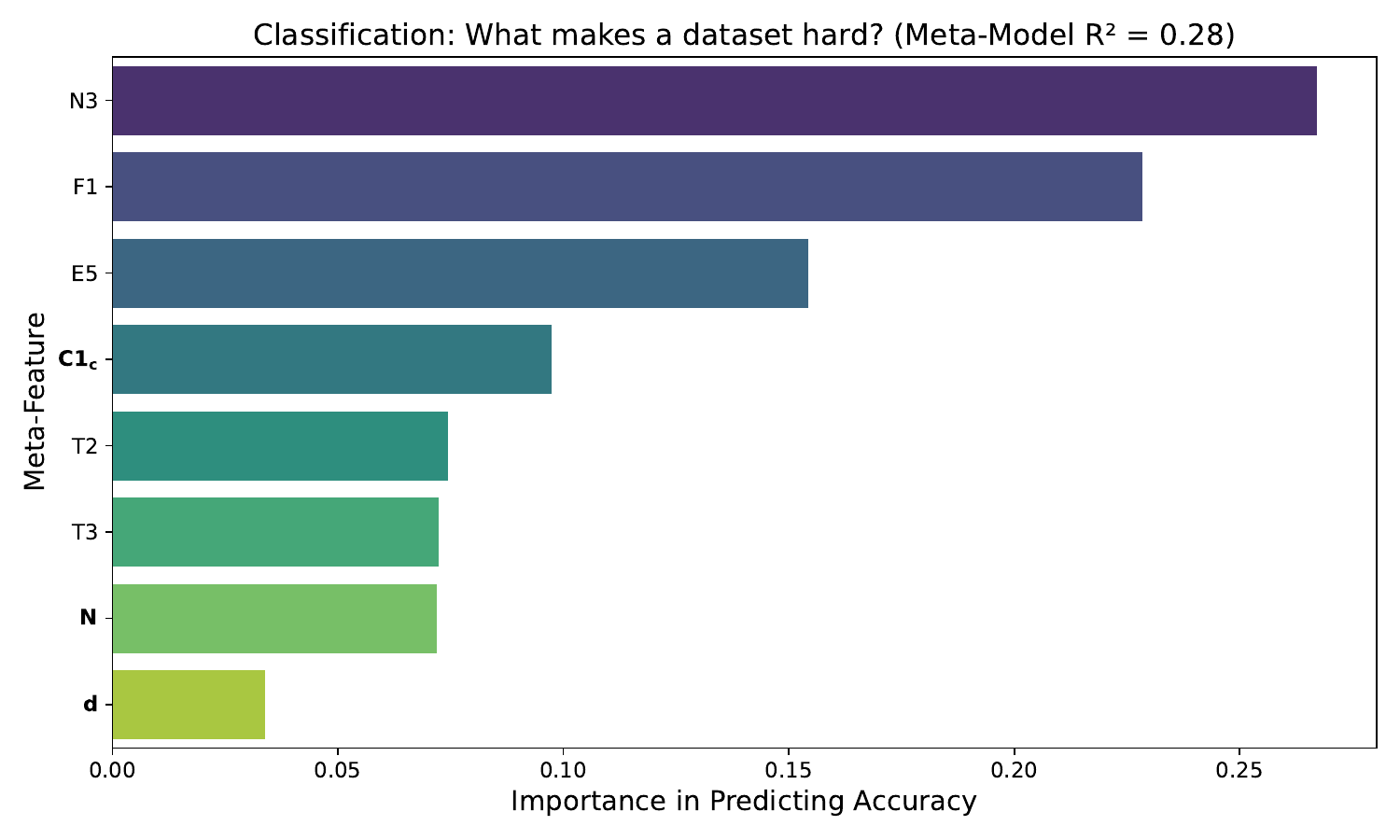}
        \caption{Feature Importance}\label{fig:clf_is_metaf}
    \end{subfigure}
    \caption{(a) $R^2$ scores of the meta-models predicting the $\Delta$ performance between pairs of models in classification. (b) Random Forest feature importances for predicting dataset difficulty in classification.}
    \label{fig:meta_clf}
\end{figure*}

As discussed in Section \ref{sec:operationalize}, we operationalized dataset meta-features to predict both pairwise algorithm superiority and overall dataset difficulty. Figure \ref{fig:meta_clf} visualizes the predictive outcomes for the classification task. The pairwise predictability matrix in Figure \ref{fig:clf_is_metap} reveals that the Random Forest meta-model struggles to forecast the performance difference between most pairs of classifiers, frequently yielding negative $R^2$ values. This confirms that meta-features provide little reliable guidance for a priori algorithm selection in the classification domain. When predicting inherent dataset difficulty, the meta-model achieves a comparatively low $R^2$ score of 0.28. Figure \ref{fig:clf_is_metaf} indicates that the most influential meta-features for determining task hardness are the Error Rate of the Nearest Neighbor Classifier (N3), the Entropy of the 5 nearest neighbors (E5), and the Maximum Fisher's Discriminant Ratio (F1). This emphasizes that classification difficulty is primarily driven by local class overlap and decision boundary complexity rather than broad structural data properties, similar to regression.

\begin{figure}
    \centering
    \begin{subfigure}{0.48\textwidth}
        \centering
        \includegraphics[width=\linewidth]{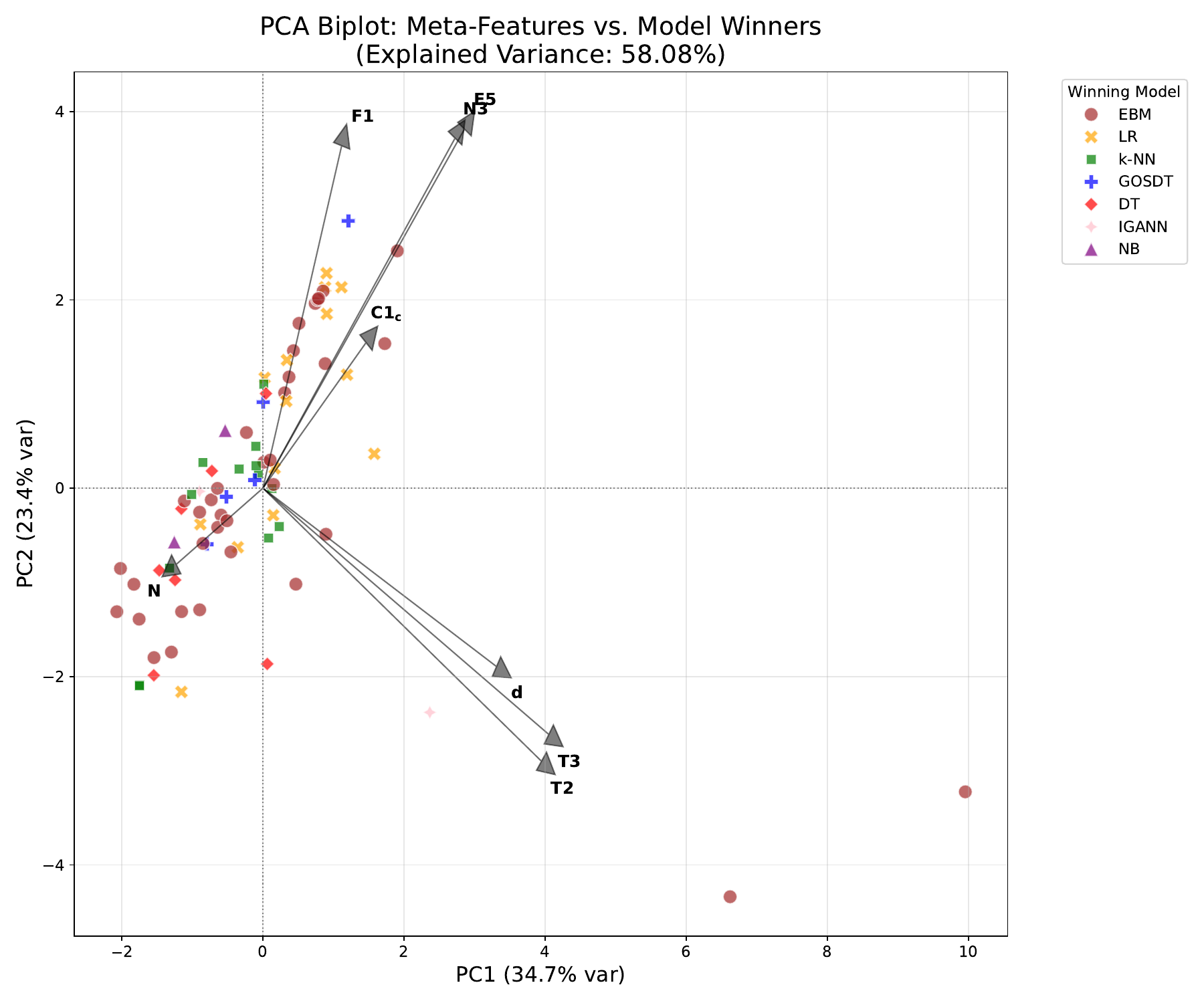}
        \caption{PCA biplot of the dataset meta-features (classification) projected onto the first two principal components. Individual datasets are color-coded by the winning algorithm based on the Brier score.}
        \label{fig:clf_is_pca_brier}
    \end{subfigure}\hfill
    \begin{subfigure}{0.48\textwidth}
        \centering
        \includegraphics[width=\linewidth]{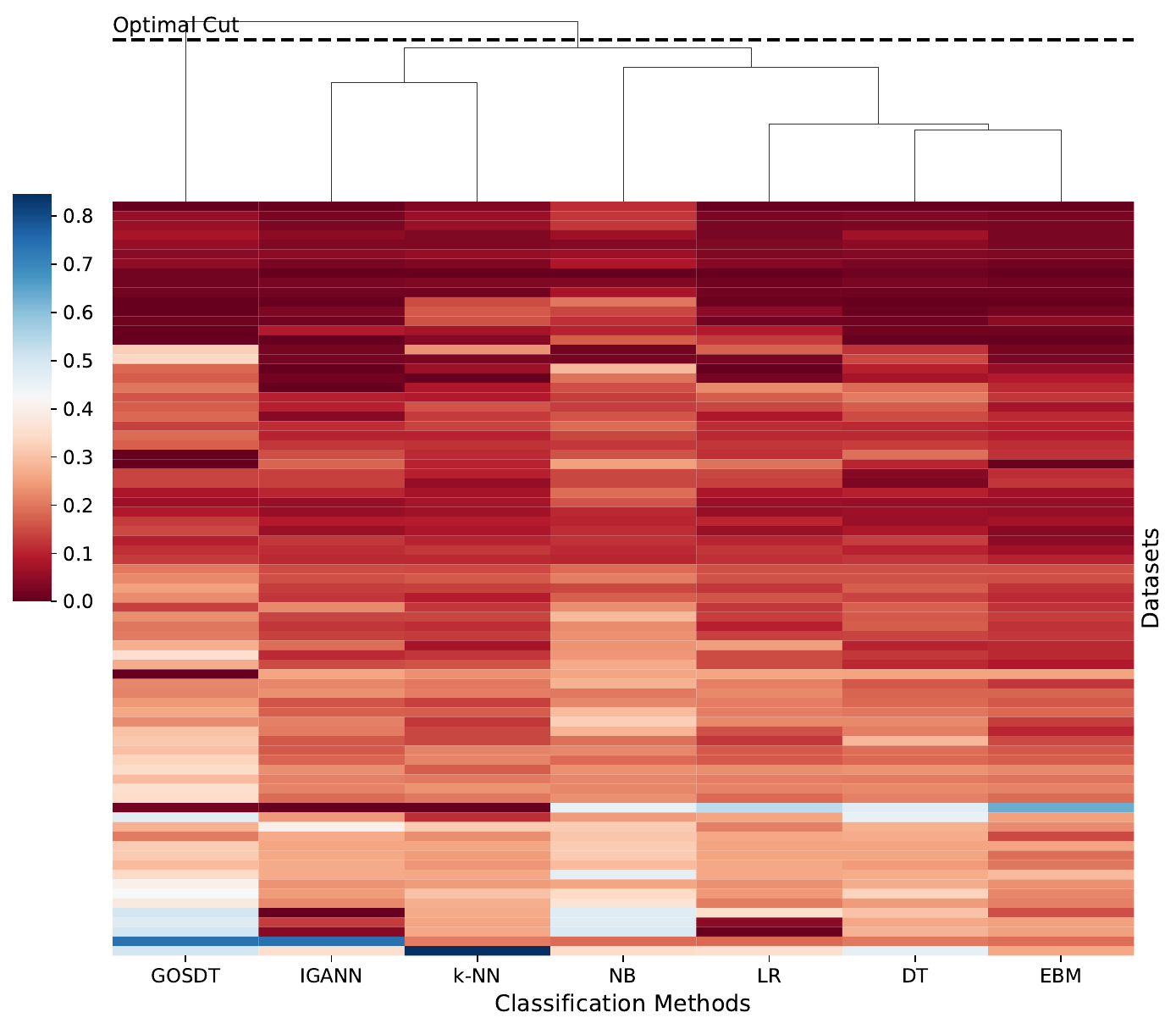}
        \caption{Hierarchical clustering of method performance profiles across classification datasets (Brier score).}
        \label{fig:clf_is_cluster_brier}
    \end{subfigure}
    \caption{Landscape analysis for classification tasks evaluated using the Brier score.}
    \label{fig:clf_landscape_brier}
\end{figure}

\begin{figure}
    \centering
    \begin{subfigure}{0.48\textwidth}
        \centering
        \includegraphics[width=\linewidth]{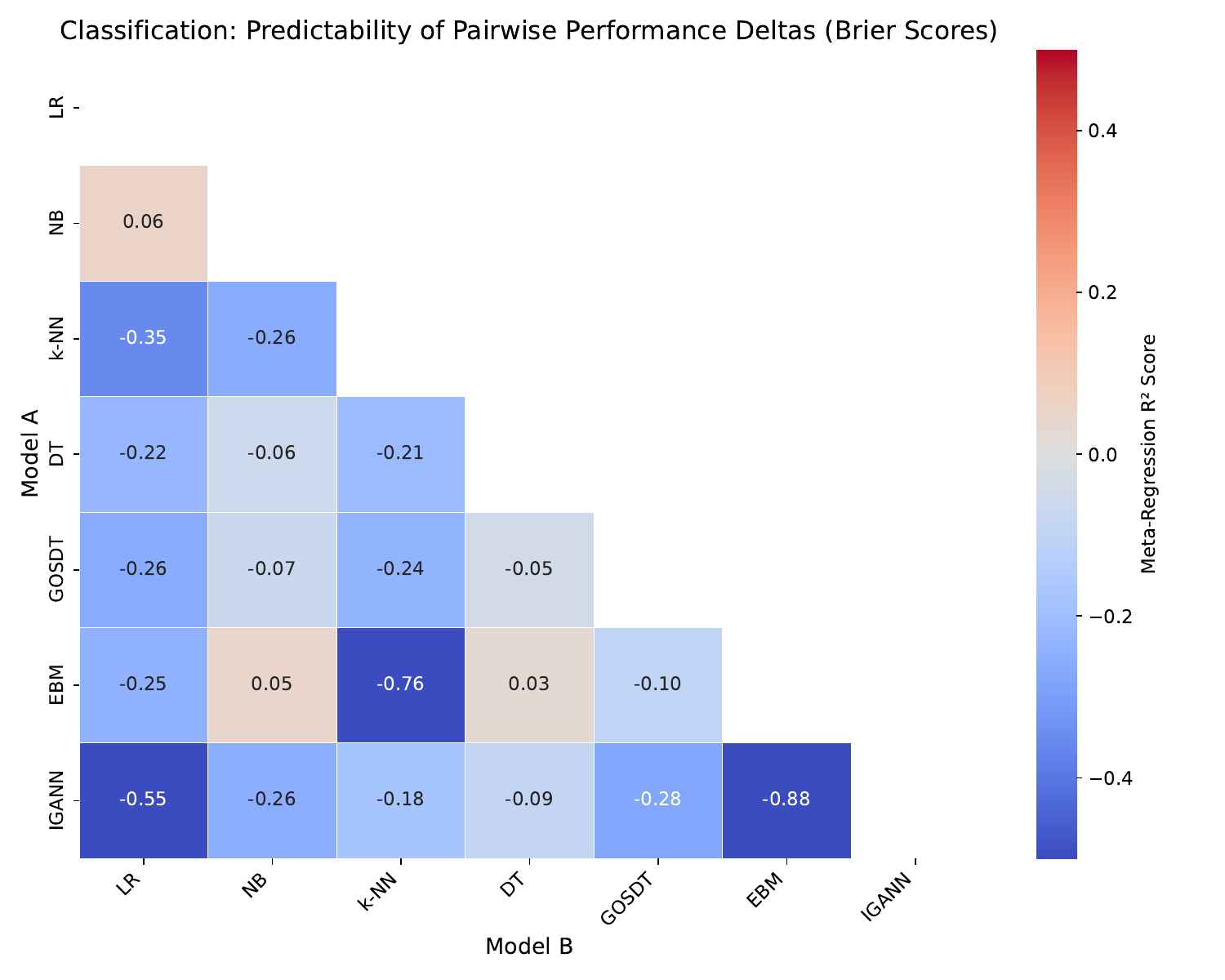}
        \caption{$R^2$ scores of the meta-models predicting the $\Delta$ performance (Brier score) between pairs of models in classification.}
        \label{fig:clf_is_metap_brier}
    \end{subfigure}\hfill
    \begin{subfigure}{0.48\textwidth}
        \centering
        \includegraphics[width=\linewidth]{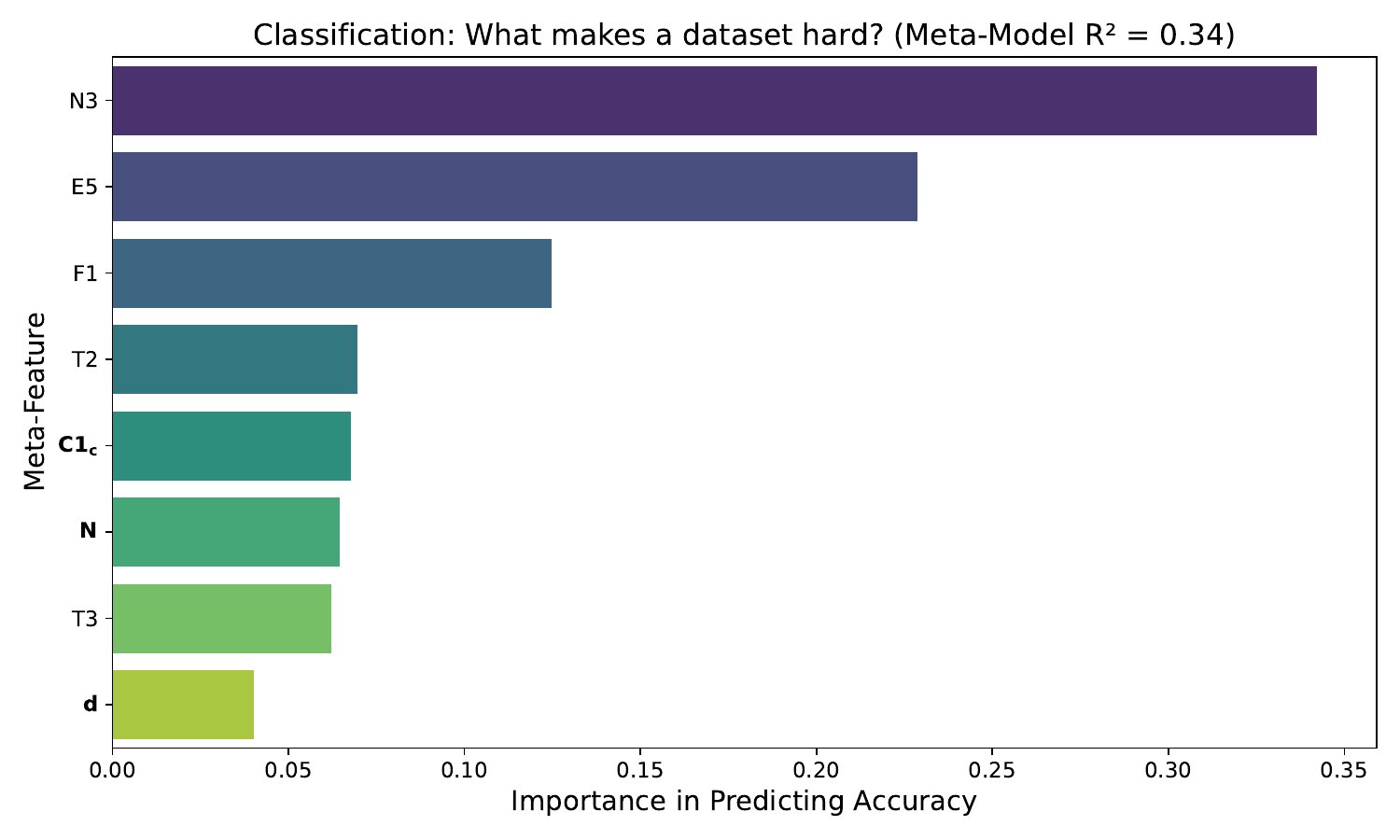}
        \caption{Random Forest feature importances for predicting dataset difficulty (Brier score) in classification tasks.}
        \label{fig:clf_is_metaf_brier}
    \end{subfigure}
    \caption{Predictability of pairwise performance deltas and dataset difficulty in classification tasks (Brier score).}
    \label{fig:clf_predictability_brier}
\end{figure}

In the main text, we demonstrated that classification performance lacks a stable hierarchy when evaluated using the F1 score. To confirm that this observation is not merely an artifact of the discrete nature of the F1 metric, we replicated the landscape analysis using the Brier score, which is a continuous proper scoring rule. Figure \ref{fig:clf_landscape_brier} presents the results of this extended analysis. The PCA biplot in Figure \ref{fig:clf_is_pca_brier} projects the dataset meta-features onto the first two principal components, with datasets colored by the best performing algorithm according to the Brier score. Similar to the F1 score results, there is no discernible clustering of winning models within the meta-feature space. Furthermore, the hierarchical clustering of the method performance profiles across datasets, shown in Figure \ref{fig:clf_is_cluster_brier}, fails to identify cohesive performance clusters. Specifically, for the Brier score clustermap, the overall Silhouette score was 0.22, which further decreased to 0.15 when the algorithm was forced to find a number of clusters equal to the number of methods. These findings validate that the highly interleaved and idiosyncratic nature of classifier performance remains consistent across different evaluation metrics. Finally, as reported in Figure \ref{fig:clf_predictability_brier}, predicting dataset difficulty using the Brier score achieves an $R^2$ of 0.34, closely mirroring the low predictability observed with the F1 score, and fully supporting these conclusions.

\section{Regression Performance Landscape}\label{sec:regr_compl_res}
For completeness and to provide a direct comparison to the classification landscape discussed in the main text, we report the corresponding landscape analyses for the regression tasks.

\begin{figure}
    \centering
    \begin{subfigure}{0.48\textwidth}
        \centering
        \includegraphics[width=\linewidth]{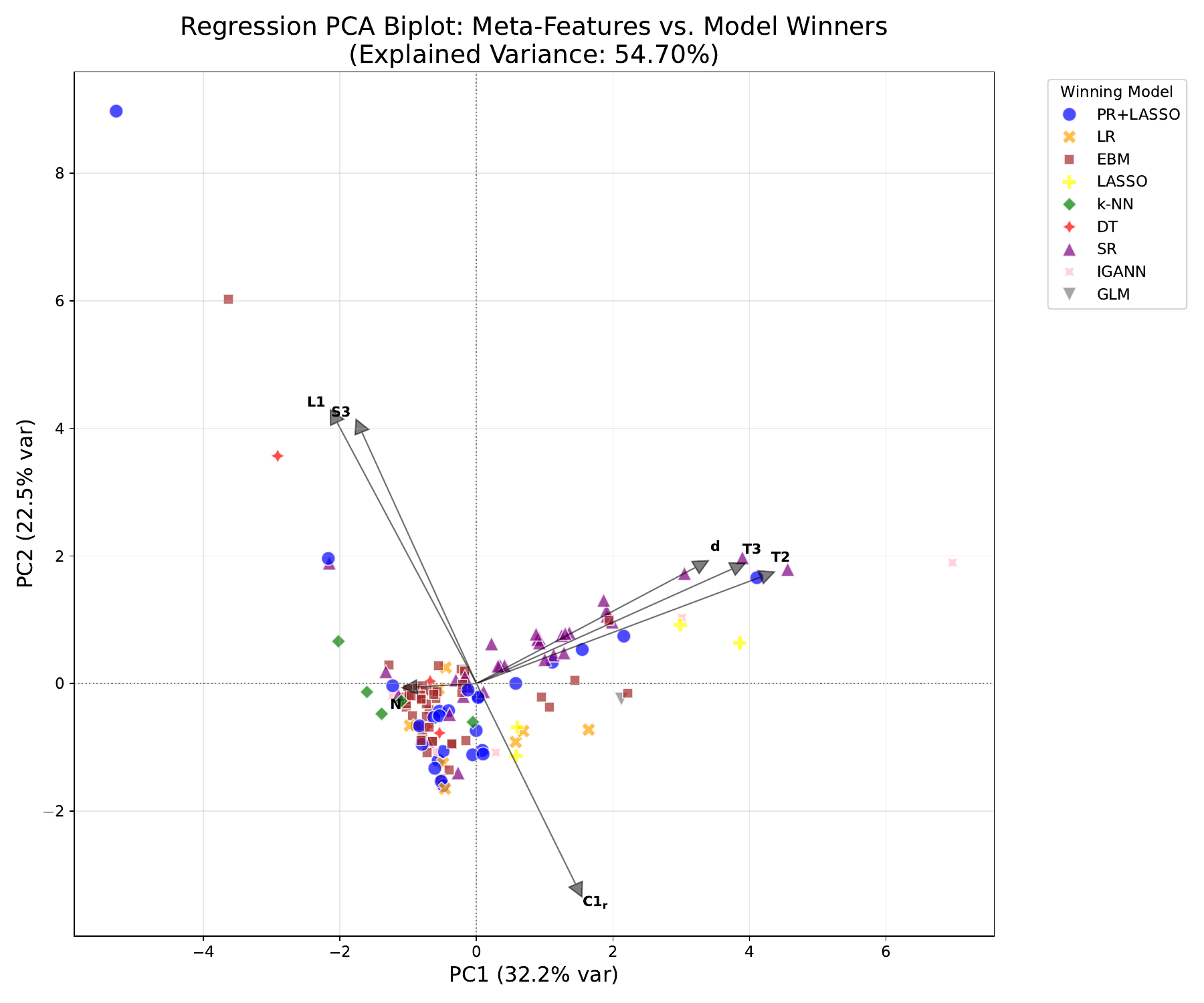}
        \caption{PCA biplot of the dataset meta-features (regression) projected onto the first two principal components. Individual datasets are color-coded by the winning algorithm.}
        \label{fig:regr_is_pca}
    \end{subfigure}\hfill
    \begin{subfigure}{0.48\textwidth}
        \centering
        \includegraphics[width=\linewidth]{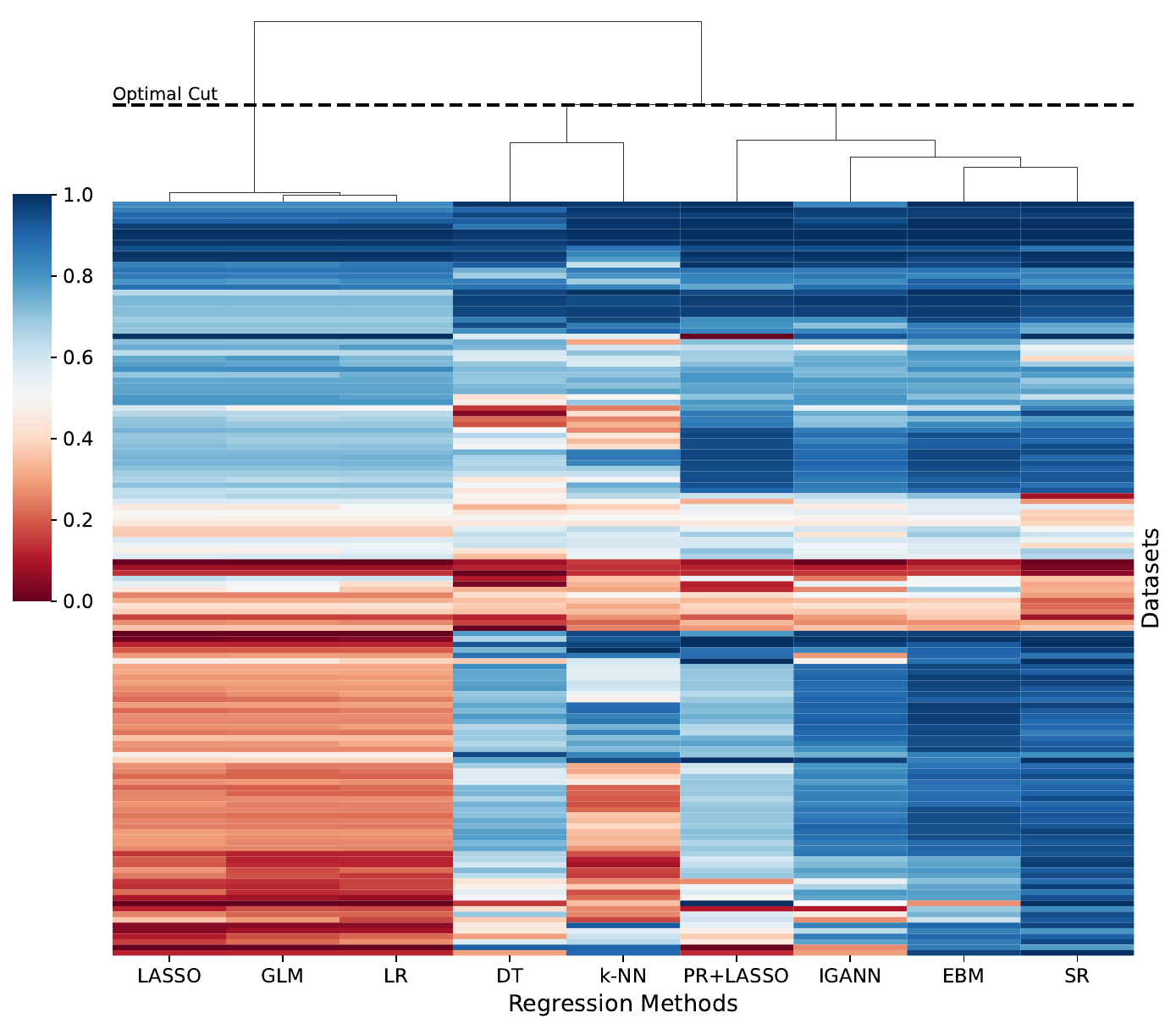}
        \caption{Hierarchical clustering of method performance profiles across regression datasets.}
        \label{fig:regr_is_cluster}
    \end{subfigure}
    \caption{Landscape analysis for regression tasks, displaying the PCA biplot of meta-features alongside the hierarchical clustering of method performance.}
    \label{fig:regr_landscape}
\end{figure}

As noted in our primary analysis, regression datasets show a clearer performance hierarchy and stronger stratification induced by the meta-features than classification datasets. To visualize the underlying structure of the regression benchmark, we first project the dataset meta-features into a reduced dimensional space using PCA. Figure \ref{fig:regr_is_pca} illustrates the resulting biplot, with datasets color-coded by the winning regression algorithm.

To further quantify the grouping of the performance profiles, we also apply the same hierarchical clustering approach to the algorithms' performances across all regression datasets. Figure \ref{fig:regr_is_cluster} presents the resulting dendrogram and clustermap. 

This clustering yielded a notably higher overall silhouette score of 0.54. When the algorithm was constrained to find a number of clusters equal to the number of evaluated methods, the silhouette score remained relatively stable at 0.30 (compared to 0.05 for classification). These visualizations and metrics corroborate the findings presented in the main text: the successes and failures of regression methods present distinct, measurable clusters that are more structurally cohesive than those observed in the classification benchmark.

\clearpage
\addtocounter{page}{-1}

\end{document}